\documentclass[a4paper,fleqn]{cas-dc}

\usepackage[authoryear,longnamesfirst]{natbib}
\usepackage{grffile} % 处理特殊路径

\usepackage[commandnameprefix=always]{changes}

\usepackage{xcolor}
\DeclareRobustCommand{\revised}[1]{#1}  % 直接显示内容，无颜色
\usepackage{graphicx}
\usepackage{subcaption}
\usepackage{adjustbox} % 高级对齐控制
\usepackage{caption}        % 必须

\usepackage{array}

\usepackage{tikz}
\usetikzlibrary{calc}

\def\tsc#1{\csdef{#1}{\textsc{\lowercase{#1}}\xspace}}
\tsc{WGM}
\tsc{QE}
\begin{document}
\let\WriteBookmarks\relax
\def\floatpagepagefraction{1}
\def\textpagefraction{.001}

\shorttitle{}    

\shortauthors{}  

\title [mode = title]{Depth-Consistent 3D Gaussian Splatting via Physical Defocus Modeling and Multi-View Geometric Supervision}  

\author[1]{Yu Deng}%[<options>]
\ead{202311093429@mail.scut.edu.cn}
% \ead[url]{202311093429@mail.scut.edu.cn}
\credit{Methodology, Software, Validation, Formal analysis, Investigation, Resources, Data Curation, Writing - Original Draft, Writing - Review \& Editing, Visualization}

\author[1]{Baozhu Zhao}%[<options>]
\ead{202320163293@mail.scut.edu.cn}
% \ead[url]{202320163293@mail.scut.edu.cn}
\credit{Conceptualization, Methodology, Investigation, Resources, Writing - Original Draft, Writing - Review \& Editing}

\author[1]{Junyan Su}%[<options>]
\ead{ft_su.junyan@mail.scut.edu.cn}
% \ead[url]{ft_su.junyan@mail.scut.edu.cn}
\credit{Validation, Formal analysis, Writing - Review \& Editing}

\author[1]{Xiaohan Zhang}%[<options>]
% \fnmark[3]
\ead{ftxiaohanzhang@mail.scut.edu.cn}
% \ead[url]{ftxiaohanzhang@mail.scut.edu.cn}
\credit{Writing - Review \& Editing}

\author[1]{Qi Liu}[orcid=0000-0001-5378-6404]
\cormark[1]
% \fnmark[4]
\ead{drliuqi@scut.edu.cn}
\ead[url]{https://drliuqi.github.io/}
\credit{Supervision, Writing - Review \& Editing, Funding acquisition}

\affiliation[1]{organization={Department of Future Technology, South China University of Technology},
            city={Guangzhou},
            postcode={511400}, 
            country={China}}

\cortext[1]{Corresponding author}

% \fntext[1]{Co-first authors.}


% \title [mode = title]{Depth-Consistent 3D Gaussian Splatting via Physical Defocus Modeling and Multi-View Geometric Supervision}  

\begin{abstract}
Three-dimensional reconstruction in scenes with extreme depth variations remains challenging due to inconsistent supervisory signals between near-field and far-field regions. Existing methods fail to simultaneously address inaccurate depth estimation in distant areas and structural degradation in close-range regions. This paper proposes a novel computational framework that integrates depth-of-field supervision and multi-view consistency supervision to advance 3D Gaussian Splatting. Our approach comprises two core components: (1) Depth-of-field Supervision employs a scale-recovered monocular depth estimator (e.g., Metric3D) to generate depth priors, leverages defocus convolution to synthesize physically accurate defocused images, and enforces geometric consistency through a novel depth-of-field loss, thereby enhancing depth fidelity in both far-field and near-field regions; (2) Multi-View Consistency Supervision employing LoFTR-based semi-dense feature matching to minimize cross-view geometric errors and enforce depth consistency via least squares optimization of reliable matched points. By unifying defocus physics with multi-view geometric constraints, our method achieves superior depth fidelity, demonstrating a 0.8 dB PSNR improvement over the state-of-the-art method on the Waymo Open Dataset. This framework bridges physical imaging principles and learning-based depth regularization, offering a scalable solution for complex depth stratification in urban environments.  
\end{abstract}


% \begin{highlights}
% \item Novel framework integrates depth-of-field and multi-view constraints for 3D Gaussian Splatting 
% \item Defocus convolution enforce geometric consistency
% \item Enhances depth fidelity in both far-field and near-field regions
% \end{highlights}

%\nocite{*}

% Keywords

\maketitle

% Main text

\section{Introduction}
Three-dimensional scene reconstruction from multi-view images remains a cornerstone capability for applications ranging from autonomous driving to immersive virtual reality. While Neural Radiance Fields (NeRF) \citep{mildenhall2021nerf} revolutionized photorealistic novel view synthesis, subsequent advances in 3D Gaussian Splatting (3DGS) \citep{kerbl20233d} have achieved unprecedented real-time rendering speeds through differentiable Gaussian primitives. However, these methods face critical limitations when reconstructing scenes with substantial depth variations, as distant structures often exhibit positional inaccuracies due to insufficient supervision signals and overfitting to training viewpoints.

Current approaches for large-scale scene reconstruction rely predominantly on multi-view stereo (MVS) techniques \citep{furukawa2015multi} or volumetric neural representations \citep{barron2022mip}. The former establishes geometric consistency through hand-crafted features \citep{xu2019multi}, while the latter optimizes implicit fields through photometric loss \citep{wang2023f2}. Recent extensions such as GaussianPro \citep{cheng2024gaussianpro} attempt to mitigate extreme distant interference through sky segmentation masks, yet introduce new artifacts from imperfect matting and sparse depth supervision. This limitation stems from a fundamental challenge: when objects appear at varying distances across frames, conventional supervision struggles to resolve scale ambiguities, particularly for distant regions receiving insufficient pixel-level gradients.

Our work addresses these limitations through two synergistic innovations. First, we leverage depth-of-field effects as implicit geometric supervision. By modeling the physical correlation between defocus blur and scene depth via adaptive kernel convolution, we derive gradient signals that guide Gaussians towards their geometrically consistent positions. Second, we introduce a hybrid depth estimation framework that integrates multi-view feature matching (LoFTR) \citep{wang2024efficient} with monocular depth completion (Metric3D) \citep{hu2024metric3d}, resolving scale ambiguities while preserving structural details. This dual strategy effectively constrains Gaussian distributions across varying depth layers, particularly benefiting distant regions reconstruction where traditional methods fail.

The technical contributions of this work are threefold:
\begin{itemize}
    \item A physics-aware defocus convolution model that translates optical principles into geometric constraints, using adaptive kernel designs (Gaussian, Polygonal, or SmoothStep) to adapt to different types of cameras and dynamic focus optimization to enhance depth consistency.
    
    \item A multiscale depth alignment framework combining global monocular depth recovery with local grid-based correction, achieving view-consistent depth estimation without manual masking.
    
    \item A gradient-aware density control mechanism that prioritizes structurally critical regions through depth-gradient statistics.
\end{itemize}

Extensive validation on urban (Waymo) and unbounded (Mip-NeRF 360) scenes demonstrates the efficacy of our approach. Quantitative results show that our model achieves 35.17 PSNR on Waymo, outperforming SOTA methods. Qualitative analyzes reveal significant improvements in near region structure recovery, particularly for vehicles and buildings. These advancements establish new state-of-the-art performance for depth-aware scene reconstruction while preserving the computational efficiency central to 3DGS frameworks.

\section{Related Work}

\subsection{Multi-View Stereo}
\label{subsec:MVS}

MVS represents a fundamental computer vision task that aims to reconstruct high-fidelity 3D models from a collection of calibrated images. The existing MVS methods can be broadly categorized into traditional geometry-based approaches and contemporary learning-based approaches.

Traditional MVS approaches typically derive their camera parameters predominantly from Structure-from-Motion (SfM) methods \citep{snavely2006photo, schonberger2016structure} or Simultaneous Localization and Mapping (SLAM) frameworks \citep{mur2015orb, engel2014lsd}. Within this paradigm, seminal methods proposed by \cite{campbell2008using}, \cite{furukawa2015multi}, and \cite{xu2019multi} establish explicit pixel correspondences through hand-crafted features and rigorous geometric constraints.

Learning-based approaches \citep{kar2017learning, ji2017surfacenet, zhou2023miper} have revolutionized MVS, initiated by the pioneering end-to-end architecture introduced by \cite{yao2018mvsnet}. Contemporary methods \citep{ma2022multiview, feng2023cvrecon} leverage learned representations for robust depth regression, while advanced techniques such as cascade cost volumes \citep{gu2020cascade} and feature matching networks \citep{giang2021curvature} have substantially enhanced both performance and computational efficiency.

\subsection{Neural Radiance Field}
\label{subsec:nerf}

A radiance field establishes a mapping from a 3D spatial coordinate $(x, y, z)$ and viewing directions parameterized by polar angle $\theta$ and azimuthal angle $\phi$ to a nonnegative radiance value, characterizing the light-matter interaction within the environment \citep{chen2024survey}. NeRF \citep{mildenhall2021nerf} pioneered the paradigm of representing a scene as an emissive volumetric function, implemented via a position-encoded neural network that enables differentiable rendering through volumetric quadrature.

Volumetric representations integrated with deep-learning techniques and volumetric ray-marching were initially proposed by \cite{sitzmann2019deepvoxels} and \cite{henzler2019escaping}. Significant advancements in this domain include Instant-NGP \citep{muller2022instant}, which employs multi-resolution hash grids, and Plenoxels \citep{fridovich2022plenoxels}, which utilizes sparse voxel grids for efficient optimization. Several approaches enhance rendering efficiency through scene reparameterization to generate more compact representations, notably Mip-NeRF 360 \citep{barron2022mip}, Zip-NeRF \citep{barron2023zip}, and F2-NeRF \citep{wang2023f2}.

Recent research has increasingly focused on addressing defocus blur in neural rendering, with seminal works including RawNeRF \citep{mildenhall2022nerf}, AR-NeRF \citep{kaneko2022ar}, and NeRFocus \citep{wang2022nerfocus}, all capable of synthesizing depth-of-field effects. Subsequently, Deblur-NeRF \citep{ma2022deblur} mitigates image blurriness resulting from defocus by implementing a Deformable Sparse Kernel module, while DP-NeRF \citep{lee2023dp} specifically addresses geometric and appearance consistency challenges in defocused scenarios. Nevertheless, as noted by \citep{wang2024dof}, these approaches continue to encounter substantial limitations regarding computational efficiency and real-time rendering capabilities.

\subsection{3D Gaussian Splatting}
\label{subsec:3dgs}

Novel view synthesis has evolved significantly with NeRF \citep{mildenhall2021nerf} setting a milestone for photorealistic rendering. Building upon this foundation, 3DGS \citep{kerbl20233d} introduces a paradigm shift that models scenes as a collection of 3D Gaussian primitives rendered via differentiable rasterization, simultaneously achieving high-quality reconstruction and real-time rendering capabilities. This approach extends traditional splatting-based rasterization \citep{zwicker2002ewa} by optimizing Gaussian primitives with explicit geometry and appearance attributes \citep{cheng2024gaussianpro, li2024mipmap}.

Recent advances in 3DGS research have predominantly focused on efficiency and quality enhancements. Innovative methods have been proposed to optimize Gaussian contributions through scale-based evaluation strategies \citep{lee2024compact} and sophisticated visibility assessment techniques \citep{fan2023lightgaussian}. Concurrent developments have yielded substantial improvements in rendering fidelity \citep{yu2024mip, blanc2024raygaussvolumetricgaussianbasedray, huang20242d}, computational efficiency \citep{girish2024eagles}, and the capacity to handle large-scale scenes with complex geometry \citep{kerbl2024hierarchical, liu2024citygaussian}.

Despite these significant advancements, fundamental limitations persist in the 3DGS representation framework. The inherent absence of true volumetric density fields manifests as view-dependent consistency issues and rendering artifacts \citep{radl2024stopthepop, mai2024ever}. Furthermore, the conventional pinhole camera model employed in standard 3DGS implementations inherently restricts its application domain to All-in-Focus (AiF) scenarios \citep{wang2024dof}.

Recent research has addressed these fundamental challenges through innovative depth-of-field rendering approaches. DOF-GS \citep{wang2024dof} introduces a finite aperture camera model coupled with explicit, differentiable defocus rendering guided by the Circle-of-Confusion (CoC), thereby enabling both adjustable depth-of-field effects and the generation of AiF images from defocused training data. Similarly, Cinematic Gaussians \citep{wang2024cinematic} leverages multi-view LDR images with varying exposure times, apertures, and focus distances to reconstruct high-dynamic-range (HDR) radiance fields, incorporating analytical convolutions of Gaussians based on a thin-lens camera model. Unlike these prior works that primarily focus on either depth-of-field effects rendering or HDR reconstruction, our approach uniquely leverages depth-of-field information as an additional structural supervision signal to achieve more geometrically accurate 3D reconstruction.

% physical_imaging_model
\begin{figure}
\centering
\includegraphics[width=0.95\columnwidth]{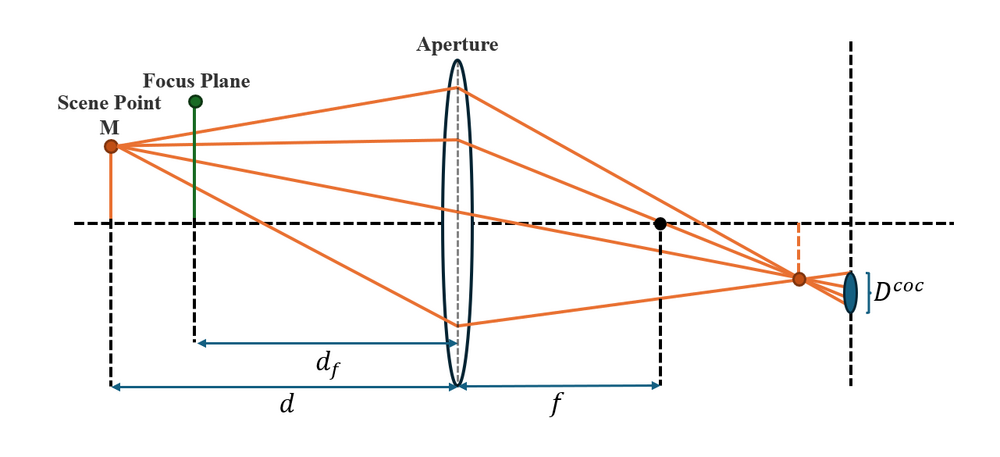}  
\caption{\revised{A schematic illustrating the principle of depth of field blur. When a scene point M at a distance d does not lie on the focus plane (at distance $d_f$), it creates a blurred spot on the image plane known as the circle of confusion (with a diameter of $D^{coc}$), causing the image to be out of focus. f represents the focal length of the lens.}}
\label{fig:image_model}
\end{figure}

\section{Method}

\begin{figure*}[t]  
\centering  
\includegraphics[width=0.95\textwidth]{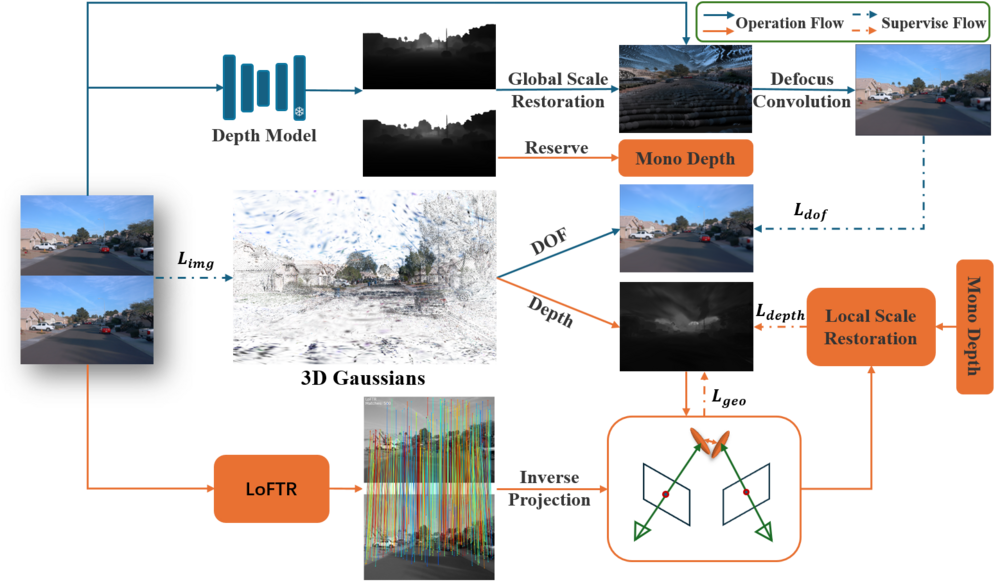}  
\caption{Our framework consists of two core technical components: (a) Depth-of-Field Supervision (Blue Flow) addressing inaccuracies in distant scenes and difficulties in recovering structures in near-field scenes. The pipeline takes multi-view images as input, obtains scale-ambiguous depth predictions through a monocular depth estimator (e.g., Metric3D), and calculates true depth maps via a multi-view depth scale recovery algorithm. Defocus convolution is then utilized to generate defocused images from both rendered and ground truth images, with the final $\mathcal{L}_{\text{dof}}$ loss between these defocused images supervising the 3DGS training. (b) Multi-View Consistency Supervision (Orange Flow) resolving cross-view geometric alignment issues. Initially, semi-dense feature matching is performed across multi-view images using LoFTR, minimizing the error $\mathcal{L}_{\text{geo}}$ between 3D points corresponding to matched pixels to enhance cross-view geometric consistency. Simultaneously, a depth consistency loss $\mathcal{L}_{\text{depth}}$ employs local depth maps recovered through least squares optimization from accurately matched points with reliable depth information to optimize the depth rendered by 3DGS.}  
\label{fig:pipeline}  
\end{figure*}

Our computational framework integrates dual supervisory paradigms for geometrically consistent 3D Gaussian Splatting, as depicted in Figure~\ref{fig:pipeline}. The architecture operates on multi-view inputs through dual-branch processing, establishing metric-scale depth priors while enforcing cross-view consistency constraints. Guided by this computational paradigm, we first introduce our depth-aware defocus modeling that formulates physical optics principles as differentiable geometric constraints. Subsequent sections systematically elaborate our hybrid depth estimation framework comprising global scale calibration and local depth refinement modules, culminating in multi-view geometric consistency enforcement.  

\subsection{Defocus Convolution}
\label{subsec:defocus_conv}

Accurate modeling of defocus effects constitutes an essential component in establishing comprehensive geometric and radiometric supervision signals \citep{cui2024dual} for 3D Gaussian Splatting-based scene reconstruction. This section presents a physics-driven defocus convolution framework that emulates physical imaging characteristics to achieve high-fidelity depth-of-field reconstruction through optical principle formulation.

\subsubsection{Physical Imaging Model}
\label{sssec:physical_model}

% \revised
\revised{As shown in \autoref{fig:image_model}, in optical imaging systems, scene points at distinct depth layers generate varying Circle of Confusion (CoC) dimensions through lens propagation.} Considering a 3D scene point with object distance \(d\), focal length \(f\), and focus distance \(d_f\), the CoC diameter in optical space is derived as:

\begin{equation}
D^{\text{(coc)}} = \frac{f^2 |d - d_f|}{F \cdot d \cdot (d_f - f)}
\label{eq:coc}  
\end{equation}

where \(F\) represents the f-number (defined as \(f/A\), with \(A\) denoting the physical aperture diameter). For digital imaging applications, we convert the optical CoC diameter to pixel space through sensor-image scaling:

\begin{equation}
D^{\text{(pixel)}} = D^{\text{(coc)}} \cdot \frac{w_i}{w_s}
\end{equation}

where \(D^{\text{(pixel)}}\) corresponds to the effective defocus diameter in digital coordinates, \(w_i\) indicates the image resolution width (pixels), and \(w_s\) specifies the sensor's physical width (mm).

\subsubsection{Adaptive Kernel Design}
\label{sssec:kernel_design}

\revised{Our framework incorporates three distinct convolution kernels to address diverse defocus characteristics found in optical systems. This multi-kernel approach provides the flexibility to balance physical realism with computational efficiency.}

\textbf{Gaussian Blur Kernel} \revised{For baseline defocus simulation, we employ a standard Gaussian kernel. Its softness provides a natural-looking blur. The kernel is defined as}:

\begin{equation}
G(x,y) = \frac{1}{2\pi\sigma^2}\exp\left(-\frac{x^2 + y^2}{2\sigma^2}\right)
\end{equation}

where \(\sigma\) relates to the circle of confusion (CoC) radius \(R\) through the formulation:

\begin{equation}
\sigma_G = \frac{D^{\text{(pixel)}}}{k_s}
\end{equation}

Here, \(k_s\) represents a normalization coefficient (default: 20) that scales the physical CoC diameter to kernel space. To ensure energy conservation, we enforce unitary integral constraint via kernel normalization:

\begin{equation}
K_{\text{norm}} = \frac{G}{\sum_{x,y} G(x,y)}
\end{equation}

\textbf{SmoothStep Blur Kernel} \revised{To better preserve sharp edges in regions with high depth discontinuity, a known limitation of Gaussian blur, we implement a hyperbolic tangent-based kernel. Its S-shaped transition profile offers a superior trade-off between blurring and edge preservation. The formulation is}:
\begin{equation}
K(x,y) = 0.5 + 0.5\tanh\left(0.25(r^2 - x^2 - y^2) + 0.5\right)
\end{equation}

where r denotes the kernel radius, and (x,y) represent the coordinates of each position within the kernel. This formulation achieves controlled edge preservation through its S-shaped transition profile, particularly effective for depth discontinuity regions.

% \textbf{Polygonal Blur Kernel} Modern camera lenses typically utilize 5-9 blades for aperture, while higher-end lenses often incorporating 9 or more rounded blades to approximate circle more closely (Fig.~\ref{fig:aperture_shape}). Our polygonal blur kernel employs 8 blades, which is a common configuration in mid-range lenses, to simulate polygonal bokeh produced by standard straight-blade design. Our parametric model explicitly encodes aperture geometry:

\begin{figure}[htbp]
\centering
\includegraphics[width=0.9\columnwidth]{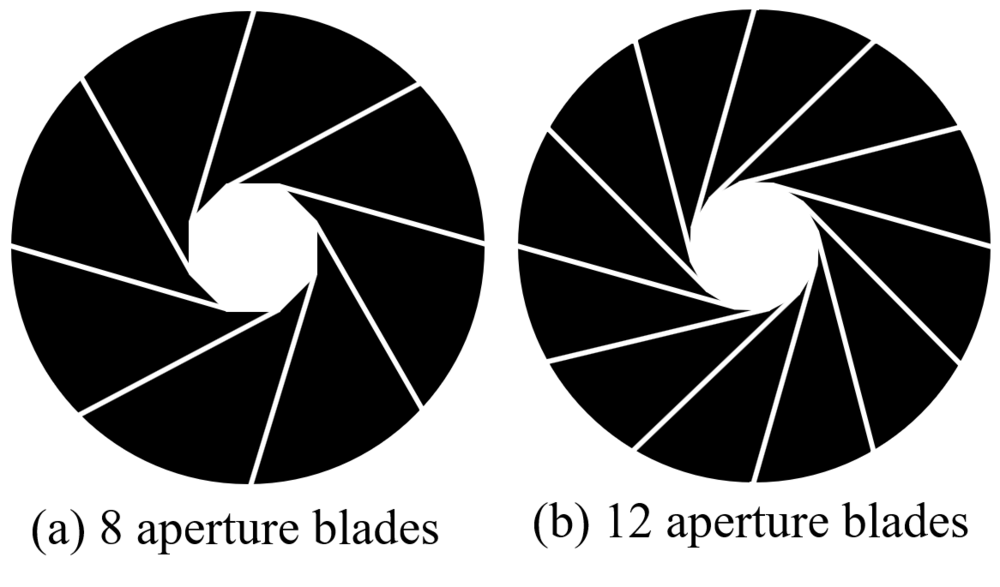}
\caption{Illustration of polygonal aperture mechanisms in a camera lens: (a) octagon aperture blades and (b) dodecagon aperture blades.}
\label{fig:aperture_shape}
\end{figure}

\textbf{Polygonal Blur Kernel} \revised{To achieve the highest degree of physical realism, especially for simulating the characteristic 'bokeh' from a lens's aperture blades, we introduce a parametric polygonal kernel. As shown in Fig.~\ref{fig:aperture_shape}, this model explicitly encodes the aperture geometry. The formulation is as follows:}

\begin{equation}
\alpha_i^{\text{Poly}}(p) = o_i \cdot \beta_i \cdot K(p)
\end{equation}

where \(N\) (default: 8) represents the number of aperture blades, \(\beta_i = 1/\sum_{p \in \Omega} K(p)\) ensures normalization, and \(K(p)\) integrates radial attenuation \(W(r)\) with geometric containment \(H(p)\):

\begin{equation}
K(p) = H(p) \cdot W(r(p))
\end{equation}

where $r(p) = \sqrt{x^2 + y^2}$ represents the Euclidean distance from point p to the center. The radial weight function employs cosine attenuation:

\begin{equation}
W(r) = \cos\left(\frac{\pi}{2} \cdot \frac{r}{R}\right) \cdot \mathbf{1}[r \leq R]
\end{equation}

where \( R \) denotes the predefined kernel radius, \(\mathbf{1}[r \leq R]\) is the indicator function. The vertices of the polygon are given by \(v_i = \left( R \cos\left(\frac{2\pi i}{N}\right), R \sin\left(\frac{2\pi i}{N}\right) \right)\) for \(i = 1, 2, \dots, N\), where \(v_{N+1} = v_1\) ensures the polygon closure.

The cross-product function \( C(p, v_i, v_{i+1}) \), which determines the relative position of point \( p \) with respect to the edge formed by vertices \( v_i \) and \( v_{i+1} \): 

\begin{equation}
\begin{split}
C(p, v_i, v_{i+1}) &= (v_{i+1,x} - v_{i,x})(p_y - v_{i,y}) \\
&\quad - (v_{i+1,y} - v_{i,y})(p_x - v_{i,x}).
\end{split}
\end{equation}

Using this cross-product criterion, the indicator function \( H(p) \), which determines whether a point \( p \) resides inside the \( N \)-sided polygon based on the cross-product criterion:

\begin{equation}
H(p) = \begin{cases}
1, & \text{if } \bigwedge_{i=1}^N C(p,v_i,v_{i+1}) < 0 \\
0, & \text{otherwise}
\end{cases}
\end{equation}

This parametric modeling accurately reproduces the optical characteristics observed in real-world Polygonal apertures, as evidenced by the qualitative comparison in Figure~\ref{fig:aperture_shape}(b).

% To better understand the blurring effects of different kernels, we visualize the results of various blur_kernels, as shown in Figure~\ref{fig:kernels}.

\begin{figure}[htbp]
\centering
\begin{subfigure}[b]{0.23\textwidth}
    \centering
    \includegraphics[width=\textwidth]{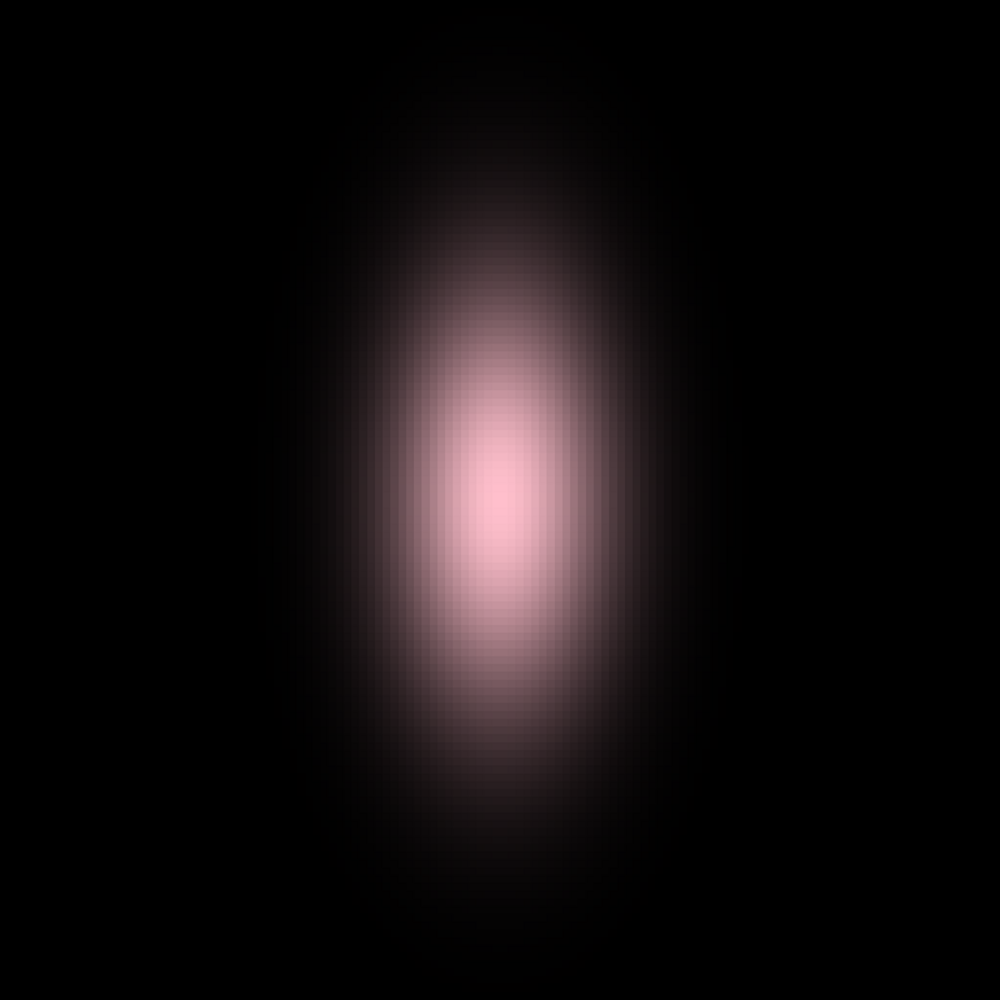}
    \caption{No Blur Kernel}
    \label{fig:no_kernel}
\end{subfigure}
\hfill
\begin{subfigure}[b]{0.23\textwidth}
    \centering
    \includegraphics[width=\textwidth]{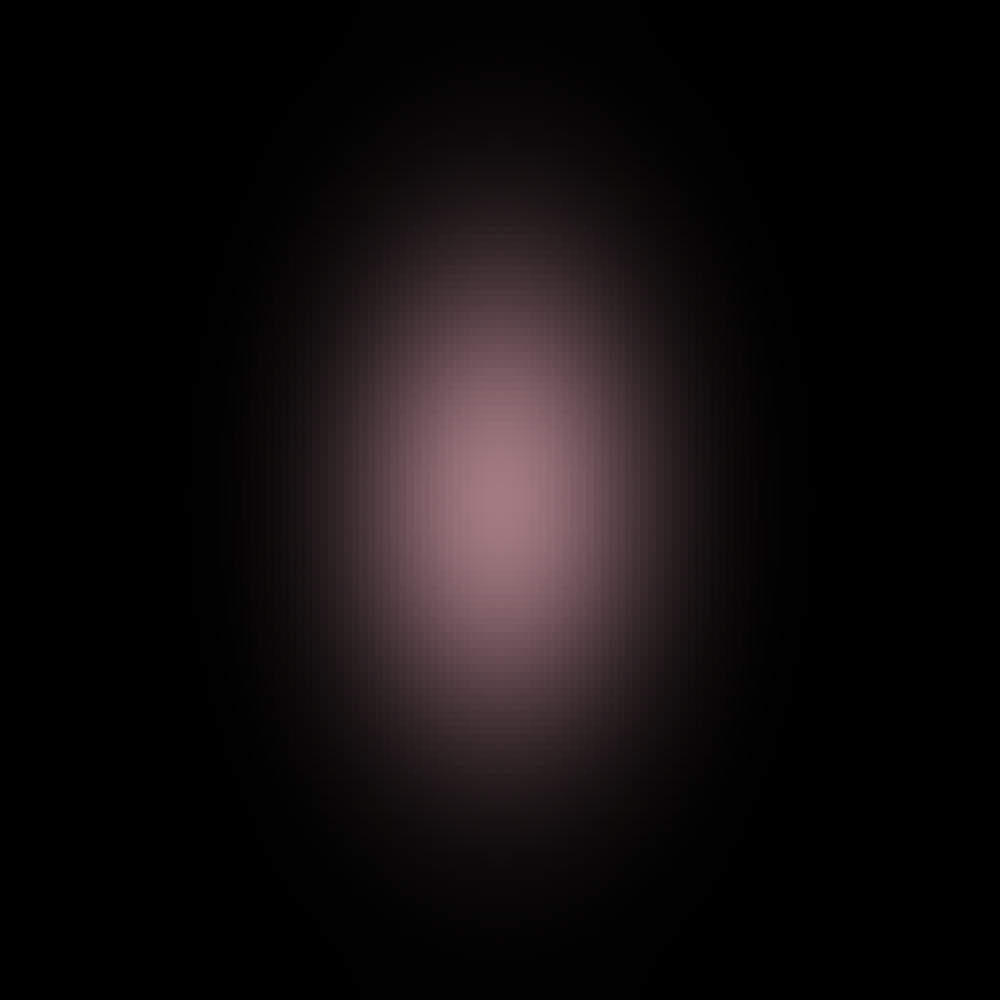}
    \caption{Gaussian Blur Kernel}
    \label{fig:gaussian_kernel}
\end{subfigure}
\hfill
\begin{subfigure}[b]{0.23\textwidth}
    \centering
    \includegraphics[width=\textwidth]{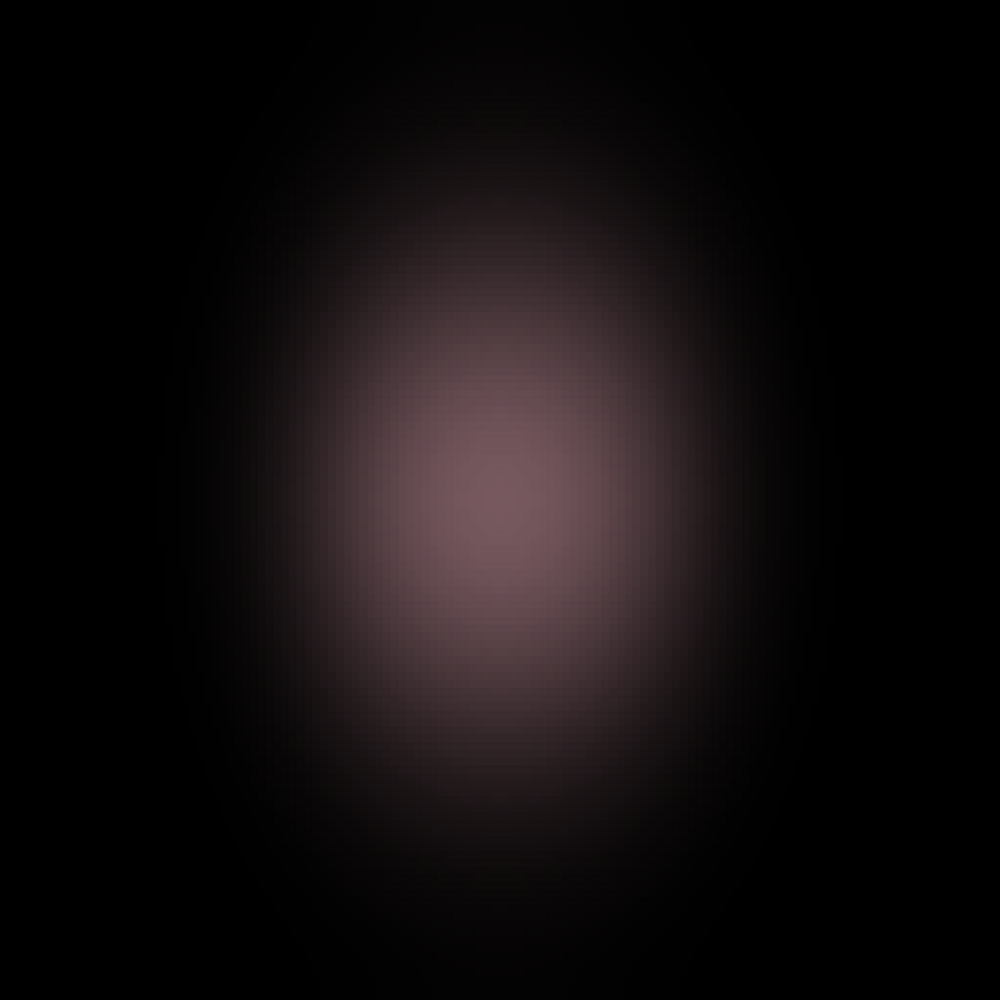}
    \caption{SmoothStep Blur Kernel}
    \label{fig:SmoothStep_kernel}
\end{subfigure}
\hfill
\begin{subfigure}[b]{0.23\textwidth}
    \centering
    \includegraphics[width=\textwidth]{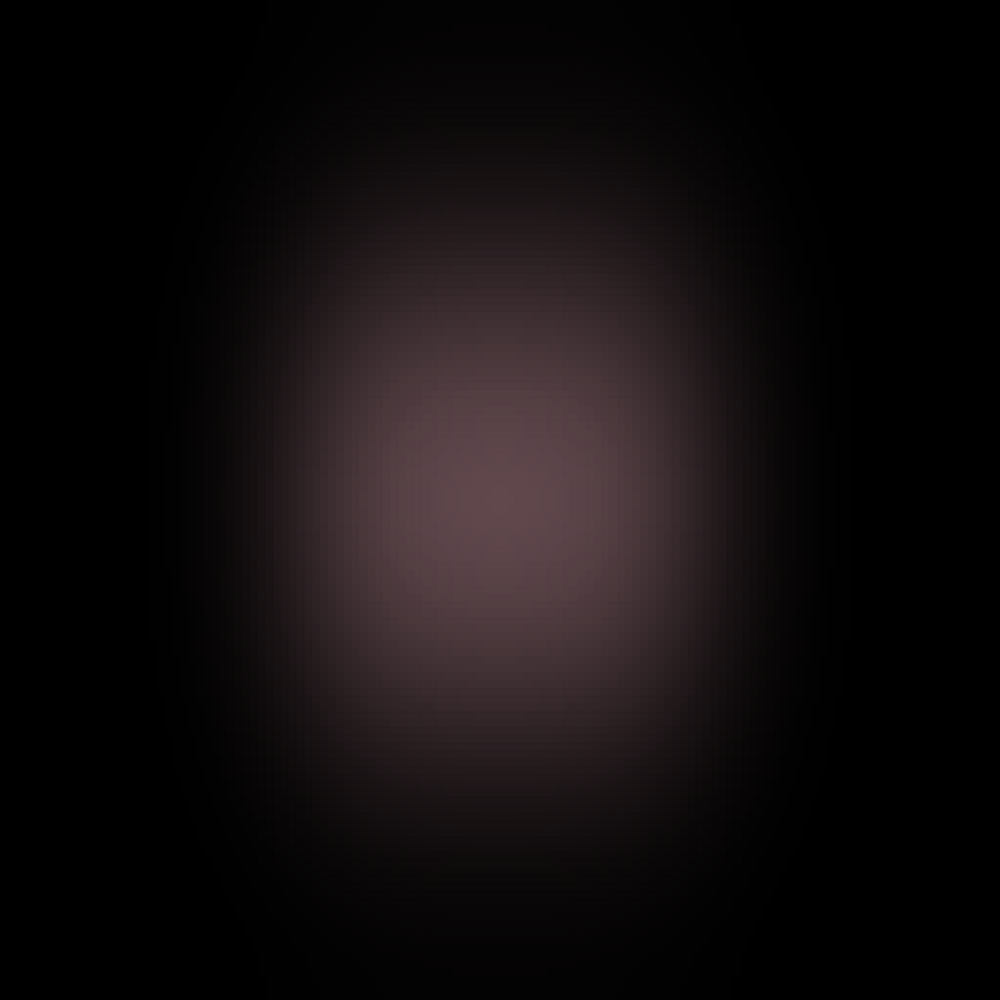}
    \caption{Polygonal Blur Kernel}
    \label{fig:Polygonal_kernel}
\end{subfigure}
\caption{Comparative analysis of defocus convolution techniques: 
(a) Original (no blur) provides baseline sharpness; 
(b) implementing radially symmetric blur via bell-shaped intensity profiles to simulate natural defocus; 
(c) preserving edge structures through S-curve transitions using hyperbolic tangent functions; 
(d) emulating optical apertures with geometric containment and radial attenuation for realistic bokeh effects.}
\label{fig:kernels}
\end{figure}

\revised{The distinct visual effects of these three kernels are compared in Figure~\ref{fig:kernels}, guiding the choice of kernel for different application requirements.}

\subsubsection{Dynamic Focus Optimization}
\label{sssec:focus_opt}

We optimize focus distance \(d_f\) using depth distribution statistics \(\{d_{1/3}, d_{1/2}, d_{2/3}, \mu_d\}\) (terciles and mean depth). The optimal focus distance minimizes:

\begin{equation}
d_f^* = \arg\min_{d_f} \sum_{p \in \Omega} w(p) \|d(p) - d_f\|^2
\end{equation}

\subsubsection{Defocus Synthesis}  
\label{sssec:defocus_synth}  

Given an input image \( I \), the final defocused image \( I_{\text{out}} \) is synthesized as follows:  

\begin{equation}  
I_{\text{out}}(p) = \sum_{q \in \mathcal{N}(p)} I(q) K\left(R(d_p)\right)(p - q) 
\label{eq:defocus_synthesis}  
\end{equation}  

where \(\mathcal{N}(p)\) denotes the neighborhood of pixel \( p \), \( K(R) \) represents the convolution kernel associated with the CoC size \( R \), which is determined by the depth \( d_p \) of the scene point corresponding to pixel \( p \).  

To enhance computational efficiency while maintaining physical accuracy, we implement a separable convolution approach. Specifically, we decompose the 2D convolution into two sequential 1D convolutions along the horizontal and vertical directions:  

\begin{equation}  
I_{\text{out}} = \left( I * k_x \right) * k_y  
\label{eq:separable_conv}  
\end{equation}  

where \( k_x \) and \( k_y \) are 1D kernel functions in the horizontal and vertical directions, respectively. This separable implementation reduces the computational complexity from \( \mathcal{O}(n^2) \) to \( \mathcal{O}(2n) \) for an \( n \times n \) kernel, significantly accelerating the processing while preserving the physical fidelity of the defocus effects.

\subsubsection{Depth-of-Field Loss}  
\label{subsec:dof_loss}  

% Our loss design addresses two critical aspects: standard image reconstruction and physical defocus effect alignment. This dual supervision ensures both geometric accuracy and optical plausibility.
\revised{Our loss design addresses two critical and distinct aspects of the reconstruction: ensuring photometric accuracy in sharp, in-focus regions, and enforcing physical plausibility of blur in out-of-focus regions. This dual supervision is key to achieving both geometric accuracy and optical realism.}

% The base reconstruction loss optimizes sharp in-focus regions through traditional metrics:
\revised{\textbf{Supervision for In-Focus Regions.} To optimize sharp regions, we employ a standard reconstruction loss, $\mathcal{L}_{\text{rgb}}$, which combines a pixel-wise L1 loss and a structural similarity (SSIM) term:}

\begin{equation}
\mathcal{L}_{\text{rgb}} = (1-\lambda_{\text{dssim}})\mathcal{L}_{\text{L1}}^{\text{(rgb)}} + \lambda_{\text{dssim}}\mathcal{L}_{\text{SSIM}}^{\text{(rgb)}}
\end{equation}

The constituent losses are defined as:
\begin{equation}
\mathcal{L}_{\text{L1}}^{\text{(rgb)}} = \frac{1}{|\Omega|}\sum_{p \in \Omega} \|I_{\text{rend}}(p) - I_{\text{gt}}(p)\|_1 
\label{eq:l1_rgb}
\end{equation}

\begin{equation}
\mathcal{L}_{\text{SSIM}}^{\text{(rgb)}} = 1 - \frac{(2\mu_{\text{rend}}\mu_{\text{gt}} + C_1)(2\sigma_{\text{rend}}\sigma_{\text{gt}} + C_2)}{(\mu_{\text{rend}}^2 + \mu_{\text{gt}}^2 + C_1)(\sigma_{\text{rend}}^2 + \sigma_{\text{gt}}^2 + C_2)}
\label{eq:ssim_rgb}
\end{equation}

where $\mu$ and $\sigma$ represent local means and standard deviations computed over $11\times11$ windows, with constants $C_1=0.01^2$ and $C_2=0.03^2$ preventing numerical instability during division. This component primarily supervises in-focus regions through direct pixel-wise comparison ($\mathcal{L}_{\text{L1}}$) and structural similarity preservation ($\mathcal{L}_{\text{SSIM}}$).
 
% To enforce physical accuracy in out-of-focus regions, we introduce defocus-specific losses:
\revised{\textbf{Supervision for Out-of-Focus Regions.} To specifically supervise the physically-based defocus effects, we introduce a dedicated loss term, $\mathcal{L}_{\text{dof}}$. This loss is computed between the ground-truth image convolved with our physics-based kernel and the rendered image similarly convolved. It shares the same structure as the in-focus loss but operates on the defocused images:}

\begin{equation}
\mathcal{L}_{\text{L1}}^{\text{(dof)}} = \frac{1}{|\Omega|}\sum_{p \in \Omega} \|I_{\text{rend}}^{\text{(dof)}}(p) - I_{\text{gt}}^{\text{(dof)}}(p)\|_1 
\label{eq:l1_dof}
\end{equation}

\begin{equation}
\mathcal{L}_{\text{SSIM}}^{\text{(dof)}} = 1 - \text{SSIM}(I_{\text{rend}}^{\text{(dof)}}, I_{\text{gt}}^{\text{(dof)}})
\label{eq:ssim_dof}
\end{equation}

% The composite defocus loss adaptively combines these metrics:
\revised{\textbf{Composite Defocus Loss.} These components are combined into the final depth-of-field loss $\mathcal{L}_{\text{dof}}$:}

\begin{equation}
\mathcal{L}_{\text{dof}} = (1-\lambda_{\text{dssim}}^{\text{(dof)}})\mathcal{L}_{\text{L1}}^{\text{(dof)}} + \lambda_{\text{dssim}}^{\text{(dof)}}\mathcal{L}_{\text{SSIM}}^{\text{(dof)}}  
\label{eq:dof_loss}  
\end{equation}

% This hierarchical loss structure enables separate optimization of focus accuracy and defocus physics, which is crucial for high-quality depth-of-field synthesis.
\revised{This hierarchical structure allows for targeted optimization of focus accuracy and defocus physics, which is crucial for high-quality synthesis.}

\subsubsection{Depth-Aware Gaussian Density Control}
\label{subsec:dof_density_control}

Traditional 3D Gaussian reconstruction struggles in defocused regions due to uniform density allocation, which limits the effectiveness of Gaussian points for downstream tasks \citep{keypoint}. Our gradient-aware strategy adaptively modulates point density using optical gradients derived from the CoC physics. The proposed approach achieves three fundamental improvements over conventional methods: (1) Edge preservation through gradient-sensitive prioritization of points in depth discontinuity regions, (2) Adaptive efficiency via automatic density balancing based on quantile-based gradient statistics, and (3) Physical consistency inherited from the optical imaging model through CoC constraints. The control parameter $\tau \in [0,1]$ governs the quality-efficiency trade-off, where higher $\tau$ values enhance reconstruction fidelity at the cost of increased computational resources, as formalized in the preservation criterion.

\textbf{Gradient Computation}  
The CoC gradient magnitude $\nabla\mathcal{L}_{\text{DoF}}$ is computed through differentiable rendering:
\begin{equation}
\nabla\mathcal{L}_{\text{DoF}} = \frac{\partial \mathcal{L}_{\text{\revised{dof}}}}{\partial \mathbf{x}}
\end{equation}

\revised{the definition of $\mathcal{L}_{\text{dof}}$ refers to Section~\ref{subsec:dof_loss}.}

\textbf{Adaptive Density Modulation}  
A quantile-based preservation criterion prioritizes structurally critical regions:
\begin{equation}
\mathcal{M}_{\text{keep}} = \mathbb{I}\left[ \|\nabla\mathcal{L}_{\text{DoF}}\| \geq Q_\tau\left( \{\|\nabla\mathcal{L}_{\text{DoF}}\|\} \right) \right]
\end{equation}
where $\tau=0.2$ preserves the top 20\% of points with the highest gradients, and $Q_\tau$ denotes the $\tau$-th quantile.

\textbf{Pruning Criterion}  
The pruning logic combines opacity thresholding with gradient-aware selection:
\begin{equation}
\mathcal{M}_{\text{prune}} = (\alpha < \alpha_{\min}) \lor \left[(\|\nabla\mathbf{x}\| < g_{\min}) \land \neg\mathcal{M}_{\text{keep}}\right]
\end{equation}

where $\alpha_{\min}$ and $g_{\min}$ are empirically determined thresholds for opacity and spatial gradients, respectively. The logical operators $\lor$ (OR), $\land$ (AND), and $\neg$ (NOT) implement the compound pruning condition. 

\revised{The base values for $\alpha_{\min}$ and $g_{\min}$ are identical to those in the original 3DGS framework (\citep{kerbl20233d}), ensuring a fair comparison. Our key contribution is the adaptive criterion $\mathcal{M}_{\text{keep}}$, which introduces a principled, physics-aware mechanism that modulates the pruning process based on geometric significance. It is important to note that our method acts as a protective supplement to the original densification logic, rather than adaptively lowering the global threshold. As detailed in Appendix~\ref{sec:appendix_densify}, our experiments confirm that simply tuning the global threshold is a fragile strategy that can lead to training instability, whereas our approach provides a stable performance improvement.}

\textbf{Gradient Statistics}  
Interquartile-range weighting prevents outlier dominance in gradient accumulation:

\begin{equation}
w_i = \exp\left(-\frac{\|\nabla\mathcal{L}_{\text{DoF},i}\| - Q_{0.25}}{Q_{0.75} - Q_{0.25} + \epsilon}\right)
\end{equation}
\begin{equation}
\hat{G}_i = \hat{G}_i + w_i\|\nabla\mathcal{L}_{\text{DoF},i}\|
\end{equation}

with $Q_{0.25}$ and $Q_{0.75}$ denoting the 25th and 75th percentiles of gradient magnitudes.

\subsection{Global Scale Recovery for Monocular Depth Maps}
\label{sec:scale_recovery}

\subsubsection{Motivation for Scale Recovery}
Monocular depth estimation methods \citep{almalioglu2022selfvio, liu2024joint} estimate depth values up to an unknown scale factor for each image, resulting in metric inconsistencies across multi-view observations. These scale ambiguities lead to misaligned geometries when integrating depth maps into 3DGS. Our scale recovery algorithm jointly optimizes per-image parameters to ensure view-consistent depth relationships, which are critical for realistic depth-of-field synthesis.

\subsubsection{Scale Recovery via Geometric Consistency}
\label{ssec:scale_optimization}

% we model scaled depth as:
% Given $N$ images with camera parameters $\{(\mathbf{K}_i, \mathbf{R}_i, \mathbf{t}_i)\}$ and monocular depth maps $\{D_i\}$, we model scaled depth as:
\revised{Given $N$ images with camera parameters $\{(\mathbf{K}_i, \mathbf{R}_i, \mathbf{t}_i)\}$ and their corresponding raw monocular depth maps $\{D_m\}$, we model the scaled depth at a pixel coordinate $\mathbf{p}$ as:}

\begin{equation}
    % \tilde{D}_i(\mathbf{p}) = s_i \cdot D_i(\mathbf{p}) + b_i
    \tilde{D}_m(\mathbf{p}) = s_i \cdot D_m(\mathbf{p}) + b_i
    \label{eq:depth_model}
\end{equation}

where $s_i > 0$ and $b_i$ are learnable scale and shift parameters, respectively.

\textbf{Feature Matching} For image pairs $(I_i, I_j)$, we employ the LoFTR descriptor to establish semi-dense local feature correspondences.
    
\textbf{Theoretical Depth Ratio} For valid matches $(\mathbf{p}_i^k \leftrightarrow \mathbf{p}_j^k)$, we compute:

\begin{equation}
    \gamma_{ij}^k = \mathbf{e}_z^\top \left( \mathbf{R}_{ji}\mathbf{x}_i^k + \frac{\mathbf{t}_{ji}}{Z_i^k} \right)
    \label{eq:theory_ratio}
\end{equation}

where $\mathbf{x}_i^k = \mathbf{K}_i^{-1}\mathbf{p}_i^k$ \revised{and $Z_i^k \approx \tilde{D}_m(\mathbf{p}_i^k)$}.
    
\textbf{Joint Optimization} \revised{We define an objective function that is minimized over the set of all scale and shift parameters $\{s_i, b_i\}_{i=1}^N$ across all images. The function combines a reprojection term and a depth ratio consistency term:}

\begin{equation}
\begin{split}
loss = & \sum_{\substack{(i,j,k)}} \underbrace{\left\| \pi(\mathbf{R}_j\tilde{\mathbf{X}}_i^k + \mathbf{t}_j) - \mathbf{p}_j^k \right\|^2}_{\text{Reprojection}} \\
       & + \lambda \sum_{\substack{(i,j,k)}} \underbrace{\left( \frac{\revised{D_m(\mathbf{p}_j^k)}}{\revised{D_m(\mathbf{p}_i^k)}} - \gamma_{ij}^k \right)^2}_{\text{Ratio Consistency}}
\end{split}
\label{eq:loss}
\end{equation}

where $\tilde{\mathbf{X}}_i^k = \revised{\tilde{D}_m^k}\mathbf{x}_i^k$ and $\lambda=0.5$. \revised{This optimization yields the optimal parameters $\{s_i^*, b_i^*\}$, which represent the globally consistent scale and shift for each view.}
    
\textbf{Depth Alignment} \revised{We apply the optimized parameters $\{s_i^*, b_i^*\}$ to the raw monocular depths to obtain the final, globally aligned depth maps $\{D_a\}$:}

\begin{equation}
    % \tilde{D}_i^* = s_i^*D_i + b_i^*
    D_a(\mathbf{p}) = s_i^* D_m(\mathbf{p}) + b_i^*
    \label{eq:final_depth}
\end{equation}

The aligned depths \revised{$\{D_a\}$} from Equation \eqref{eq:final_depth} enable consistent blur synthesis through 3DGS as described in Section~\ref{sssec:defocus_synth}.

\subsection{Geometric Consistency Supervision via Feature Matching}

\subsubsection{Depth Rendering with 3DGS}

For a given viewpoint, \revised{the rendered depth value $D_r(\mathbf{x})$ at pixel coordinate $\mathbf{x}$} is computed through alpha compositing of $K$ ordered Gaussians:

\begin{equation}
    % \mathbf{D}(\mathbf{x}) = \sum_{k=1}^K \alpha_k T_k d_k
    \revised{D_r(\mathbf{x})} = \sum_{k=1}^K \alpha_k T_k d_k
    \label{eq:depth_rendering}
\end{equation}

where $T_k = \prod_{l=1}^{k-1}(1-\alpha_l)$ denotes the cumulative transmittance up to the $k$-th Gaussian, $\alpha_k$ is the opacity of the $k$-th Gaussian, and $d_k$ represents the depth value of the Gaussian center.

\revised{Note that this formulation is the standard for depth rendering via alpha compositing, analogous to the color rendering equation in the original 3DGS framework \citep{kerbl20233d}. The sum of weights, $\sum \alpha_k T_k$, represents the total accumulated alpha. For the opaque surfaces targeted by our reconstruction, the optimization process naturally drives this sum to converge to a value near 1. Deviations from 1 are physically meaningful, representing rays viewing empty space or semi-transparent geometry.}

\subsubsection{Cross-view Feature Matching}
For image pairs $(I_i, I_j)$, we extract semi-dense correspondences using a pre-trained LoFTR model:

\begin{equation}
    \mathcal{M}_{ij} = \{(\mathbf{x}_i^{(m)}, \mathbf{x}_j^{(m)})\}_{m=1}^M
    \label{eq:matches}
\end{equation}

where matches are filtered by a confidence threshold $\tau=0.5$.

\begin{figure}
    \centering
    \includegraphics[width=0.9\linewidth]{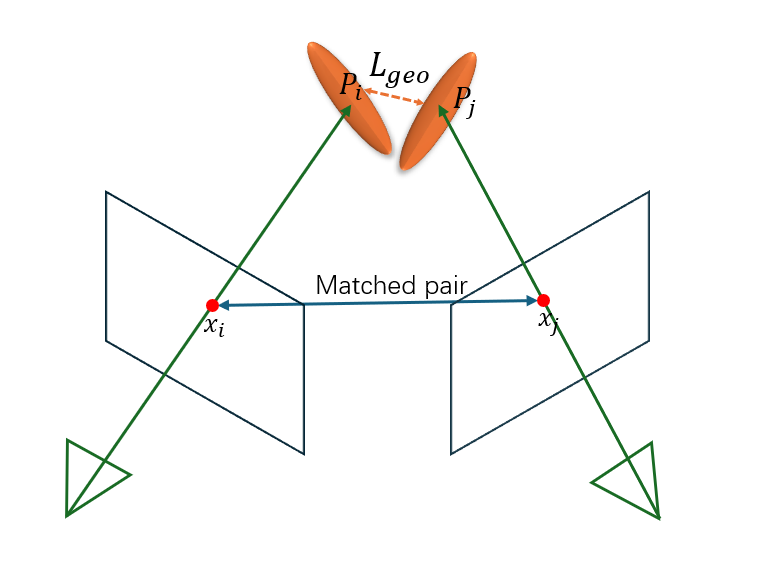}
    \caption{\revised{Schematic Diagram of Geometric Consistency Loss Calculation: Feature points matched in two views are projected into 3D space using rendered depth, and geometric constraints are imposed by minimizing the distance between these projected points.}}
    \label{fig:L_geo}
\end{figure}

\subsubsection{Geometric Consistency Loss}

% \revised
\revised{As shown in \autoref{fig:L_geo}, for each matched pair $(\mathbf{x}_i^{(m)}, \mathbf{x}_j^{(m)}) \in \mathcal{M}_{ij}$}:

1) Project 2D matches to 3D space:
\begin{align}
    \mathbf{P}_i^{(m)} &= \pi_i^{-1}(\mathbf{x}_i^{(m)}, \revised{D_r(\mathbf{x}_i^{(m)})}) \label{eq:proj1} \\
    \mathbf{P}_j^{(m)} &= \pi_j^{-1}(\mathbf{x}_j^{(m)}, \revised{D_r(\mathbf{x}_j^{(m)})}) \label{eq:proj2}
\end{align}

2) Compute position discrepancy:
\begin{equation}
    \mathcal{L}_{\text{geo}} = \frac{1}{|\mathcal{M}|}\sum_{m=1}^M \|\mathbf{P}_i^{(m)} - \mathbf{P}_j^{(m)}\|_1
    \label{eq:loss_geo}
\end{equation}

\subsection{Local Scale Restoration}  
\label{subsec:local_scale}  

Accurate depth estimation constitutes a fundamental requirement for effectively harnessing depth-of-field effects to strengthen structural supervision within 3D Gaussian Splatting. While monocular depth estimation offers initial depth cues, its inherent scale ambiguity prevents direct application. To address this limitation, we propose a local scale restoration framework through adaptive regional analysis.  

\subsubsection{Grid-based Local Regions}  
\label{sssec:grid}  

% Given a rendered depth map \(D_r \in \mathbb{R}^{H \times W}\) and monocular depth map \(D_m \in \mathbb{R}^{H \times W}\),
\revised{Given a rendered depth map $D_r \in \mathbb{R}^{H \times W}$ and monocular depth map $D_m \in \mathbb{R}^{H \times W}$, we decompose the image domain into a grid of local regions. To ensure our method is robust to varying image resolutions, we employ an adaptive strategy. Instead of fixing the number of grid cells $(h, w)$, we constrain the size of each cell, $(g_h, g_w)$, to be within an empirically established range $[g_{\min}, g_{\max}]$ (15 to 60 pixels). This range balances two factors: cells must be large enough for stable parameter estimation ($g_{\min}$) but small enough to capture local depth variations ($g_{\max}$). The number of cells $(h, w)$ is then derived from the image dimensions $(H, W)$ to meet these size constraints. Integer division is used to tile the grid, ensuring all boundary pixels are included in the final row and column of cells without clipping.}  

To guarantee robust parameter estimation, each grid cell must maintain a minimum of 5 valid feature points. This density constraint prevents numerical instability during local linear transformation computation, particularly addressing challenges from sparse or non-uniform feature distributions.  

\subsubsection{Local Linear Transformation}  
\label{sssec:linear_trans}  

Within each grid cell \(G_{i,j}\), we establish a parametric mapping between rendered depth \(D_r\) and monocular depth \(D_m\) through:  

\begin{equation}  
D_r^{(i,j)} = s_{i,j} D_m^{(i,j)} + t_{i,j},  
\end{equation}  

where \(s_{i,j}\) (scale factor) and \(t_{i,j}\) (translation offset) denote grid-specific transformation parameters. These parameters are optimized via \textbf{Tikhonov-regularized least squares} minimization:  

\begin{equation}  
\min_{s_{i,j}, t_{i,j}} \sum_{p \in G_{i,j}} \left\| D_r(p) - \left( s_{i,j} D_m(p) + t_{i,j} \right) \right\|_2^2 + \lambda \left\| \begin{bmatrix} s_{i,j} \\ t_{i,j} \end{bmatrix} \right\|_2^2,  
\end{equation}  

where \(p\) indexes valid feature points within the grid cell, and \(\lambda = 10^{-6}\) serves as the regularization coefficient to ensure numerical stability in ill-posed conditions.  

The closed-form solution derives from the normal equations formulation:  

\begin{equation}  
\begin{bmatrix}  
s_{i,j} \\  
t_{i,j}  
\end{bmatrix}  
= \left( X^\top X + \lambda I \right)^{-1} X^\top y,  
\end{equation}  

with design matrix \(X = \begin{bmatrix} D_m(p) & \mathbf{1} \end{bmatrix}\) containing monocular depth measurements and an intercept term, and observation vector \(y = D_r(p)\) comprising rendered depth values.  

\subsubsection{Depth Error Map Generation and Visualization}  
\label{sssec:error_map}  

The depth error \(E_{i,j}\) for each grid cell is computed as the mean absolute difference between the rendered depth values and their corresponding transformed monocular depth estimates:  

\begin{equation}  
E_{i,j} = \frac{1}{|G_{i,j}|} \sum_{p \in G_{i,j}} \left| D_r(p) - \left( s_{i,j} D_m(p) + t_{i,j} \right) \right|,  
\end{equation}  

where \(|G_{i,j}|\) represents the number of valid points in grid cell \(G_{i,j}\), \(D_r(p)\) denotes the rendered depth at point \(p\), and \(s_{i,j}D_m(p) + t_{i,j}\) is the transformed monocular depth using the optimized local linear transformation parameters. 

These grid-level errors are then interpolated to the original image resolution to form a comprehensive depth error map \(E \in \mathbb{R}^{H \times W}\). To enhance visualization contrast and handle grid boundaries and regions with insufficient feature points, we apply the following min-max normalization:  

\begin{equation}  
\hat{E}(p) = \begin{cases}  
\frac{E(p) - E_{\min}}{E_{\max} - E_{\min}}, & \text{if } E(p) \neq 1 \text{ and } E(p) \text{ is finite}, \\  
1, & \text{otherwise}.  
\end{cases}  
\end{equation}  

where \(E_{\min}\) and \(E_{\max}\) represent the minimum and maximum error values among all valid pixels, respectively. The default value of 1 is preserved for pixels in invalid regions or grid cells with insufficient feature points.  

\begin{figure*}[htbp]
\centering
% 容器确保整体左右有边距且居中
\begin{minipage}{0.98\textwidth}
    % --- 步骤 1: 确定一个基准高度 ---
    % 我们需要为所有图片选择一个统一的高度。
    % 一个好的起点是根据您最“正常”的图片来估算。
    % 假设我们选择 4cm 作为目标高度。
    \newcommand{\myfigureheight}{4cm} % 定义一个变量，方便统一修改

    \begin{subfigure}[t]{0.32\linewidth} % 使用 [t] 或 [b] 都可以，因为现在高度一致
        \centering
        % --- 关键修改: 同时强制 width 和 height ---
        % width=\linewidth 确保图片填满子图容器的宽度
        % height=\myfigureheight 确保图片具有我们设定的统一高度
        \includegraphics[width=\linewidth, height=\myfigureheight]{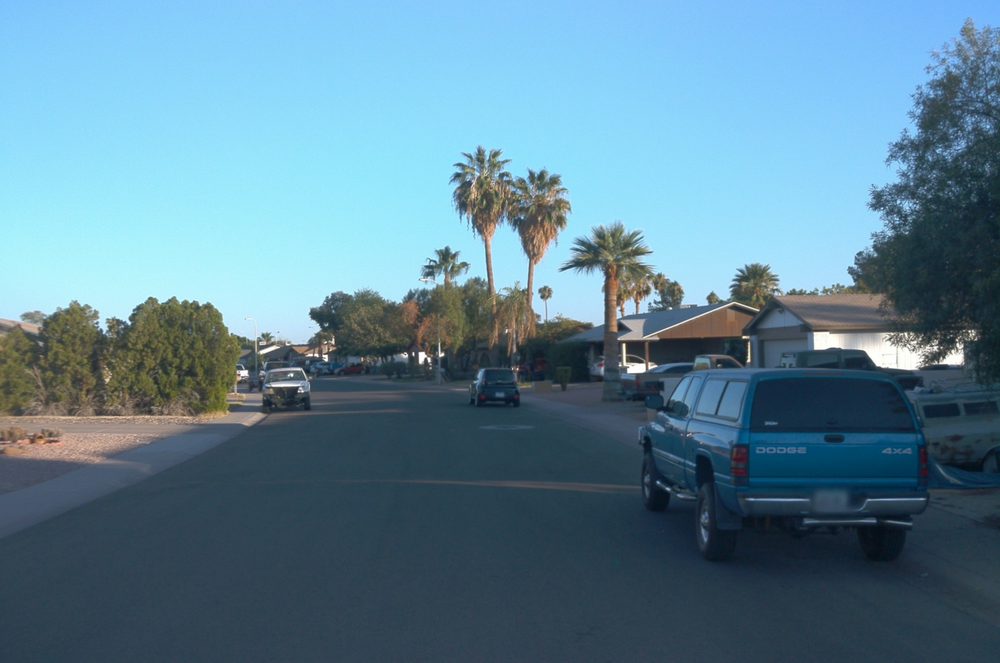}
        \caption{Ground Truth}
        \label{fig:no_kernel}
    \end{subfigure}
    \hfill 
    \begin{subfigure}[t]{0.32\linewidth}
        \centering
        \includegraphics[width=\linewidth, height=\myfigureheight]{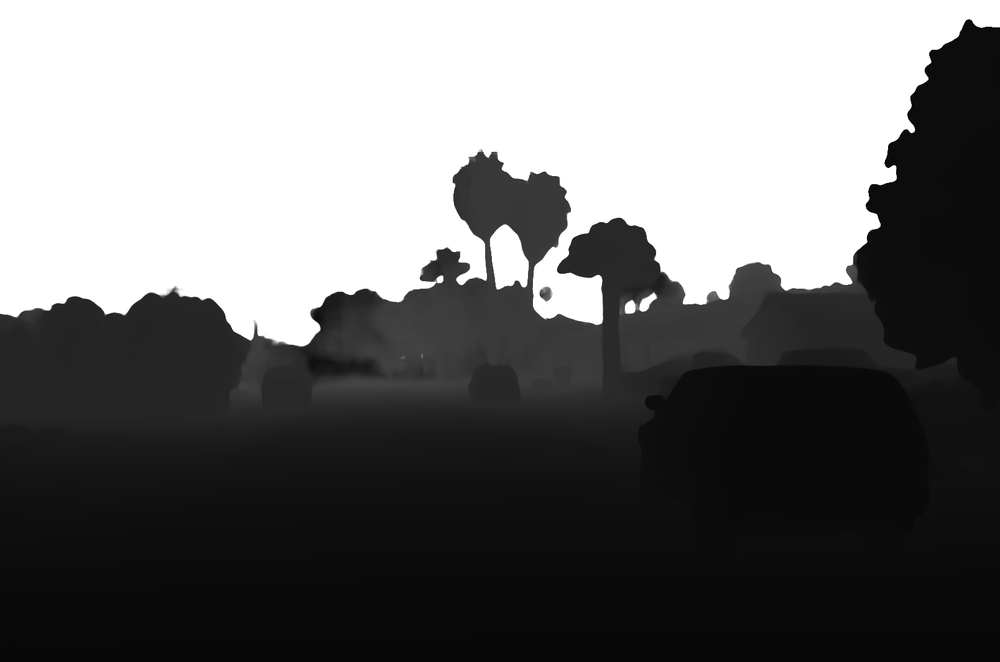}
        \caption{Monocular Estimation}
        \label{fig:gaussian_kernel}
    \end{subfigure}
    \hfill
    \begin{subfigure}[t]{0.32\linewidth}
        \centering
        \includegraphics[width=\linewidth, height=\myfigureheight]{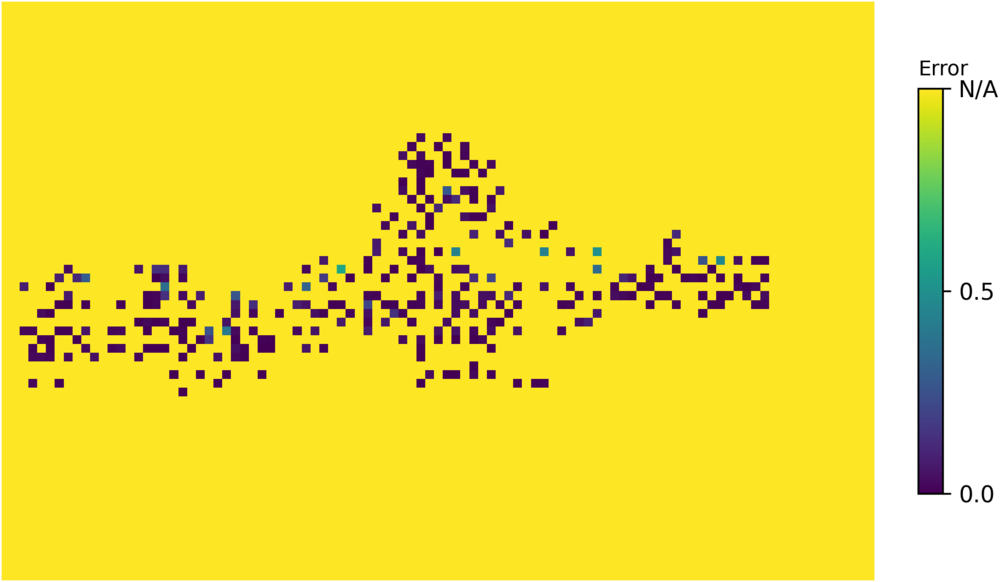}
        \caption{Error Distribution}
        \label{fig:SmoothStep_kernel}
    \end{subfigure}
\end{minipage}
\caption{Comparative visualization of depth estimation components: (a) Ground truth, (b) Monocular depth prediction from our proposed method, (c) \revised{Spatial error mapping. Error is computed only in regions with sufficient feature matches (non-yellow areas) and normalized to [0, 1), where cooler colors indicate lower error. The 'N/A' label on the color bar denotes regions where error was not computed.}}
\label{fig:depth_visualization}
\end{figure*}

To quantitatively analyze the spatial distribution characteristics of depth estimation errors, Figure~\ref{fig:depth_visualization} provides a multi-modal comparison comprising three aligned representations:
(a) Ground truth RGB image,
(b) Monocular depth estimation results, and
(c) Corresponding disparity error heatmap generated by our method.

This multi-view visualization quantitatively reveals spatial error concentration patterns and pinpoints challenging regions characterized by complex geometric configurations, occlusion boundaries, and fine-scale structural details.

\subsubsection{Depth Consistency Loss}  
\label{sssec:depth_loss}  

We design a depth consistency loss function to optimize the alignment between rendered and monocular depths:  
\begin{equation}  
\mathcal{L}_{\text{depth}} = \mathcal{L}_{\text{abs}} + \alpha \mathcal{L}_{\text{corr}},  
\end{equation}  
where $\alpha$ is a balancing coefficient empirically determined to regulate the influence of correlation error.  

where the \textbf{absolute error term} \(\mathcal{L}_{\text{abs}}\) penalizes deviations in valid regions:  
  \begin{equation}  
  \mathcal{L}_{\text{abs}} = \frac{1}{|\Omega_{\text{valid}}|} \sum_{p \in \Omega_{\text{valid}}} \hat{E}(p),  
  \end{equation}  
  
with valid regions \(\Omega_{\text{valid}}\) defined as pixels where both error map values are non-zero and both rendered and monocular depths are positive.  
  
The \textbf{correlation error term} \(\mathcal{L}_{\text{corr}}\) enforces consistency in invalid regions using min-max normalized depth maps:  

  \begin{equation}  
  \mathcal{L}_{\text{corr}} = \left| 1 - \frac{1}{|\Omega_{\text{invalid}}|} \sum_{p \in \Omega_{\text{invalid}}} \hat{D}_r(p) \hat{D}_m(p) \right|,  
  \end{equation}  
  
where \(\Omega_{\text{invalid}} = \{p \in \Omega | p \notin \Omega_{\text{valid}}\}\) is the complement of the valid region, and \(\hat{D}_r\) and \(\hat{D}_m\) are min-max normalized to \([0, 1]\) as follows:

\begin{equation}
\hat{D}(p) = \frac{D(p) - D_{\min}}{D_{\max} - D_{\min} + \epsilon},
\end{equation}

with \(\epsilon = 10^{-8}\) added for numerical stability.  

% \textbf{Advantages of the Proposed Method:}  
The proposed depth consistency loss addresses three critical challenges in depth alignment. First, the explicit min-max normalization of rendered and monocular depth maps (\(\hat{D}_r\) and \(\hat{D}_m\)) to the unit interval inherently handles scale variations between different regions, eliminating the need for ad-hoc calibration. Second, the dual-term design (\(\mathcal{L}_{\text{abs}}\) for valid regions and \(\mathcal{L}_{\text{corr}}\) for invalid regions) ensures robustness against uneven feature point distributions by decoupling the optimization constraints. Finally, the integration of normalized correlation in \(\mathcal{L}_{\text{corr}}\) and the previously computed error map \(\hat{E}(p)\) in \(\mathcal{L}_{\text{abs}}\) jointly improve numerical stability, effectively handling regions with insufficient valid matches while maintaining consistent depth relationships. These attributes are achieved with only a single hyperparameter \(\alpha\) to balance the two loss terms.

\subsection{Total Loss Formulation}
\label{subsec:total_loss}
The overall optimization objective is formulated as a weighted combination of four key components:

\begin{equation}
\mathcal{L}_{\text{total}} = \mathcal{L}_{\text{rgb}} + \mathcal{L}_{\text{dof}} + \lambda_{\text{geo}}\mathcal{L}_{\text{geo}} + \lambda_{\text{depth}}\mathcal{L}_{\text{depth}}
\end{equation}

where $\mathcal{L}_{\text{rgb}}$ ensures photometric accuracy through RGB reconstruction error minimization, $\mathcal{L}_{\text{dof}}$ enforces optical consistency by aligning depth-of-field effects, $\mathcal{L}_{\text{geo}}$ maintains multi-view geometric consistency across adjacent frames through feature matching, and $\mathcal{L}_{\text{depth}}$ aligns rendered depths with monocular depth priors using our proposed depth consistency loss.

The balancing weights $\lambda_{\text{geo}}$ and $\lambda_{\text{depth}}$ control the relative influence of geometric and depth constraints, respectively. This formulation enables joint optimization of appearance, geometry, and optical properties within a unified framework, effectively leveraging both learned monocular depth priors and multi-view geometric constraints.

\section{Experiment}  

\subsection{Datasets}  
\label{sssec:Datasets}  

Our experiments were conducted on four datasets: Waymo \citep{Sun_2020_CVPR}, Mip-NeRF360 \citep{barron2022mip}, SS3DM \citep{DBLP:conf/nips/0001WZG024}, and the YouTube \citep{DBLP:conf/icml/ChengLYY0MWC24} dataset. Waymo dataset is a large-scale urban dataset. Mip-NeRF360 is a common NeRF benchmark. The detailed descriptions and experimental results for both the SS3DM and YouTube datasets will be provided in the \revised{\ref{sssec:add_exp}}. 

\subsection{Implementation Details}
\label{sec:implementation}

Our implementation extends the original 3DGS framework \citep{kerbl20233d} by incorporating enhanced depth-of-field controls while maintaining compatibility for fair comparisons. All experiments were conducted using the original 3DGS configuration over 30,000 iterations on NVIDIA GeForce RTX 4090 GPUs. 

Key optical parameters adhere to standard photographic configurations: a 50mm focal length (equivalent to a standard lens), an $f/5.6$ aperture (providing a moderate depth of field), and a 36mm full-frame sensor. To balance defocus realism with rendering performance, we limited the maximum blur kernel size to $7\times7$ pixels.

When the Dynamic Focus Strategy is activated, the focus distance $d_f$ is optimized through statistical analysis of depth distributions. Empirical evaluation demonstrates that both median depth ($d_{\text{med}}$) and first-tercile depth ($d_{1/3}$) exhibit comparable performance metrics. The median depth configuration is adopted as the default selection to maximize computational efficiency.

Our depth-aware density control preserves the top 20\% of Gaussians by applying a $\tau=0.2$ quantile threshold to depth gradients, thereby prioritizing structural regions while maintaining computational efficiency.

\begin{table*}[t]
\caption{\revised{Revised comparison with state-of-the-art methods on Waymo and Mip-NeRF 360 datasets. Our method shows significant gains over the baseline through principled adaptation. \textbf{Ours (Default)} refers to our default configuration from the original manuscript. \textbf{Ours (All)} refers to the full framework with tuned supervision weights. \textbf{Ours (Best)} reports the per-scene best PSNR achieved across all experimental configurations, demonstrating the peak performance potential of our framework under this metric.}}
\label{tab:results}
\centering
\small
\setlength{\tabcolsep}{8pt}
\begin{tabular}{@{}l|*{3}{c}@{\hspace{8pt}}|*{3}{c}@{}}
\toprule
\multirow{2}{*}{Method} & \multicolumn{3}{c|}{Waymo} & \multicolumn{3}{c}{Mip-NeRF 360} \\
\cmidrule(lr){2-4} \cmidrule(lr){5-7}
& PSNR$\uparrow$ & SSIM$\uparrow$ & LPIPS$\downarrow$ & PSNR$\uparrow$ & SSIM$\uparrow$ & LPIPS$\downarrow$ \\ 
\midrule
Instant-NGP \citep{muller2022instant} & 30.98 & 0.886 & 0.281 & 25.59 & 0.699 & 0.331 \\
Mip-NeRF 360 \citep{barron2022mip} & 30.09 & 0.909 & 0.262 & 27.69 & 0.792 & 0.237 \\
Zip-NeRF \citep{barron2023zip} & 34.22 & 0.939 & 0.205 & \textbf{28.54} & \textbf{0.828} & \textbf{0.189} \\
3DGS \citep{kerbl20233d} & 34.04 & 0.942 & 0.224 & 27.21 & 0.815 & 0.214 \\
GaussianPro \citep{cheng2024gaussianpro} & 34.37 & 0.945 & 0.210 & 27.92 & 0.825 & 0.208 \\
\midrule
\revised{ConsistentGaussian (Ours, Default)} & \textbf{35.17} & \textbf{0.950} & \textbf{0.205} & 27.71 & 0.824 & 0.197 \\
\revised{ConsistentGaussian \textbf{(Ours, All)}} & \revised{-} & \revised{-} & \revised{-} & \revised{27.92} & \revised{0.827} & \revised{\textbf{0.189}} \\
\revised{ConsistentGaussian \textbf{(Ours, Best)}} & \revised{-} & \revised{-} & \revised{-} & \revised{27.95} & \revised{0.826} & \revised{0.195} \\
\bottomrule
\end{tabular}
\end{table*}

% 
% \begin{table*}[t]
% \caption{\revised{Revised comparison with state-of-the-art methods on Waymo and Mip-NeRF 360 datasets. Our method shows significant gains over the baseline through principled adaptation. \textbf{Ours (Default)} refers to our default configuration from the original manuscript. \textbf{Ours (All)} refers to the full framework with tuned supervision weights. \textbf{Ours (Best)} reports the per-scene best PSNR achieved across all experimental configurations, demonstrating the peak performance potential of our framework under this metric.}}
% \label{tab:results}
% \centering
% \small
% \setlength{\tabcolsep}{8pt}
% \begin{tabular}{@{}l|*{3}{c}@{\hspace{8pt}}|*{3}{c}@{}}
% \toprule
% \multirow{2}{*}{Method} & \multicolumn{3}{c|}{Waymo} & \multicolumn{3}{c}{Mip-NeRF 360} \\
% \cmidrule(lr){2-4} \cmidrule(lr){5-7}
% & PSNR$\uparrow$ & SSIM$\uparrow$ & LPIPS$\downarrow$ & PSNR$\uparrow$ & SSIM$\uparrow$ & LPIPS$\downarrow$ \\ 
% \midrule
% Instant-NGP \citep{muller2022instant} & 30.98 & 0.886 & 0.281 & 25.59 & 0.699 & 0.331 \\
% Mip-NeRF 360 \citep{barron2022mip} & 30.09 & 0.909 & 0.262 & 27.69 & 0.792 & 0.237 \\
% Zip-NeRF \citep{barron2023zip} & 34.22 & 0.939 & 0.205 & \textbf{28.54} & \textbf{0.828} & \textbf{0.189} \\
% 3DGS \citep{kerbl20233d} & 34.04 & 0.942 & 0.224 & 27.21 & 0.815 & 0.214 \\
% GaussianPro \citep{cheng2024gaussianpro} & 34.37 & 0.945 & 0.210 & 27.92 & 0.825 & 0.208 \\
% \midrule
% ConsistentGaussian (Ours) & \textbf{35.17} & \textbf{0.950} & \textbf{0.205} & 27.92 & 0.827 & \textbf{0.189} \\
% \bottomrule
% \end{tabular}
% \end{table*}

\subsection{Results}
\label{sec:Results}

% 增加名字后再改为0.85
\begin{figure*}[t] % 使用 [t] 将图像置于页面顶部
\centering
\includegraphics[width=\textwidth, height=0.8\textheight, keepaspectratio]{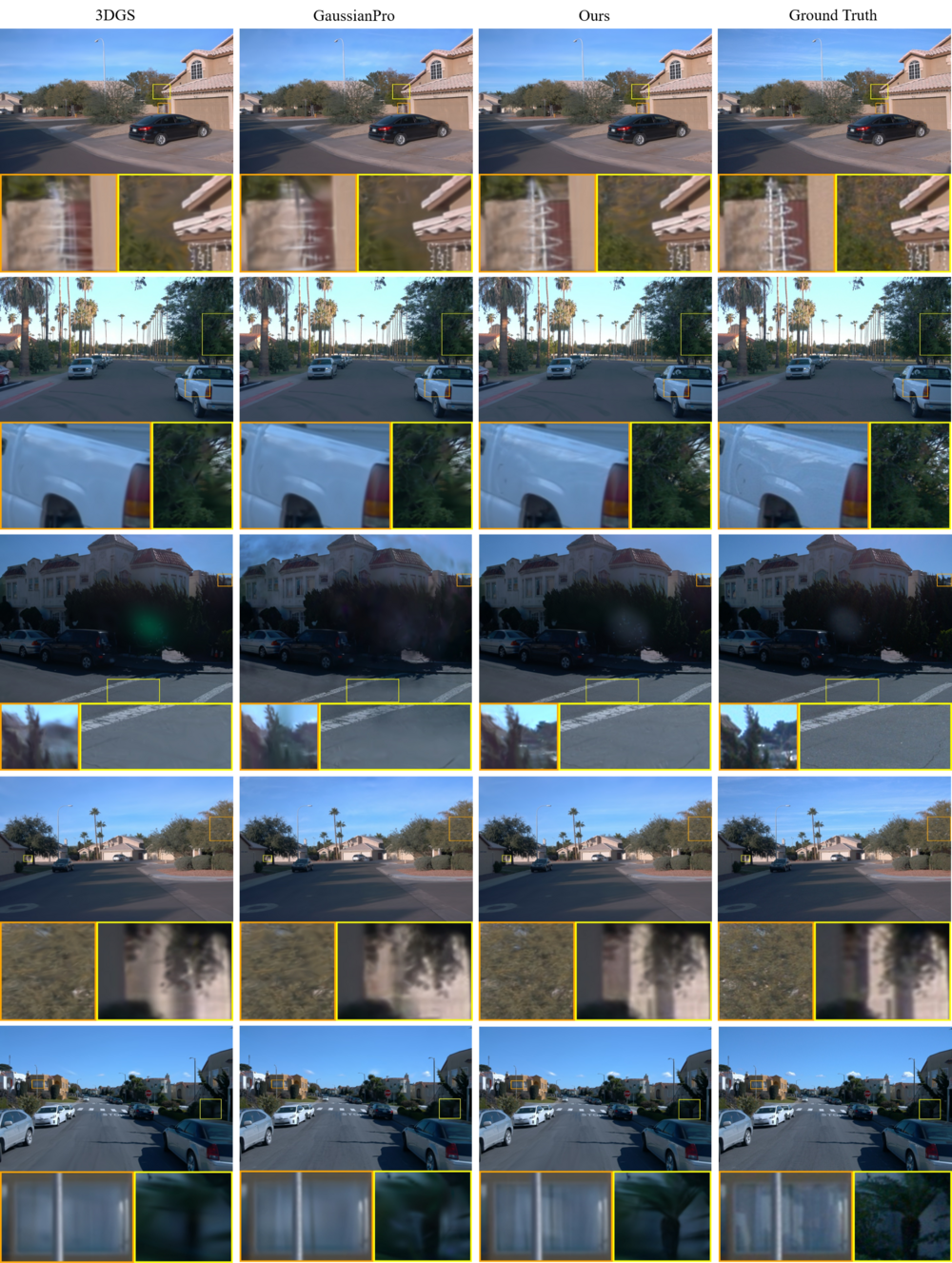}
\caption{Visual comparison on Waymo dataset. Our method achieves superior detail preservation in both mid-range and distant scenes compared to 3DGS and GaussianPro.}
\label{fig:waymo_compare}
\end{figure*}

To rigorously validate our approach, we established comparative benchmarks against five state-of-the-art novel view synthesis methods: \textbf{GaussianPro} \citep{cheng2024gaussianpro}, \textbf{Instant-NGP} \citep{muller2022instant}, \textbf{Mip-NeRF 360} \citep{barron2022mip}, \textbf{Zip-NeRF} \citep{barron2023zip}, and the baseline \textbf{3DGS} \citep{kerbl20233d}. The quantitative performance comparison between our method and these approaches is systematically presented in Table~\ref{tab:results}.

\subsubsection{Quantitative Comparisons}
\label{sssec:quant}

Table~\ref{tab:results} presents the comprehensive evaluation of our method against state-of-the-art approaches on Waymo and Mip-NeRF 360 datasets, analyzed through three key metrics: peak signal-to-noise ratio (PSNR), structural similarity index measure (SSIM) \citep{wang2004image}, and the learned perceptual image patch similarity (LPIPS) \citep{zhang2018unreasonable}. The experimental results for GaussianPro are sourced directly from its original implementation.

\revised{\textbf{Analysis on Unbounded Scenes and Adaptive Supervision.}} \revised{The Mip-NeRF 360 dataset, comprising nine distinct scenes (five outdoor and four indoor) that present diverse and challenging unbounded environments, serves as an ideal testbed for analyzing the adaptability of supervision strategies. Unlike the more structurally homogeneous forward-facing scenes in Waymo, its diversity (e.g., from the structured 'room' to the texture-rich 'treehill') motivates a deeper, scene-specific investigation into the interplay between our framework's components and scene characteristics. Our initial results with default parameters, while robust on Waymo, were less conclusive on this dataset, prompting this focused analysis.}

\revised{As shown in the revised Table~\ref{tab:results}, our full framework with tuned weights (\textbf{Ours, All}) improves the average PSNR to 27.92 dB. Furthermore, by selecting the configuration that yielded the best PSNR for each scene (\textbf{Ours, Best}), the average PSNR is further lifted to a peak of \textbf{27.95 dB}. While this does not surpass the specialized Zip-NeRF model, it highlights a key insight: for scenes with unreliable geometric priors due to repetitive textures (e.g., `treehill`), a targeted application of our supervision signals is most effective. Conversely, for well-structured scenes (e.g., `room`), the full suite of losses yields substantial improvements. This demonstrates that our framework's value lies not only in its strong baseline performance but also in its flexibility, providing a toolkit and principled guidelines for practitioners to achieve optimal results by adapting the supervision strategy to the scene at hand.}

\textbf{Cross-Dataset Analysis} The \revised{7.25} dB PSNR disparity between Waymo (35.17 dB) and Mip-NeRF 360 (\revised{27.92} dB) reflects fundamental challenges in outdoor driving scenarios with extensive depth ranges compared to indoor scenes with constrained depth variation. Our methodology specifically targets structural inconsistencies in distant views, which contrasts with the Mip-NeRF 360 dataset's characteristics where most sub-scenes exhibit limited depth variation. Furthermore, our analysis identifies implementation discrepancies between the Mip-NeRF 360 variants used in GaussianPro and our baseline experiments.

\subsubsection{Qualitative Comparisons}
\label{sssec:quali}

Our method demonstrates significant visual improvements in complex outdoor scenarios, as evidenced by comparative reconstructions on the Waymo dataset (Figure~\ref{fig:waymo_compare}). The key enhancement manifests in resolving the inherent near-far supervision conflict through depth-aware physical modeling, particularly evident in scenes containing both close-range structural elements and distant environmental features.

\textbf{Geometric Consistency Enhancement} As shown in Figure~\ref{fig:waymo_compare} (red boxes), our approach maintains coherent geometric patterns across various depth layers where baseline methods exhibit structural fragmentation or spatial artifacts. This improvement stems from two synergistic mechanisms: 1) The depth-sensitive density control prioritizes representation capacity allocation to geometrically critical regions, and 2) The hybrid depth estimation framework ensures metric consistency across multi-view observations through Equation~\ref{eq:final_depth}. These technical components jointly address the dual challenges of preserving near-field structural details while maintaining far-field depth coherence.

\textbf{Technical Implementation} All visualizations strictly adhere to the rendering configurations defined in Section~\ref{sec:implementation}, with kernel parameterization following the specifications in Section~\ref{sssec:kernel_design}. Depth-dependent blur synthesis is governed by the physical imaging model in Equation~\ref{eq:coc}.

\begin{table*}[t]
\caption{Comparison on SS3DM, YouTube, and LibraryDoF datasets.}
\label{tab:dataset_comparison}
\centering
\footnotesize
\setlength{\tabcolsep}{4pt}
\begin{tabular}{@{}l|*{3}{c}|*{3}{c}|*{3}{c}@{}}
\hline
\multirow{2}{*}{Method} 
& \multicolumn{3}{c|}{SS3DM} 
& \multicolumn{3}{c|}{YouTube} 
& \multicolumn{3}{c}{LibraryDoF} \\
& PSNR$\uparrow$ & SSIM$\uparrow$ & LPIPS$\downarrow$ 
& PSNR$\uparrow$ & SSIM$\uparrow$ & LPIPS$\downarrow$ 
& PSNR$\uparrow$ & SSIM$\uparrow$ & LPIPS$\downarrow$ \\ 
\hline
3DGS & 30.47 & 0.891 & 0.253 & 34.98 & 0.960 & 0.081 & 23.13 & 0.729 & 0.340 \\
GaussianPro & 31.44 & 0.906 & 0.225 & 35.58 & 0.9645 & 0.071 & 22.62 & 0.731 & 0.330 \\
Ours & \revised{\textbf{33.83}} & \revised{\textbf{0.928}} & \revised{\textbf{0.192}} & \textbf{36.50} & \textbf{0.972} & \textbf{0.057} & \textbf{24.81} & \textbf{0.810} & \textbf{0.241} \\
\hline
\end{tabular}
\vspace{-0.2cm}
\end{table*}

% % 
% \begin{table*}[t]
% \caption{Comparison on SS3DM, YouTube, and LibraryDoF datasets.}
% \label{tab:dataset_comparison}
% \centering
% \footnotesize
% \setlength{\tabcolsep}{4pt}
% \begin{tabular}{@{}l|*{3}{c}|*{3}{c}|*{3}{c}@{}}
% \hline
% \multirow{2}{*}{Method} 
% & \multicolumn{3}{c|}{SS3DM} 
% & \multicolumn{3}{c|}{YouTube} 
% & \multicolumn{3}{c}{LibraryDoF} \\
% & PSNR$\uparrow$ & SSIM$\uparrow$ & LPIPS$\downarrow$ 
% & PSNR$\uparrow$ & SSIM$\uparrow$ & LPIPS$\downarrow$ 
% & PSNR$\uparrow$ & SSIM$\uparrow$ & LPIPS$\downarrow$ \\ 
% \hline
% 3DGS & 30.47 & 0.891 & 0.253 & 34.98 & 0.960 & 0.081 & 23.13 & 0.729 & 0.340 \\
% GaussianPro & 31.44 & 0.906 & 0.225 & 35.58 & 0.9645 & 0.071 & 22.62 & 0.731 & 0.330 \\
% Ours & \textbf{33.83} & \textbf{0.928} & \textbf{0.192} & \textbf{36.50} & \textbf{0.972} & \textbf{0.057} & \textbf{24.81} & \textbf{0.810} & \textbf{0.241} \\
% \hline
% \end{tabular}
% \vspace{-0.2cm}
% \end{table*}

\begin{figure*}[h]
    \centering
    \begin{tabular}{cccc}
        % \toprule
        \multicolumn{1}{c}{3DGS} & \multicolumn{1}{c}{GaussianPro} & \multicolumn{1}{c}{Ours} & \multicolumn{1}{c}{Ground Truth} \\ % 空表头
        % \midrule
        \includegraphics[width=0.2\textwidth]{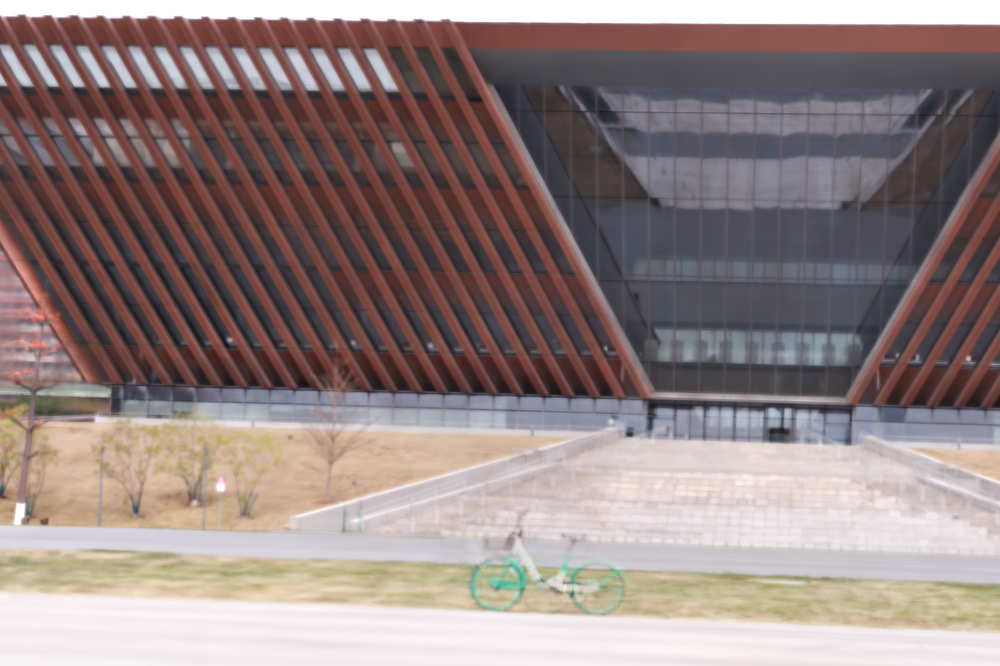} &
        \includegraphics[width=0.2\textwidth]{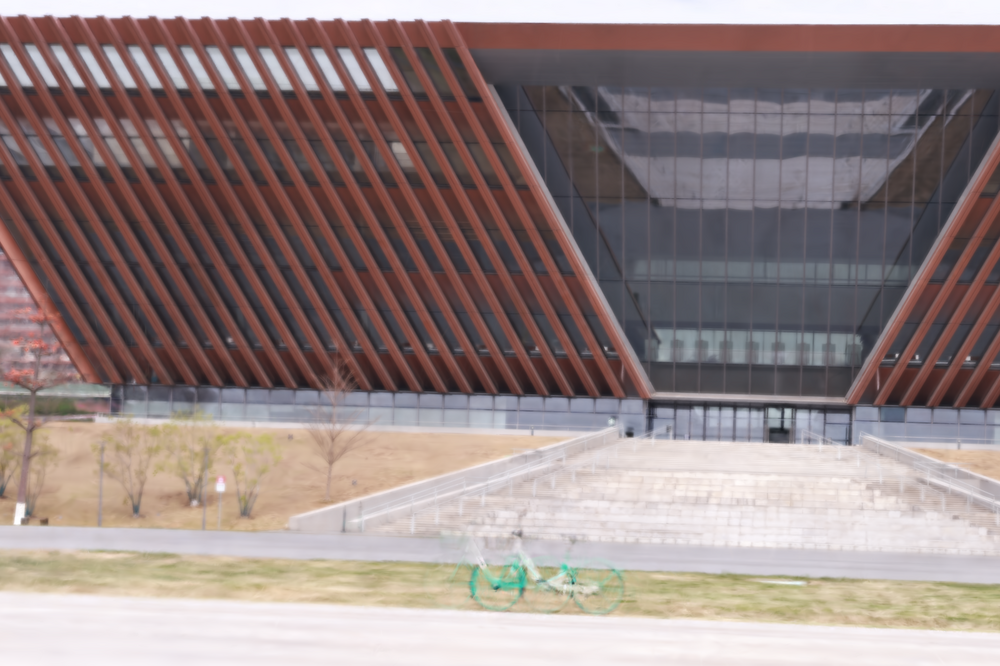} &
        \includegraphics[width=0.2\textwidth]{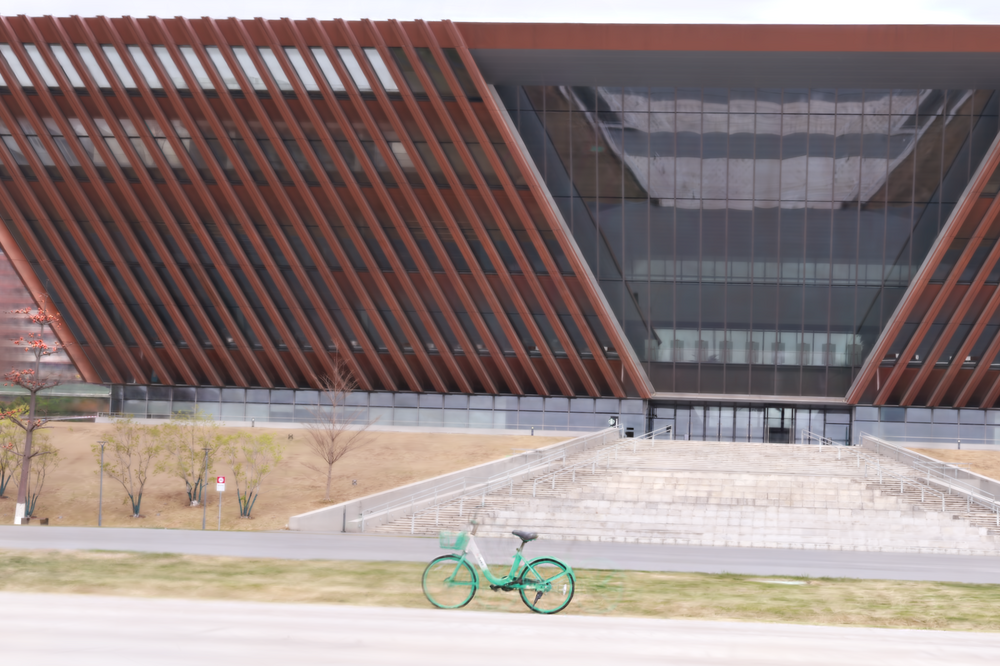} &
        \includegraphics[width=0.2\textwidth]{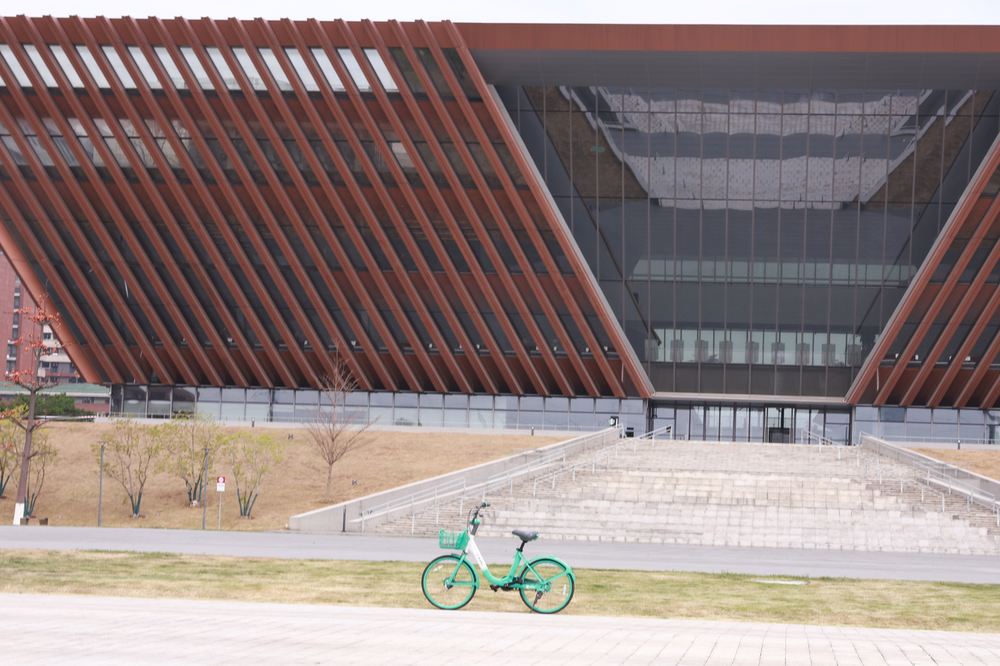} \\
        \includegraphics[width=0.2\textwidth]{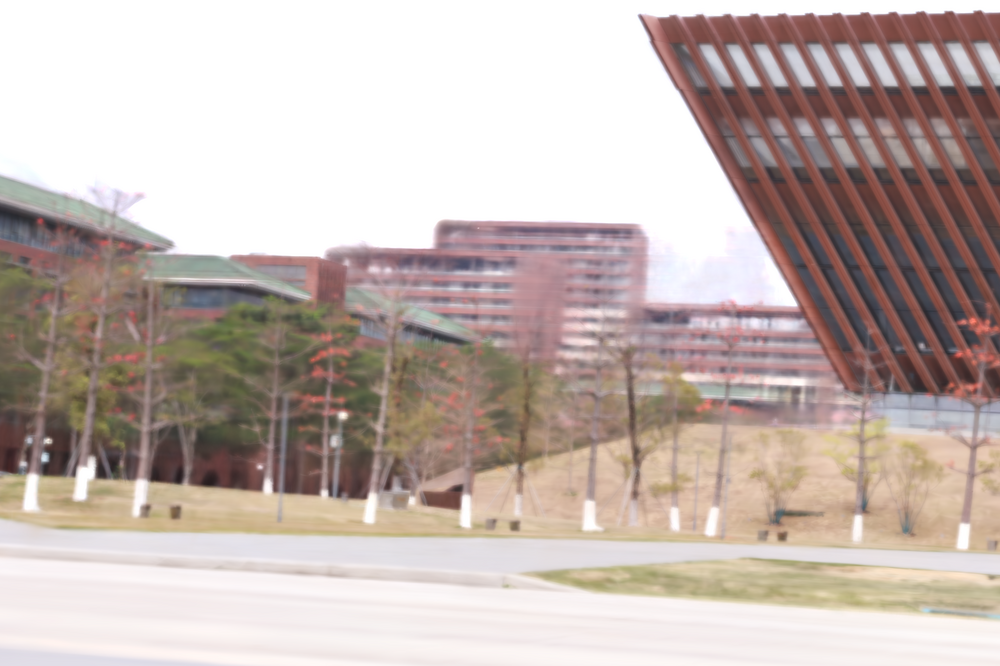} &
        \includegraphics[width=0.2\textwidth]{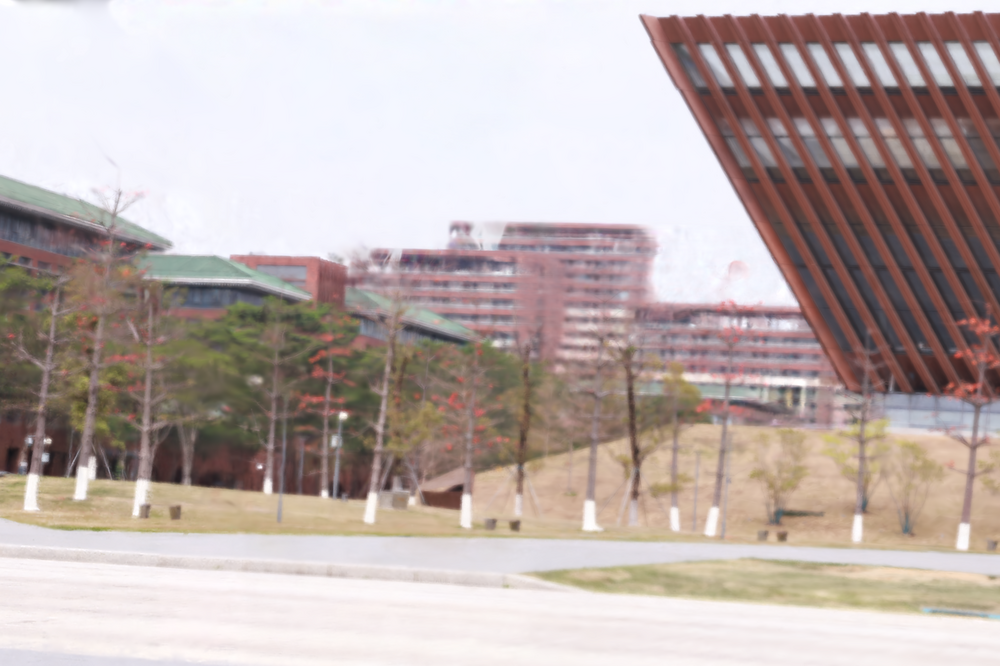} &
        \includegraphics[width=0.2\textwidth]{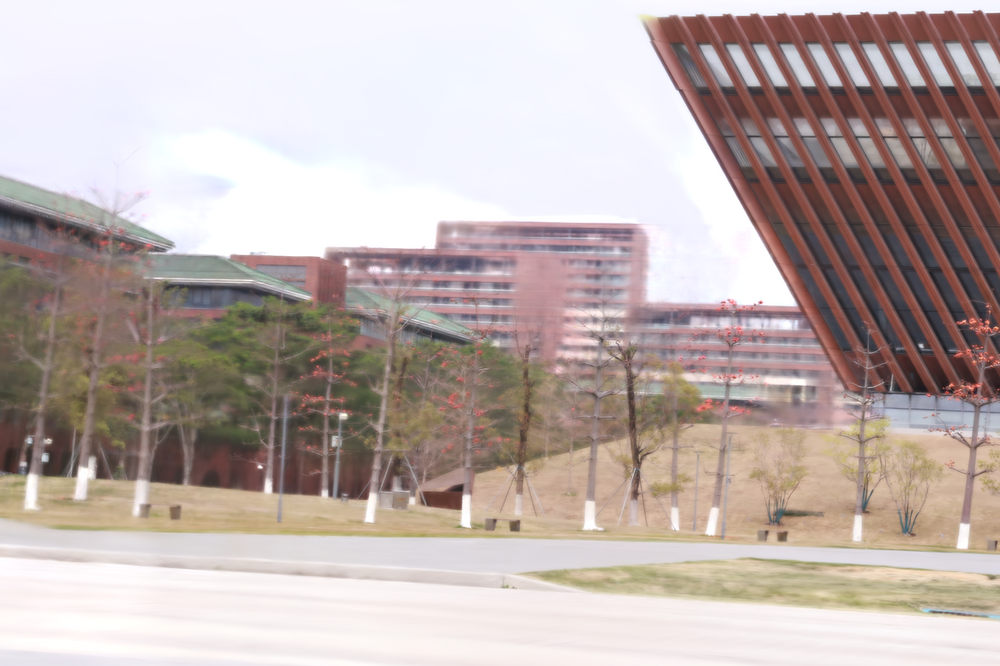} &
        \includegraphics[width=0.2\textwidth]{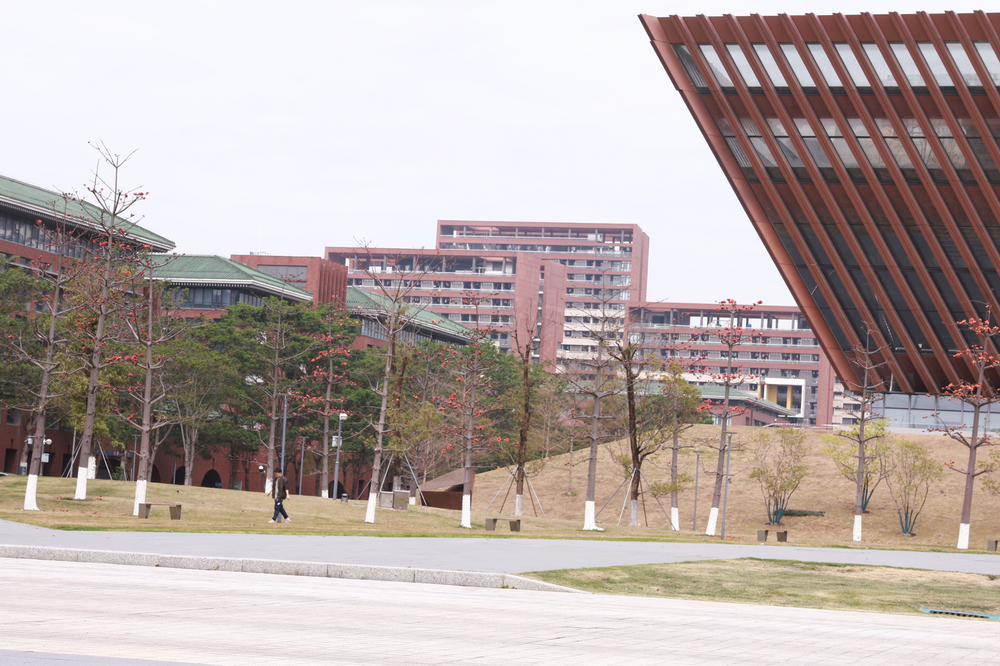} \\
        % \bottomrule
    \end{tabular}
    \caption{Comparison of model renderings on the LibraryDoF dataset among our method, GaussianPro , and 3DGS. Our method shows distinct advantages, with more accurate detail restoration, and a more immersive visual experience. }
    \label{Fig:self_collected_render}
\end{figure*}

\begin{table*}[t]
\caption{Per-scene performance comparison across SS3DM towns.}
\label{tab:ss3dm_results}
\centering
\footnotesize
\setlength{\tabcolsep}{2.5pt} % 压缩列间距
\begin{tabular}{l|*{3}{c}|*{3}{c}|*{3}{c}|*{3}{c}}
\hline
\multirow{2}{*}{Method} & \multicolumn{3}{c|}{Town01} & \multicolumn{3}{c|}{Town02} & \multicolumn{3}{c|}{Town03} & \multicolumn{3}{c}{Town04} \\
& PSNR$\uparrow$ & SSIM$\uparrow$ & LPIPS$\downarrow$ & PSNR$\uparrow$ & SSIM$\uparrow$ & LPIPS$\downarrow$ & PSNR$\uparrow$ & SSIM$\uparrow$ & LPIPS$\downarrow$ & PSNR$\uparrow$ & SSIM$\uparrow$ & LPIPS$\downarrow$ \\
\hline
3DGS & 32.466 & 0.934 & 0.220 & 25.549 & 0.844 & 0.342 & 26.356 & 0.840 & 0.325 & 31.976 & 0.910 & 0.205 \\
GaussianPro & 32.425 & 0.929 & 0.220 & 31.588 & 0.918 & 0.233 & 31.101 & 0.880 & 0.250 & 31.713 & 0.913 & 0.202 \\
Ours & \textbf{34.400} & \textbf{0.953} & \textbf{0.178} & \revised{\textbf{34.000}} & \revised{\textbf{0.947}} & \revised{\textbf{0.184}} & \revised{\textbf{35.903}} & \revised{\textbf{0.944}} & \revised{\textbf{0.168}} & \revised{\textbf{34.199}} & \revised{\textbf{ 0.934}} & \textbf{0.172} \\
\hline
\hline
\multirow{2}{*}{Method} & \multicolumn{3}{c|}{Town05} & \multicolumn{3}{c|}{Town06} & \multicolumn{3}{c|}{Town07} & \multicolumn{3}{c}{Town10} \\
& PSNR$\uparrow$ & SSIM$\uparrow$ & LPIPS$\downarrow$ & PSNR$\uparrow$ & SSIM$\uparrow$ & LPIPS$\downarrow$ & PSNR$\uparrow$ & SSIM$\uparrow$ & LPIPS$\downarrow$ & PSNR$\uparrow$ & SSIM$\uparrow$ & LPIPS$\downarrow$ \\
\hline
3DGS & 33.350 & 0.927 & 0.196 & 32.305 & 0.906 & 0.254 & 30.220 & 0.870 & 0.244 & 31.562 & 0.895 & 0.241 \\
GaussianPro & 31.959 & 0.918 & 0.204 & 32.210 & 0.908 & 0.246 & 29.372 & \textbf{0.886} & \textbf{0.213} & 31.145 & 0.895 & 0.232 \\
Ours & \revised{\textbf{34.432}} & \revised{\textbf{0.940}} & \revised{\textbf{0.174}} & \textbf{34.082} & \textbf{0.918} & \textbf{0.231} & \textbf{30.944} & 0.877 & 0.225 & \textbf{32.710} & \textbf{0.913} & \textbf{0.202} \\
\hline
\end{tabular}
\vspace{-0.2cm}
\end{table*}

\subsubsection{Additional experiments}
\label{sssec:add_exp}

We performed additional experiments on diverse datasets to validate the generalizability of the ConsistentGaussian model (ours). Beyond the widely adopted real-world Mip-Nerf 360 and Waymo datasets in the 3D reconstruction field, we specifically introduced three distinct data sources: the SS3DM synthetic autonomous driving dataset \citep{DBLP:conf/nips/0001WZG024}, YouTube video sequences \citep{cheng2024gaussianpro}, and our self-collected LibraryDoF dataset captured using consumer-grade cameras. Through experimental validation of these three types of differentiated datasets, we further corroborated the robustness advantages of our proposed method in handling multi-source heterogeneous data.

The SS3DM benchmark provides controlled-environment evaluation for street view surface reconstruction, offering CARLA-simulator-generated 3D ground-truth meshes that enable precise geometric and photometric assessments. The YouTube dataset, curated by GaussianPro, contains four distinct Subscenes extracted from publicly available YouTube videos.

The SS3DM dataset contains eight virtual towns (\texttt{Town01 - Town07, Town10}), each featuring multiple sub-scenes. We standardize our evaluation by selecting the \texttt{150\_streetsurf} sub-scene in \texttt{Town01 - Town07}, while employing \texttt{200\_streetsurf} in \texttt{Town10} which lacks the \texttt{150\_streetsurf} configuration. This strategy maintains the consistency of the evaluation while accommodating the inherent variations in the dataset.

For the YouTube dataset, each sub-scene corresponds to 360-degree aerial footage of iconic landmarks, such as the Eiffel Tower. We utilized the YouTube dataset to further validate the effectiveness of our proposed method.

To further evaluate the efficacy of our proposed method, we meticulously acquired a comprehensive dataset using a Canon R10 camera in front of the university library. The dataset comprises two distinct categories of images with approximately equal distribution and similar amounts: near-all-in-focus images and shallow depth-of-field images. The near-all-in-focus images were captured with the focus set at distant objects, simulating an all-in-focus effect, while the shallow depth-of-field images were obtained by shifting the focal plane forward under identical camera extrinsic parameters, thereby introducing defocus effects.

For model training implementation, we adhered to the splitting strategy used in the original 3DGS, with a crucial modification: we ensured that the training set contained a balanced mixture of both near-all-in-focus and shallow depth-of-field images, whereas the test set exclusively consisted of near-all-in-focus images. During the loss computation phase, the base reconstruction loss $\mathcal{L}_\text{rgb}$ is only calculated when the depth-of-field strategy is disabled or the loaded image is near all-in-focus. In contrast, for shallow depth-of-field images, we computed $\mathcal{L}_{\text{dof}}$. When training comparative baseline models, we restricted the input to near-all-in-focus images exclusively, as these approaches lack the capability to process images with pronounced depth-of-field effects.

The scene-specific evaluation results of the synthetic SS3DM dataset are presented in Table \ref{tab:ss3dm_results}, which provides a comprehensive overview of performance metrics across its distinct scenarios. Table \ref{tab:dataset_comparison} further summarizes the mean metric comparisons among SS3DM (synthetic autonomous driving data), YouTube sequences (real-world dynamic scenes), and our LibraryDoF dataset (challenging handheld captures). Figure \ref{Fig:self_collected_render} shows some rendering results from the LibraryDoF dataset. These cross-dataset comparisons demonstrate the consistent superiority of our method across diverse data modalities, highlighting its robust generalization capability from synthetic to real-world scenarios and across varying imaging conditions.

\subsection{Ablation Study}
\label{sec:ablation}

\begin{table*}[t]
\caption{Ablation study with component integration and blur kernel comparisons on Waymo dataset}
\label{tab:ablation}
\centering
\small
\setlength{\tabcolsep}{4pt}
\begin{tabular}{l|ccc|ccc}
\hline
Configuration & PSNR$\uparrow$ & SSIM$\uparrow$ & LPIPS$\downarrow$ & $\Delta$PSNR & $\Delta$SSIM & $\Delta$LPIPS \\
\hline
3DGS (Baseline) & 34.04 & 0.942 & 0.224 & - & - & - \\
\hline
\multicolumn{7}{l}{\textit{Progressive Component Integration}} \\
+ Depth-of-field (Gaussian Blur Kernel) & 34.61 & 0.946 & 0.217 & +0.57 & +0.004 & -0.007 \\
+ Gradient Control & 35.00 & 0.948 & 0.209 & +0.96 & +0.006 & -0.015 \\
+ Point Match Loss & 34.98 & 0.949 & 0.206 & +0.94 & +0.007 & -0.018 \\
\textbf{Full (Gaussian Blur Kernel)} & \textbf{35.17} & \textbf{0.950} & \textbf{0.205} & +1.13 & +0.008 & -0.019 \\
\hline
\multicolumn{7}{l}{\textit{Blur Kernel Analysis}} \\
Full (Polygonal Blur Kernel) & 35.03 & 0.948 & 0.209 & +0.99 & +0.006 & -0.015 \\
Full (SmoothStep Blur Kernel) & 35.03 & 0.948 & 0.209 & +0.99 & +0.006 & -0.015 \\
No Depth-of-field Loss & 34.42 & 0.944 & 0.221 & +0.38 & +0.002 & -0.003 \\
\hline
\end{tabular}
\end{table*}

\subsubsection{Experimental Protocol}
\label{sssec:ablation_protocol}

Our ablation analysis adopts two complementary strategies: 1) Progressive integration of core components to isolate individual contributions, and 2) Comparative evaluation of blur kernel implementations. The first protocol sequentially adds depth-of-field supervision, gradient-aware density control, point matching loss, and monocular depth alignment to the 3DGS baseline. The second protocol substitutes the Gaussian blur kernel with Polygonal blur kernel or SmoothStep blur kernel in the full configuration. The "No Depth-of-field Loss" condition removes defocus supervision while retaining other components, quantifying its geometric regularization effect.

\subsubsection{Component Effectiveness}
\label{sssec:ablation_components}

The Gaussian blur-based depth-of-field supervision establishes the foundational improvement (+0.57 dB PSNR), validating our physics-driven defocus modeling. Subsequent integration of gradient-aware density control contributes an additional +0.39 dB enhancement, demonstrating its efficacy in preserving geometrically critical regions. The marginal LPIPS improvement (0.206 vs 0.209) when adding point matching loss indicates enhanced structural consistency through multi-view correspondence constraints.

The full configuration achieves peak performance (35.17 dB PSNR), confirming the necessity of unified depth supervision. The 0.75 dB degradation in the "No Depth-of-field Loss" condition (34.42 dB vs 35.17 dB) quantitatively demonstrates defocus supervision's critical role in geometric regularization, consistent with our theoretical analysis in Section~\ref{sssec:physical_model}.

\subsubsection{Blur Kernel Analysis}
\label{sssec:ablation_kernel}

All blur kernel implementations significantly outperform the baseline, with the Gaussian configuration achieving optimal PSNR (35.17 dB) and LPIPS (0.205). The Polygonal blur kernel generates physically accurate bokeh effects through parametric aperture modeling (Figure~\ref{fig:Polygonal_kernel}), effectively replicating real camera optics. \revised{The Polygonal and SmoothStep blur kernel implementations yield identical performance}.

\textbf{Implementation Guidelines} The Gaussian blur kernel demonstrates superior overall performance across metrics, making it the default choice for general applications. The Polygonal blur kernel is recommended for scenarios requiring optical realism, particularly when synthesizing aperture-specific effects. The SmoothStep blur kernel provides enhanced edge preservation for high-frequency detail recovery and is suitable for post-processing applications. This parametric design framework maintains core reconstruction performance while accommodating diverse optical requirements.

\subsubsection{Hyperparameter Robustness}
\label{sssec:ablation_hyper}

\revised{To address the practical usability of our method, we analyze the sensitivity of the key new hyperparameter introduced: the depth-gradient preservation quantile, $\tau$ (see Section~\ref{subsec:dof_density_control}). We evaluate the final reconstruction quality (PSNR) on three diverse scenes from the Mip-NeRF 360 dataset while varying $\tau$ over a wide range from 0.1 (preserving top 90\% of gradients) to 0.9 (preserving top 10\%). As shown in Table~\ref{tab:robustness_tau}, the performance is highly stable across all scenes. The maximum PSNR deviation is less than 0.15 dB, demonstrating that our method is not sensitive to the precise choice of this parameter and that our default value ($\tau=0.2$) is a robust choice for general use.}

\revised{In summary, our framework is designed to be both principled and practical, minimizing the need for extensive hyperparameter tuning and ensuring its reproducibility.}

\begin{table*}[t]
\caption{\revised{Robustness analysis of the depth-gradient preservation quantile ($\tau$) on the Mip-NeRF 360 dataset. PSNR$\uparrow$ is stable across a wide range of $\tau$ values, indicating low sensitivity. The maximum performance deviation for each scene is shown in parentheses, confirming either "Highly Robust" ($\leq$0.1 dB) or "Good Robustness" (0.1-0.2 dB) according to our protocol. Our default value is $\tau=0.2$.}}
\label{tab:robustness_tau}
\centering
\small
\begin{tabular}{@{}l|ccccccc@{}}
\toprule
\multirow{2}{*}{Scene} & \multicolumn{7}{c}{Preservation Quantile $\tau$} \\
\cmidrule(l){2-8}
 & 0.1 & 0.2 & 0.4 & 0.5 & 0.6 & 0.8 & 0.9 \\ 
\midrule
bicycle ($\Delta$0.13 dB) & 25.31 & 25.35 & 25.39 & 25.37 & 25.35 & 25.33 & 25.26 \\
garden ($\Delta$0.10 dB)  & 27.50 & 27.58 & 27.53 & 27.56 & 27.55 & 27.58 & 27.60 \\
room ($\Delta$0.14 dB)    & 31.99 & 31.96 & 32.09 & 32.03 & 32.08 & 31.98 & 31.95 \\
\bottomrule
\end{tabular}
\end{table*}

\section{Conclusion}
\label{sec:conclusion}

We propose a physics-guided framework that enhances 3D Gaussian Splatting through depth-of-field-induced geometric supervision, addressing three fundamental challenges in neural scene reconstruction. First, our differentiable defocus convolution model physically emulates camera optics through parametric kernel design, achieving optically faithful bokeh effects while preserving computational efficiency through separable convolution operators. Second, our gradient-aware density control mechanism dynamically preserves geometrically critical structures through quantile-based pruning, particularly effective in maintaining urban scene integrity. Third, hierarchical depth alignment integrates global monocular depth estimation with local grid-based corrections, significantly enhancing geometric consistency. Comprehensive evaluations demonstrate state-of-the-art performance in structured environments, showing enhanced rendering fidelity and improved depth estimation accuracy. 

Current limitations in unbounded scene optimization suggest three research directions: 1) Temporal focus adaptation mechanisms for video-consistent dynamic scene reconstruction, 2) depth-discontinuous kernel blending strategies combining physically accurate polygonal blur kernel with detail-preserving SmoothStep blur kernel, and 3) adaptive optics simulations for extreme depth ranges using dynamic kernel scaling.

\appendix
\section{\revised{Appendix}}
\label{sec:appendix}

\subsection{\revised{Analysis of Densification Strategy}}
\label{sec:appendix_densify}

% \revised{To validate our adaptive preservation mechanism for density control (see Section~\ref{subsec:dof_density_control}), we compared it against the simpler alternative of globally tuning the densification gradient threshold. As shown in Table~\ref{tab:densify_comparison}, simply lowering the threshold from the 3DGS default of 0.0002 is a fragile strategy: a reduction to 0.00015 or below resulted in training failure on the Mip-NeRF 360 `bicycle` scene. While carefully increasing the threshold can yield marginal gains, this requires scene-specific tuning. In contrast, our method, which retains the default threshold, not only remains stable but also improves reconstruction quality, demonstrating its superior robustness and efficacy.}

\revised{To validate our adaptive preservation mechanism (see Section~\ref{subsec:dof_density_control}), we compared it against the simpler alternative of globally tuning the densification gradient threshold. As shown in Table~\ref{tab:densify_comparison}, this manual tuning is a fragile process. Lowering the threshold from the 3DGS default of 0.0002 leads to training failure. In contrast, our adaptive method, which operates on the same stable default threshold, not only avoids this instability but also delivers a consistent performance gain over the baseline. While it is possible to achieve marginal gains by carefully tuning the threshold, this requires a delicate, scene-specific process. Our method provides a robust path to high performance without such manual intervention.}

% & 0.0002 (default) & 25.30 \\
% {Baseline (Tuning Global Threshold)}
\begin{table*}[h]
\caption{\revised{Quantitative comparison of densification strategies on the Mip-NeRF 360 `bicycle` scene. Our method is compared against baselines with our adaptive mechanism disabled.}}
\label{tab:densify_comparison}
\centering
\small
\begin{tabular}{@{}lcc@{}}
\toprule
Method & densify grad threshold & PSNR$\uparrow$ \\ 
\midrule
\textbf{Ours (Adaptive)} & \textbf{0.0002} & \textbf{25.35} \\
\hline
Baseline (Default Threshold) & 0.0002 & 25.26 \\
Baseline (Lowered Threshold) & $\leq$ 0.00015 & \textit{Training Failed} \\
\bottomrule
\end{tabular}
\end{table*}

\subsection{\revised{Robustness of Empirical Parameters}}
\label{sec:appendix_robustness}

\revised{Our framework introduces several parameters that are empirically set. Here, we provide a detailed analysis to demonstrate their robustness and validate our default choices.}

\subsubsection{\revised{Geometric Supervision Weights}}

\revised{The weights for the geometric consistency loss ($\lambda_{\text{geo}}$) and depth consistency loss ($\lambda_{\text{depth}}$) are set to 0.05 and 0.005 by default. As shown in Table~\ref{tab:robustness_lambda}, these defaults are robust for standard scenes like `bicycle`, where performance varies by only 0.04 dB across a 100x change in weights. For pathological scenes with unreliable geometric priors like `bonsai`, our analysis confirms the effectiveness of a key adaptive criterion: strategically reducing the weights leads to significant performance gains. This provides a clear, principled guideline for tuning in such specific cases, rather than requiring blind experimentation.}

\begin{table*}[h]
\caption{\revised{Robustness of geometric supervision weights ($\lambda$) on Mip-NeRF 360 scenes. PSNR$\uparrow$ is stable for standard scenes, while pathological scenes benefit from principled attenuation.}}
\label{tab:robustness_lambda}
\centering
\small
\begin{tabular}{@{}lccc@{}}
\toprule
Scene & Default Weights (1x) & Attenuated (10x) & Attenuated (100x) \\ 
\midrule
bicycle & 25.35 & 25.31 & \textbf{25.35} \\
bonsai & 32.44 & 32.64 & \textbf{32.80} \\
\bottomrule
\end{tabular}
\end{table*}

\subsubsection{\revised{Adaptive Grid Strategy Parameters}}

\revised{Our adaptive grid strategy constrains the cell size to an empirically established range of $[g_{\min}, g_{\max}] = [15, 60]$ pixels. The rationale for this effective range is rooted in the bias-variance tradeoff. We validate this choice with two analyses. First, Table~\ref{tab:robustness_grid} shows that the optimal fixed grid size is highly scene-dependent, making manual tuning impractical. Our adaptive method consistently performs competitively against the best fixed grid in each case, automating this choice. Second, Table~\ref{tab:robustness_grid_bounds} shows the method's low sensitivity to the precise values of $g_{\min}$ and $g_{\max}$, confirming our default range is a robust choice.}

\begin{table}[h]
\caption{\revised{Comparison of our adaptive grid strategy vs. a range of fixed grid sizes on Mip-NeRF 360 scenes. The optimal fixed grid size (bolded) is highly scene-dependent. Our adaptive method consistently achieves near-optimal performance automatically. Performance is measured in PSNR$\uparrow$.}}
\label{tab:robustness_grid}
\centering
\small
\begin{tabular}{@{}lccc@{}}
\toprule
Grid Strategy & bicycle & garden & room \\ 
\midrule
\textbf{Ours (Adaptive)} & 25.35 & \textbf{27.59} & 32.09 \\
\hline
8x8 & 25.33 & 27.58 & \textbf{32.10} \\
16x16 & 25.34 & 27.50 & 32.01 \\
32x32 & 25.37 & 27.57 & 32.08 \\
64x64 & \textbf{25.38} & 27.53 & 31.99 \\
128x128 & 25.34 & 27.54 & 31.97 \\
\bottomrule
\end{tabular}
\end{table}

\begin{table}[h]
\caption{\revised{Sensitivity analysis of the adaptive grid boundaries $[g_{\min}, g_{\max}]$ on the `bicycle` scene. PSNR$\uparrow$ is highly stable around our default setting of [15, 60].}}
\label{tab:robustness_grid_bounds}
\centering
\small
\begin{tabular}{@{}lc@{}}
\toprule
$[g_{\min}, g_{\max}]$ & PSNR$\uparrow$ \\ 
\midrule
$[5, 50]$ & 25.26 \\
$[5, 70]$ & 25.32 \\
$[15, 50]$ & 25.37 \\
$[15, 60]$ & 25.35 \\
$[25, 50]$ & 25.37 \\
$[25, 70]$ & 25.30 \\
\bottomrule
\end{tabular}
\end{table}

\begin{table}[h]
\caption{\revised{Sensitivity analysis of maximum blur kernel size on four diverse Mip-NeRF 360 scenes. A smaller 3x3 kernel consistently provides either superior or statistically equivalent performance. Performance is measured in PSNR$\uparrow$.}}
\label{tab:robustness_kernel_size}
\centering
\small
\begin{tabular}{@{}lcc@{}}
\toprule
Scene & Kernel Size & PSNR$\uparrow$ \\ 
\midrule
\multirow{2}{*}{bicycle} & \textbf{3x3} & \textbf{25.38} \\
& 7x7 & 25.35 \\
\hline
\multirow{2}{*}{flowers} & 3x3 & 22.01 \\
& \textbf{7x7} & \textbf{22.02} \\
\hline
\multirow{2}{*}{garden} & \textbf{3x3} & \textbf{27.55} \\
& 7x7 & 27.53 \\
\hline
\multirow{2}{*}{room} & \textbf{3x3} & \textbf{32.10} \\
& 7x7 & 31.87 \\
\bottomrule
\end{tabular}
\end{table}

\begin{figure*}[t]
    \centering
    
    % Define a single, consistent width for all image columns.
    % 0.24 * 4 = 0.96, leaving a small amount of space for gaps.
    \newcommand{\myimagewidth}{0.23\textwidth}
    
    % Define the horizontal gap between columns.
    \newcommand{\mycolumngap}{2mm}
    
    % Define the vertical gap between rows.
    \newcommand{\myrowgap}{2mm}

    % The >{\centering\arraybackslash} command ensures every cell's content is centered.
    % The @{\hspace{\mycolumngap}} command places a precise, fixed space between columns.
    \begin{tabular}{
        >{\centering\arraybackslash}m{\myimagewidth}
        @{\hspace{\mycolumngap}}
        >{\centering\arraybackslash}m{\myimagewidth}
        @{\hspace{\mycolumngap}}
        >{\centering\arraybackslash}m{\myimagewidth}
        @{\hspace{\mycolumngap}}
        >{\centering\arraybackslash}m{\myimagewidth}
    }
        % === COLUMN HEADERS ===
        \small 3DGS & \small GaussianPro & \small Ours & \small Ground Truth  \\[2mm]

        % === ROW 1: FLOWERS ===
        \includegraphics[width=\linewidth]{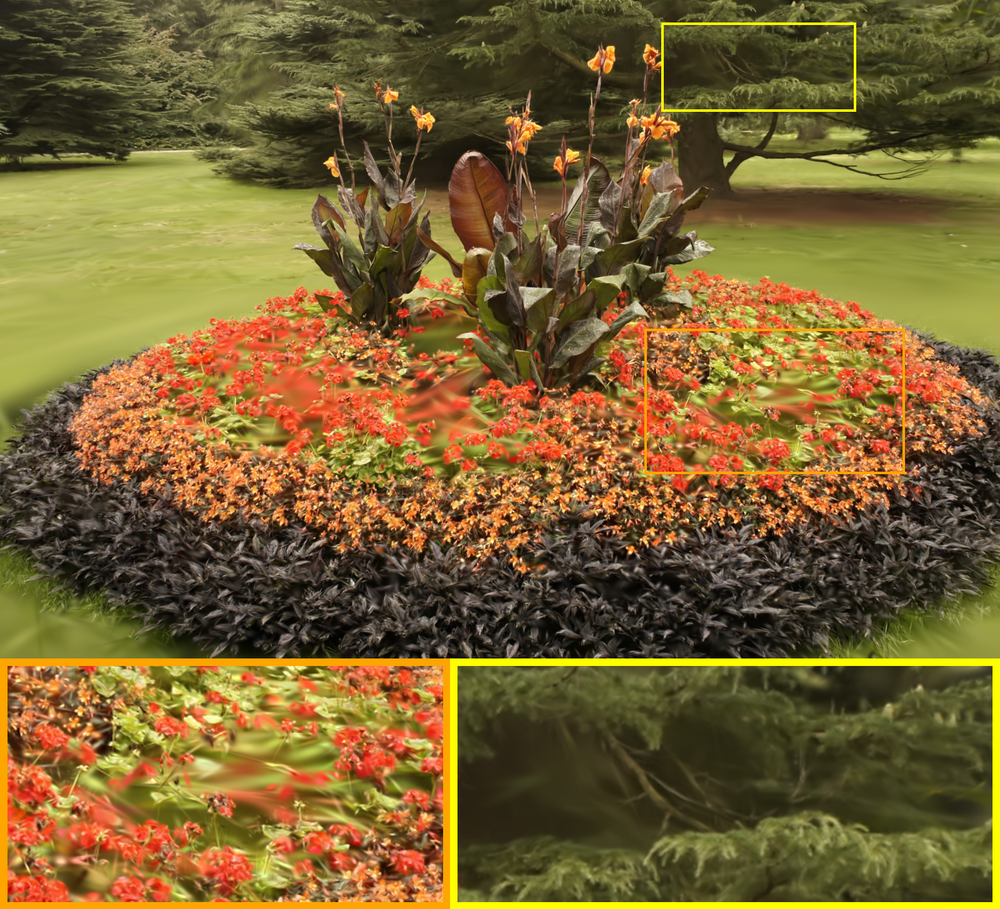} &
        \includegraphics[width=\linewidth]{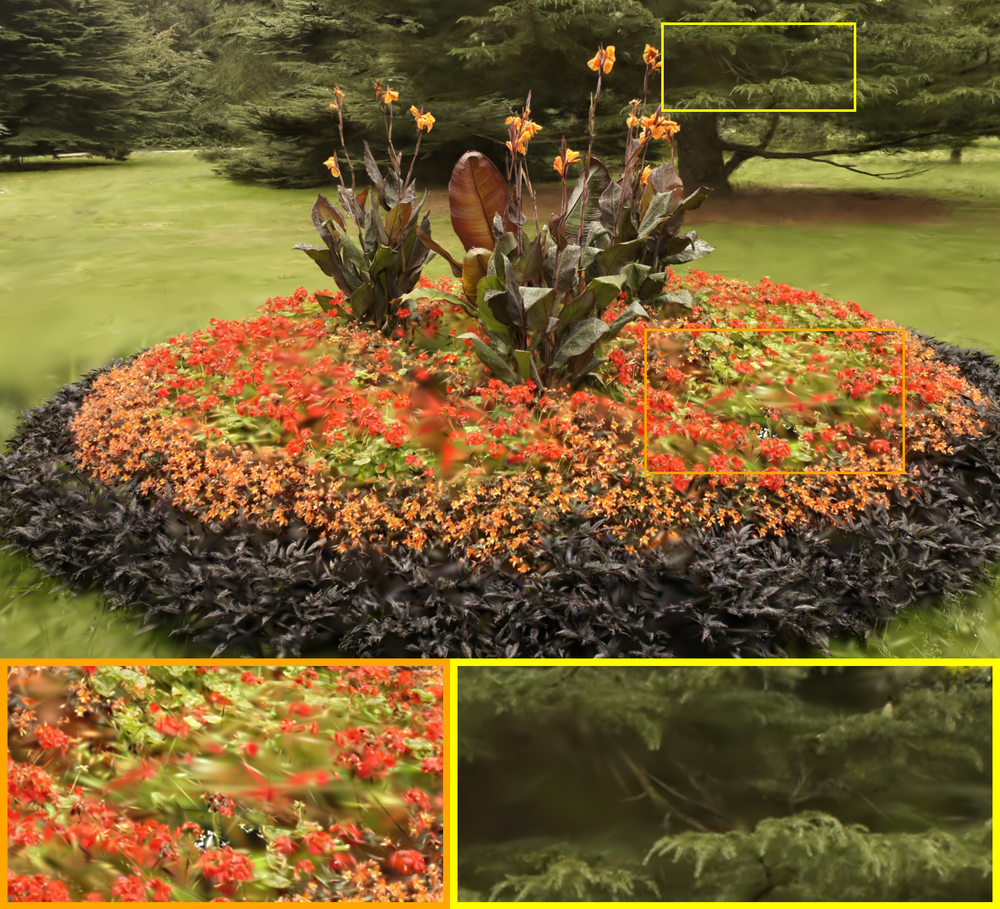} &
        \includegraphics[width=\linewidth]{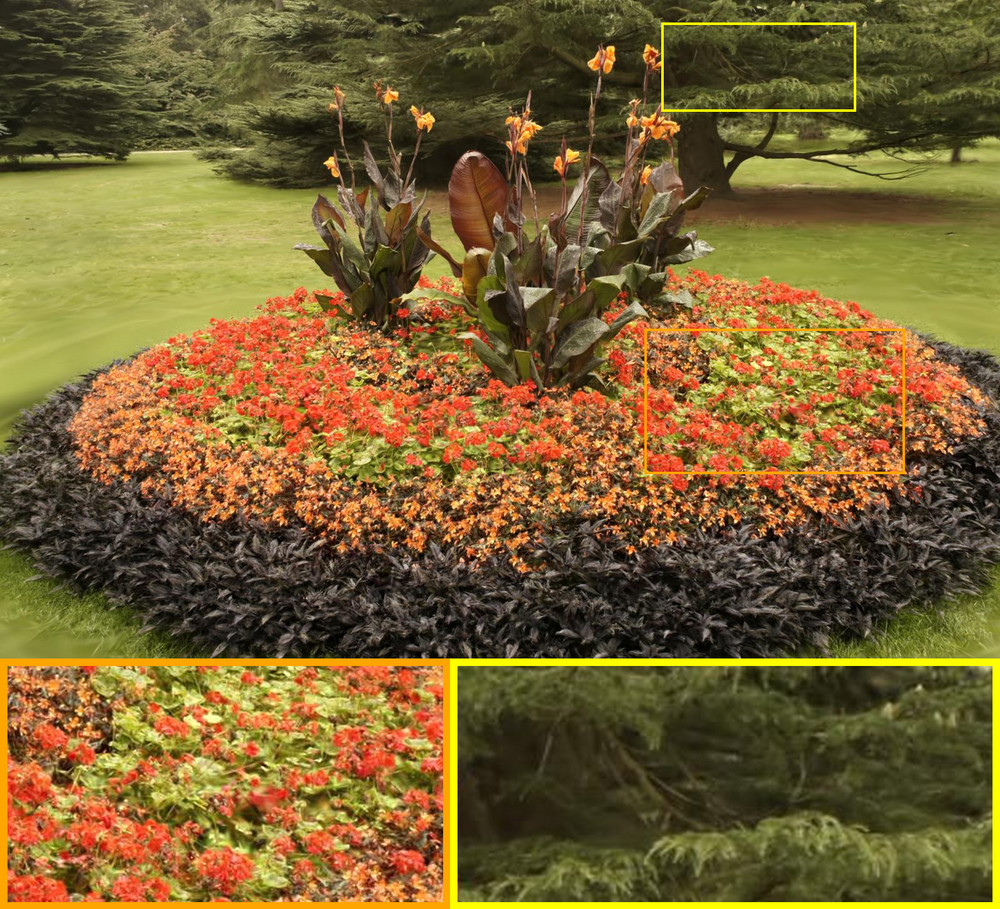} &
        \includegraphics[width=\linewidth]{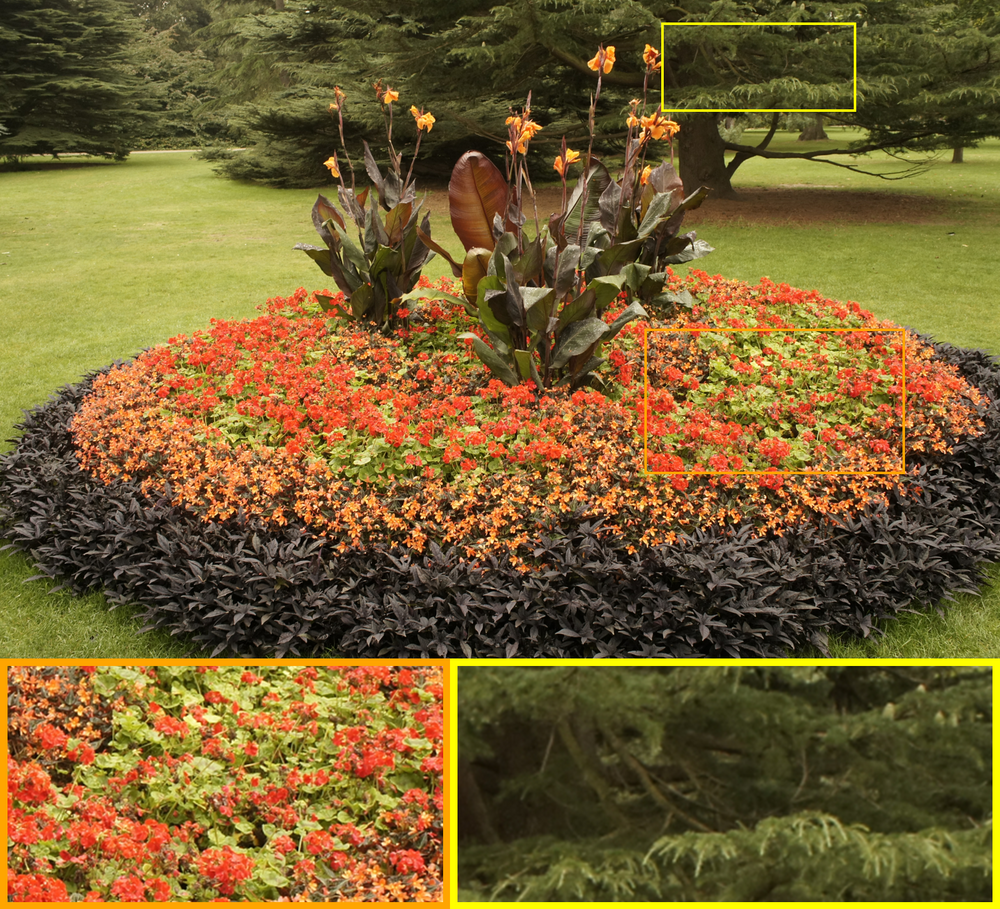} \\[\myrowgap]

        % === ROW 2: ROOM ===
        \includegraphics[width=\linewidth]
        {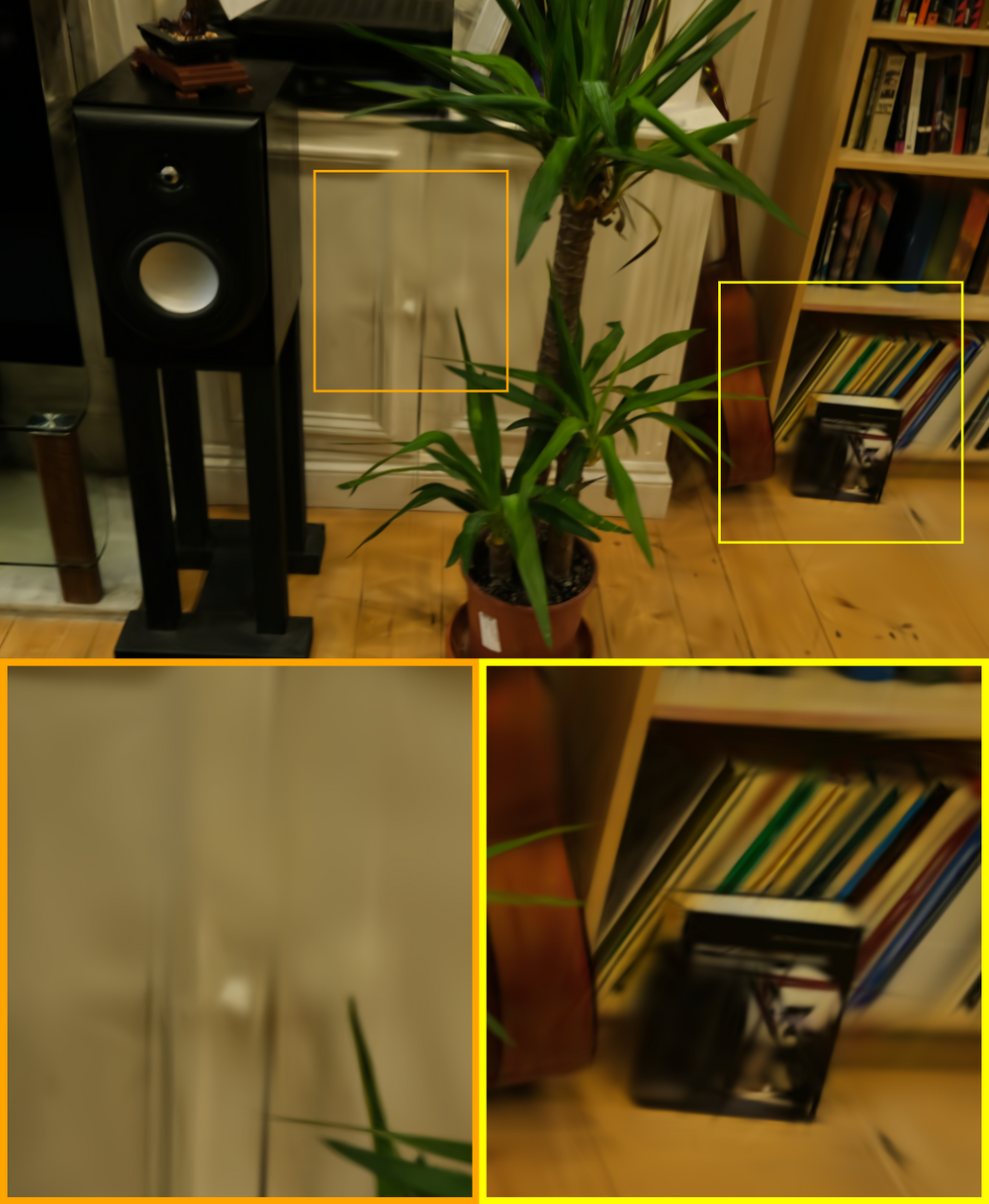} &
        \includegraphics[width=\linewidth]{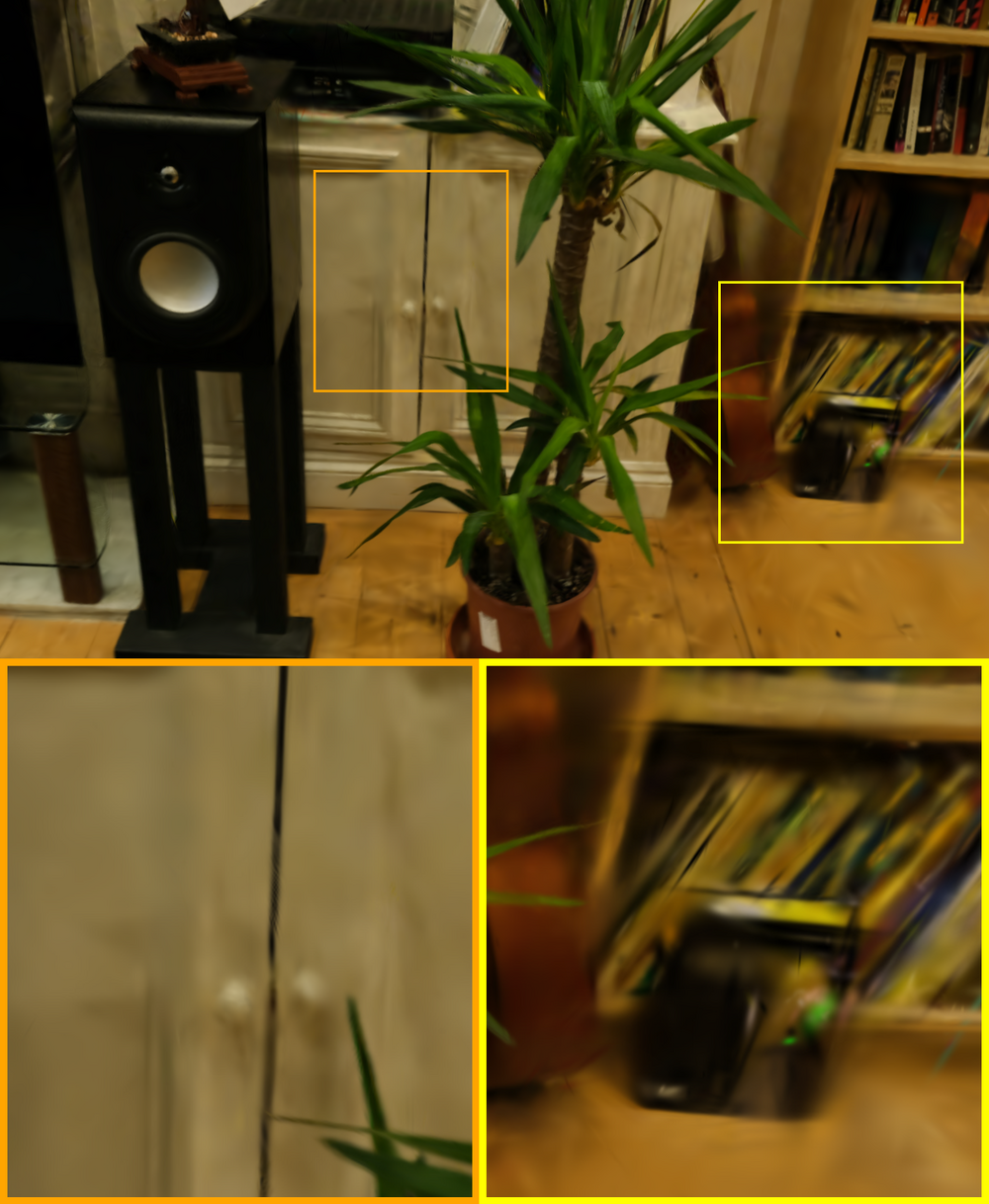} &
        \includegraphics[width=\linewidth]{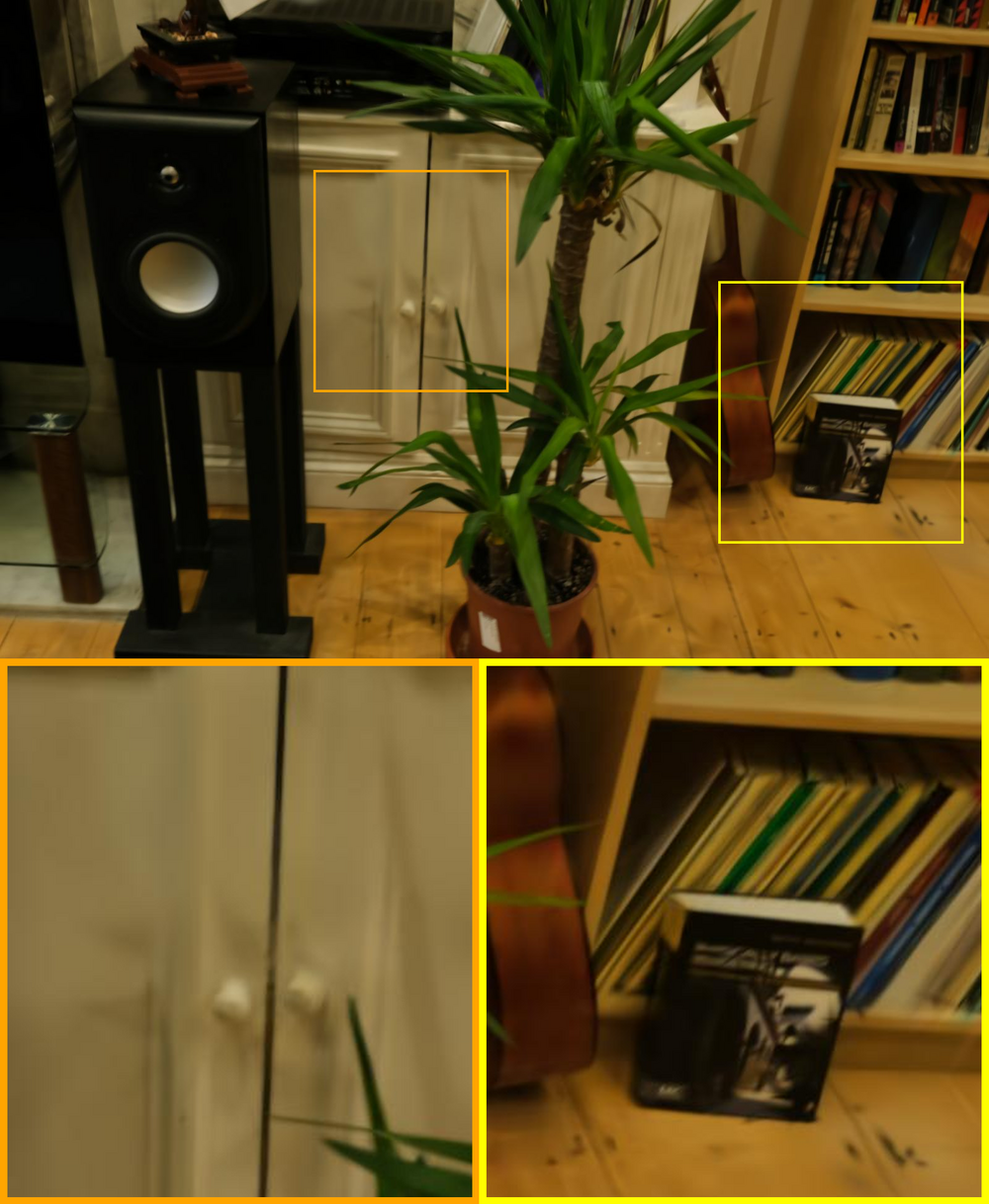} &
        \includegraphics[width=\linewidth]
        {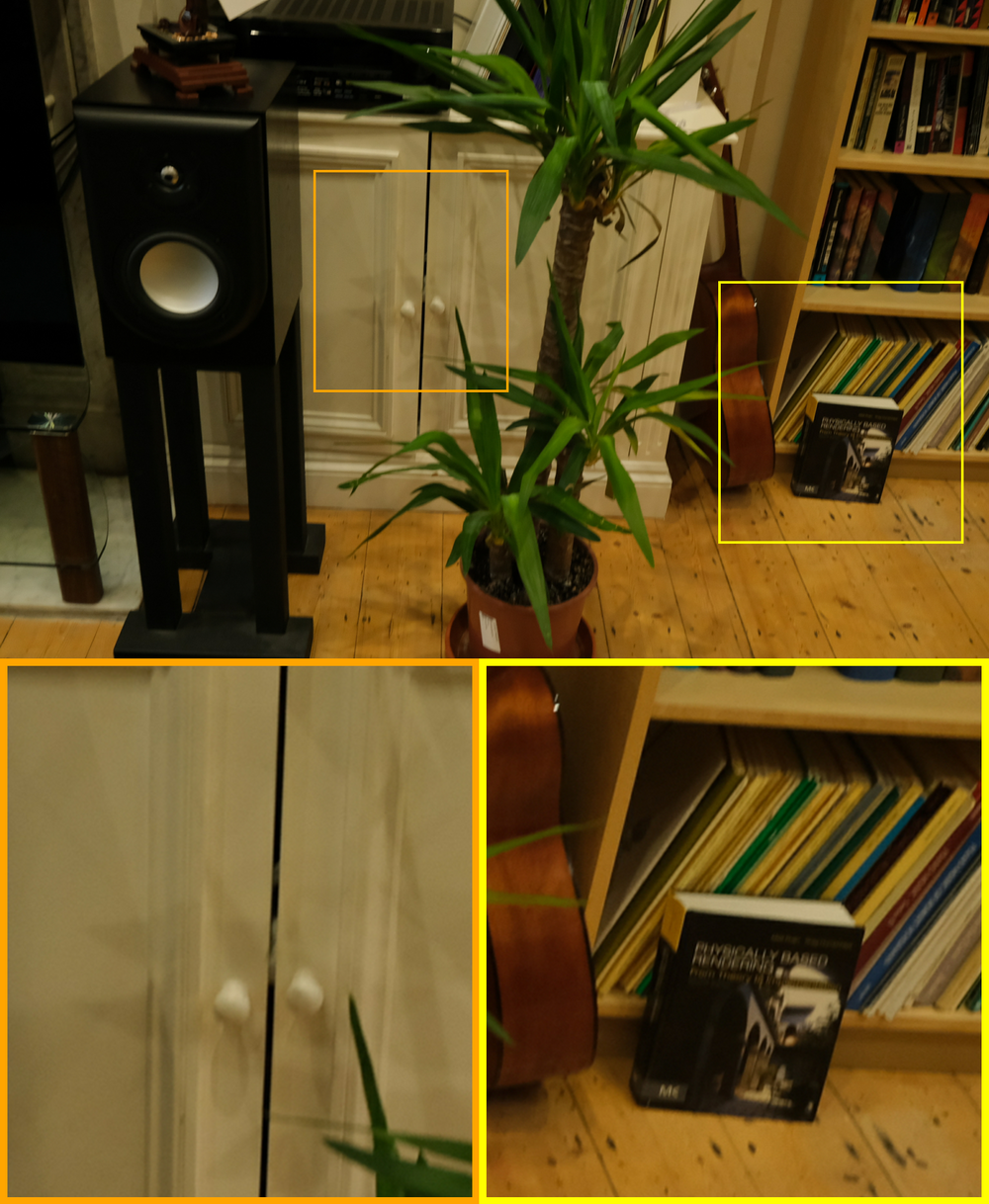} \\[\myrowgap]

        % === ROW 3: GARDEN ===
        \includegraphics[width=\linewidth]
        {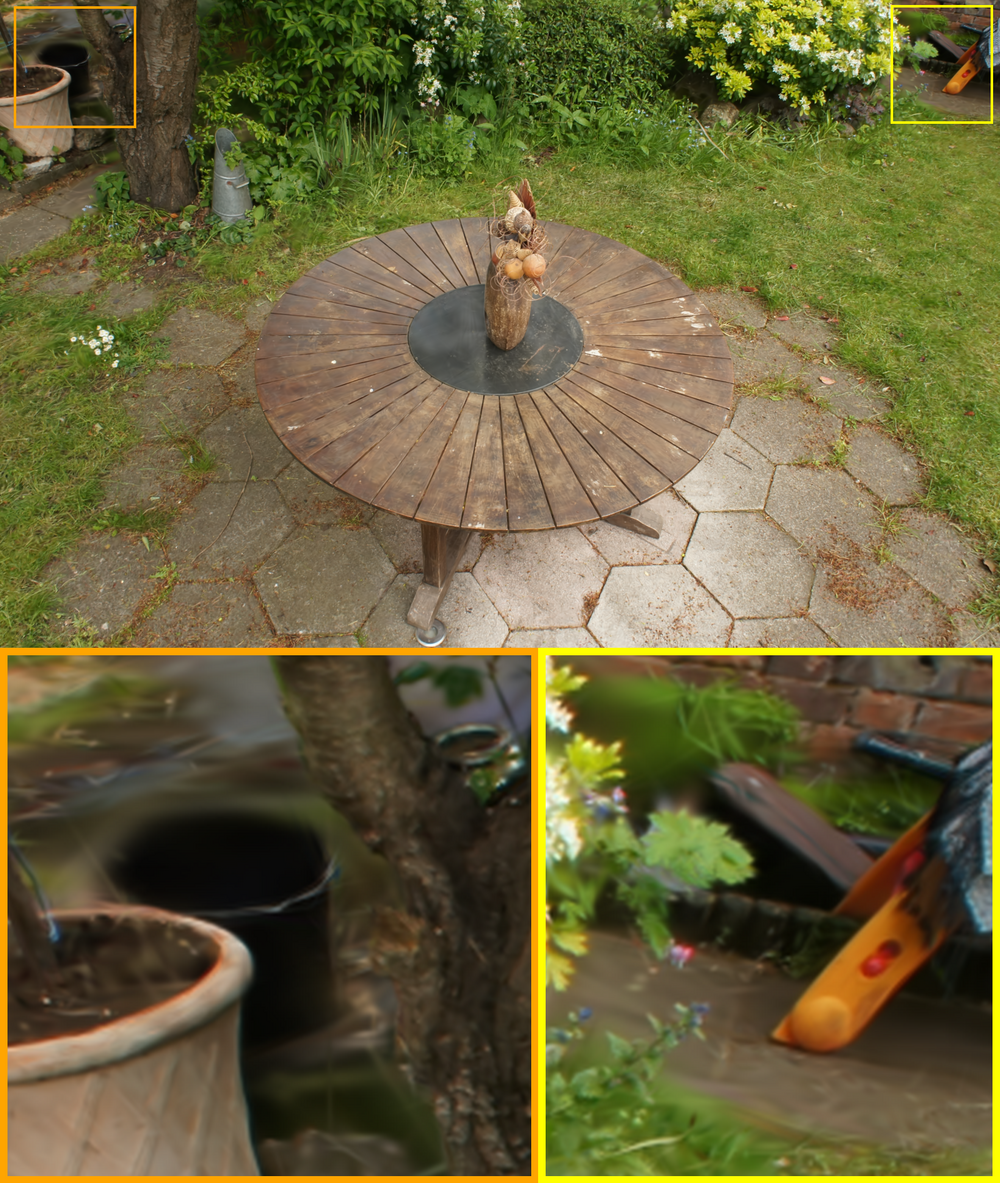} &
        \includegraphics[width=\linewidth]{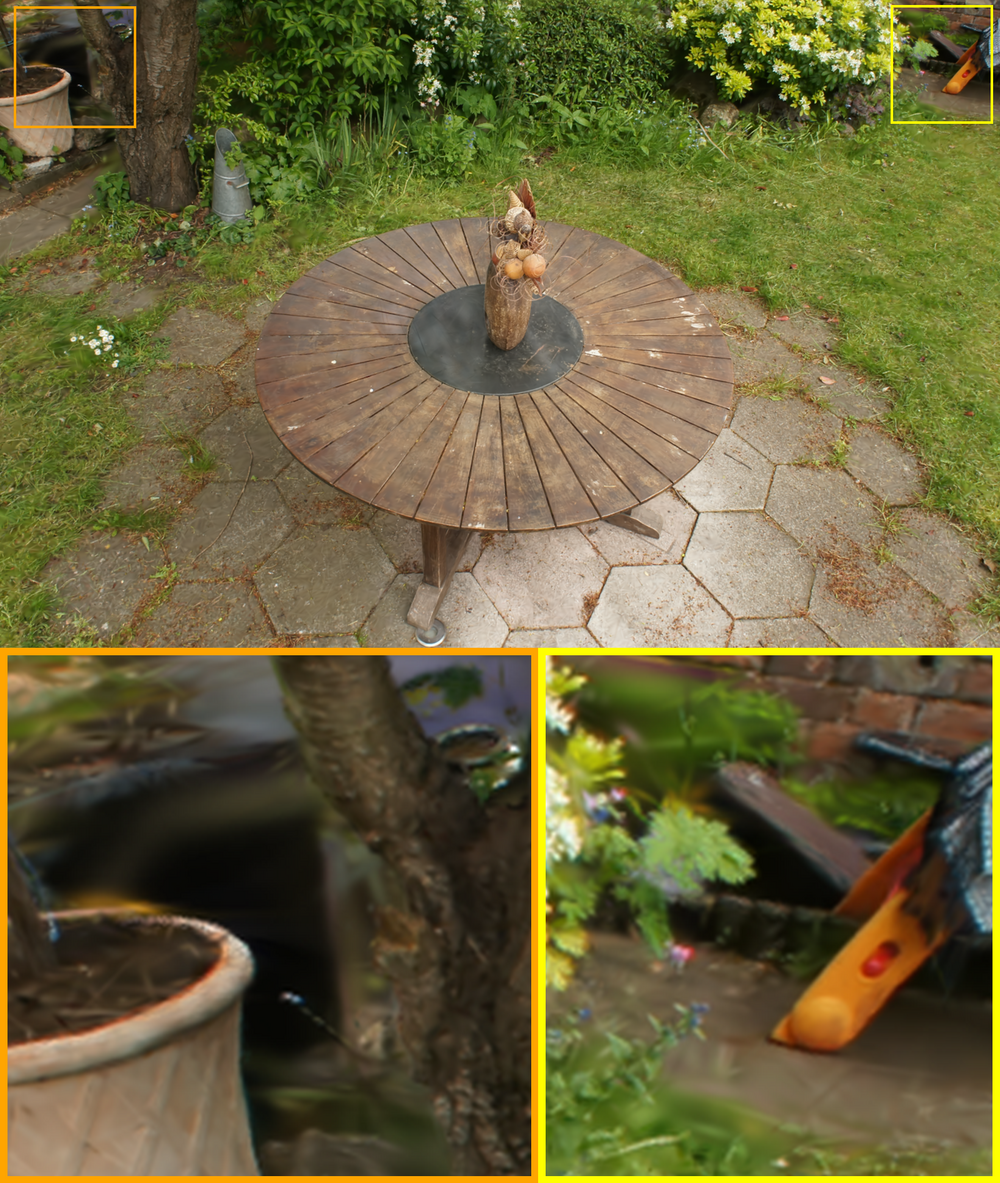} &
        \includegraphics[width=\linewidth]{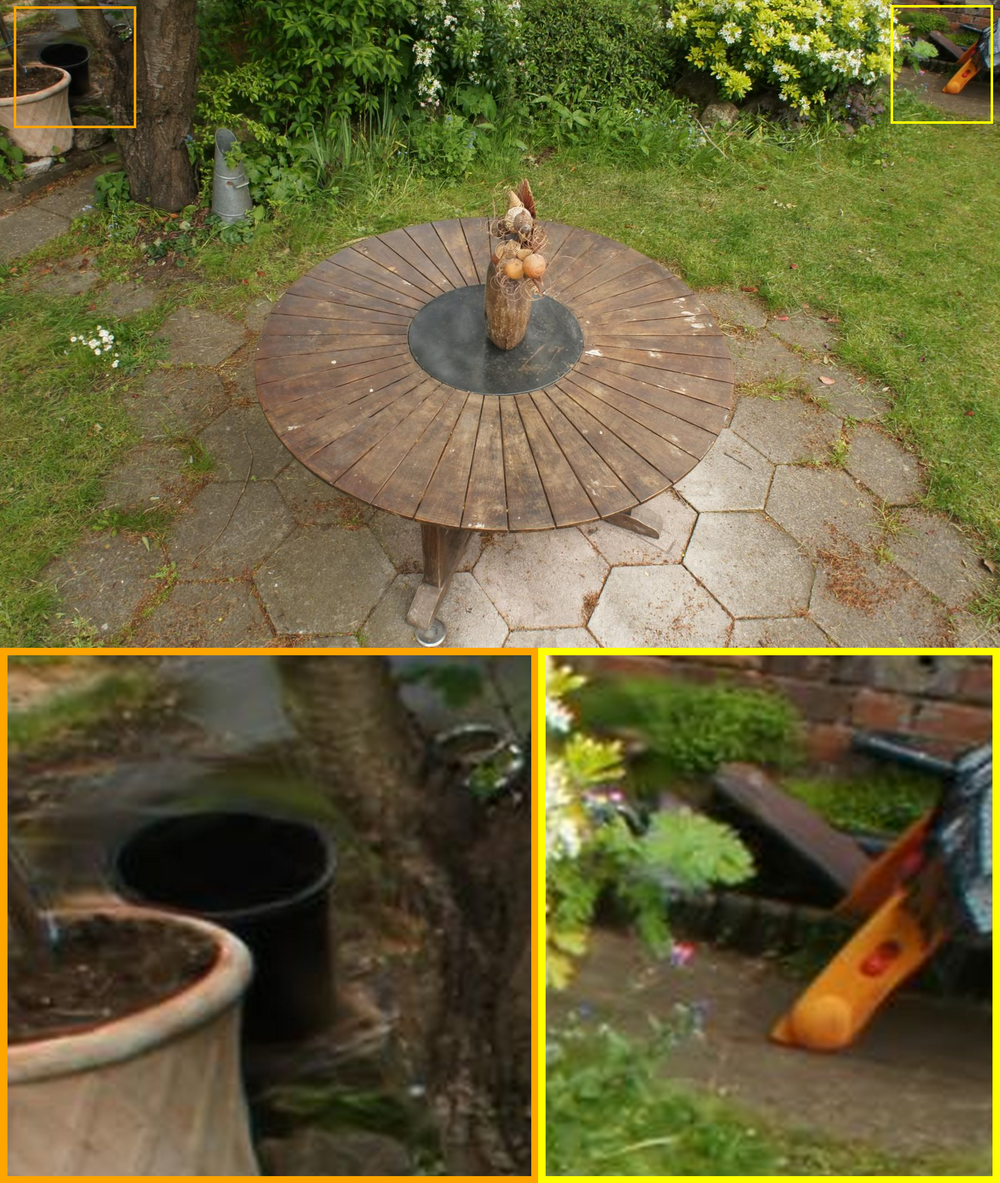} &
        \includegraphics[width=\linewidth]
        {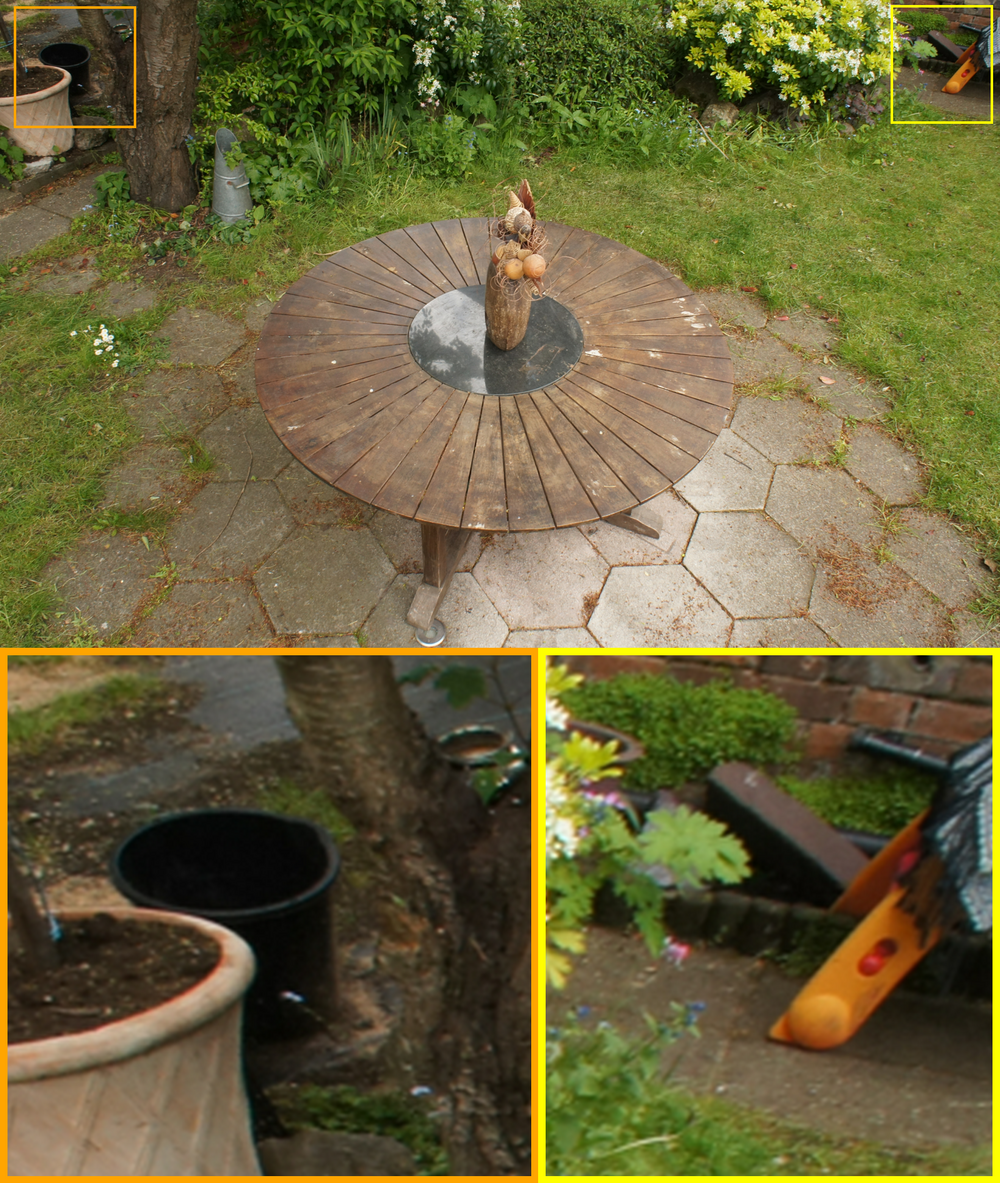} \\[\myrowgap]

        % === ROW 4: BICYCLE ===
        \includegraphics[width=\linewidth]{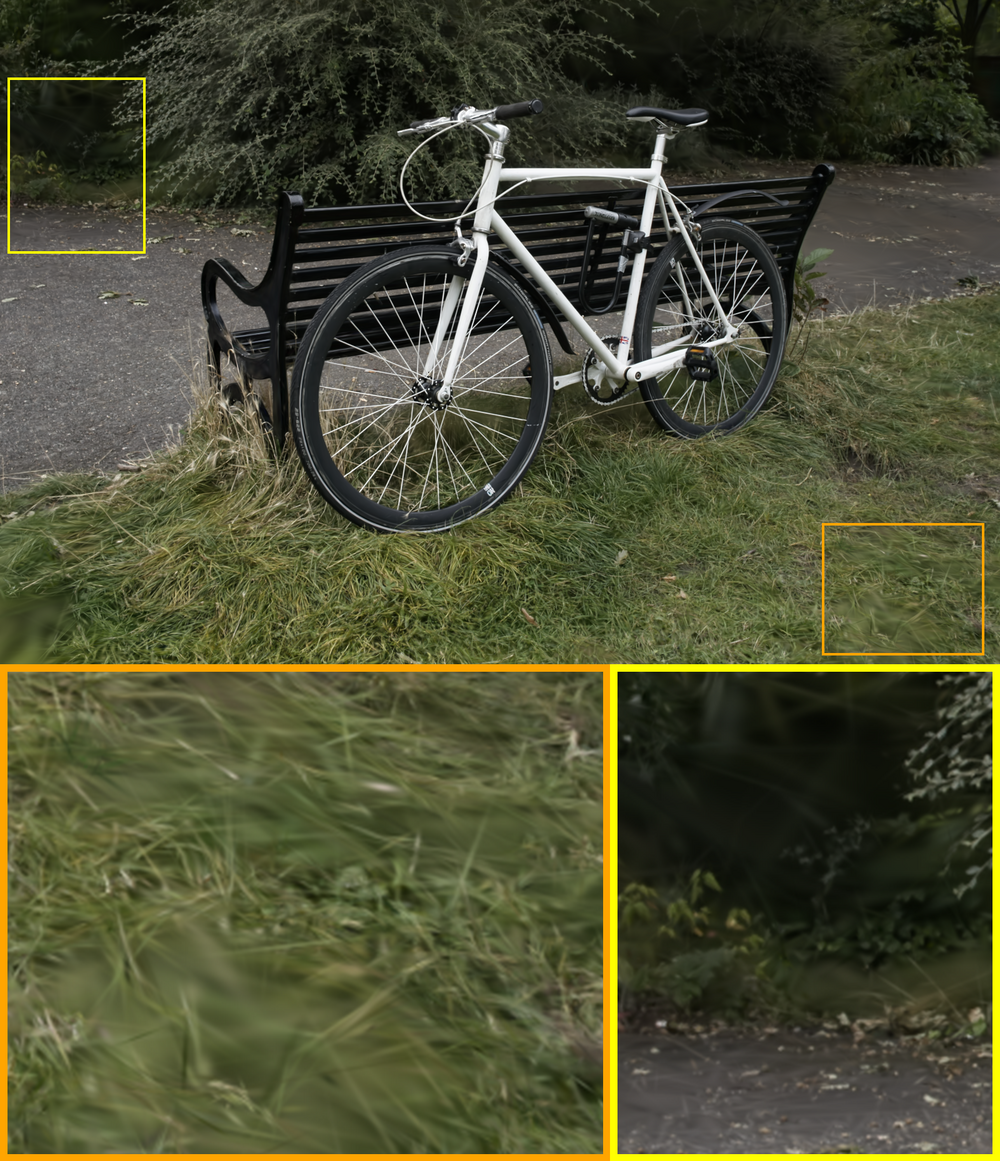} &
        \includegraphics[width=\linewidth]{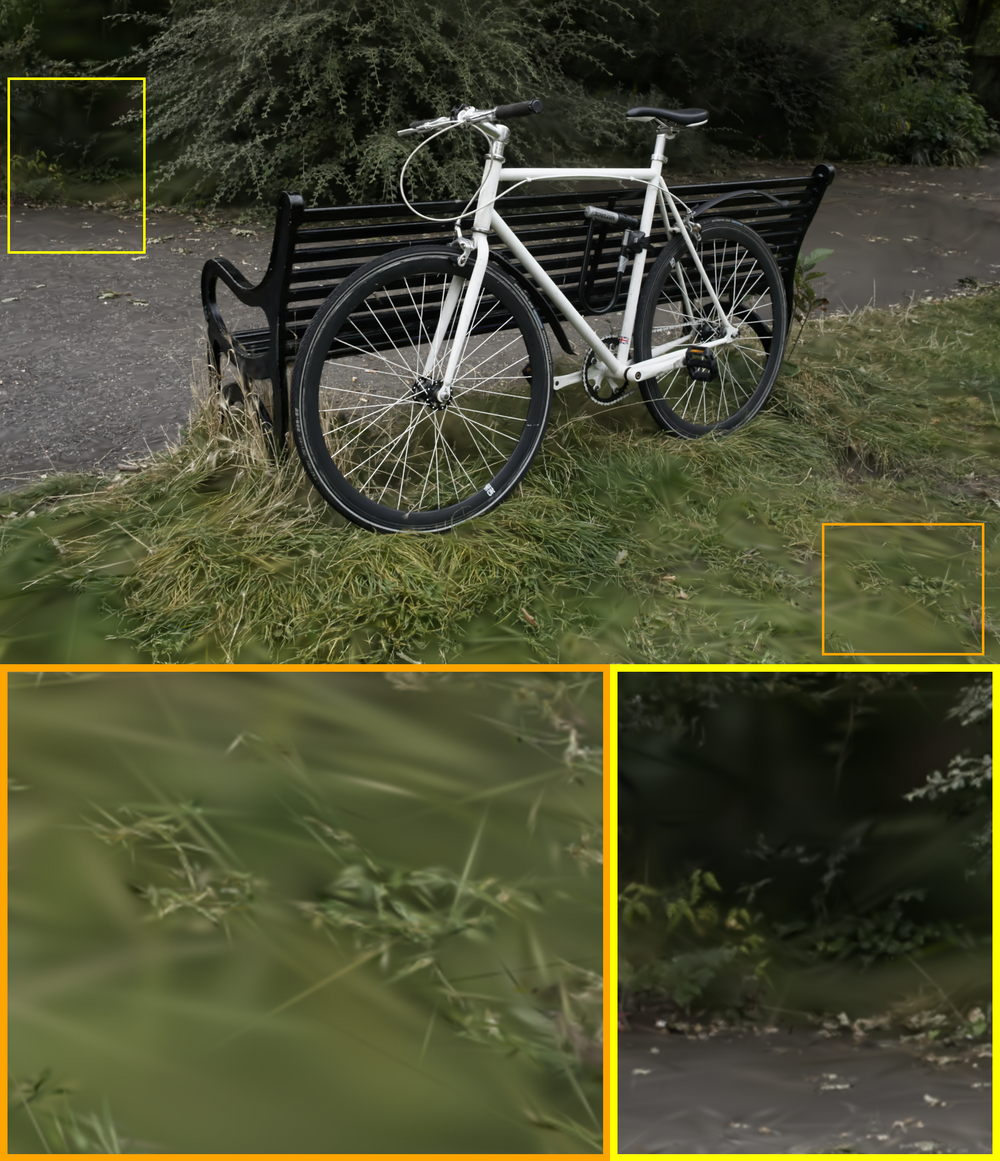} &
        \includegraphics[width=\linewidth]{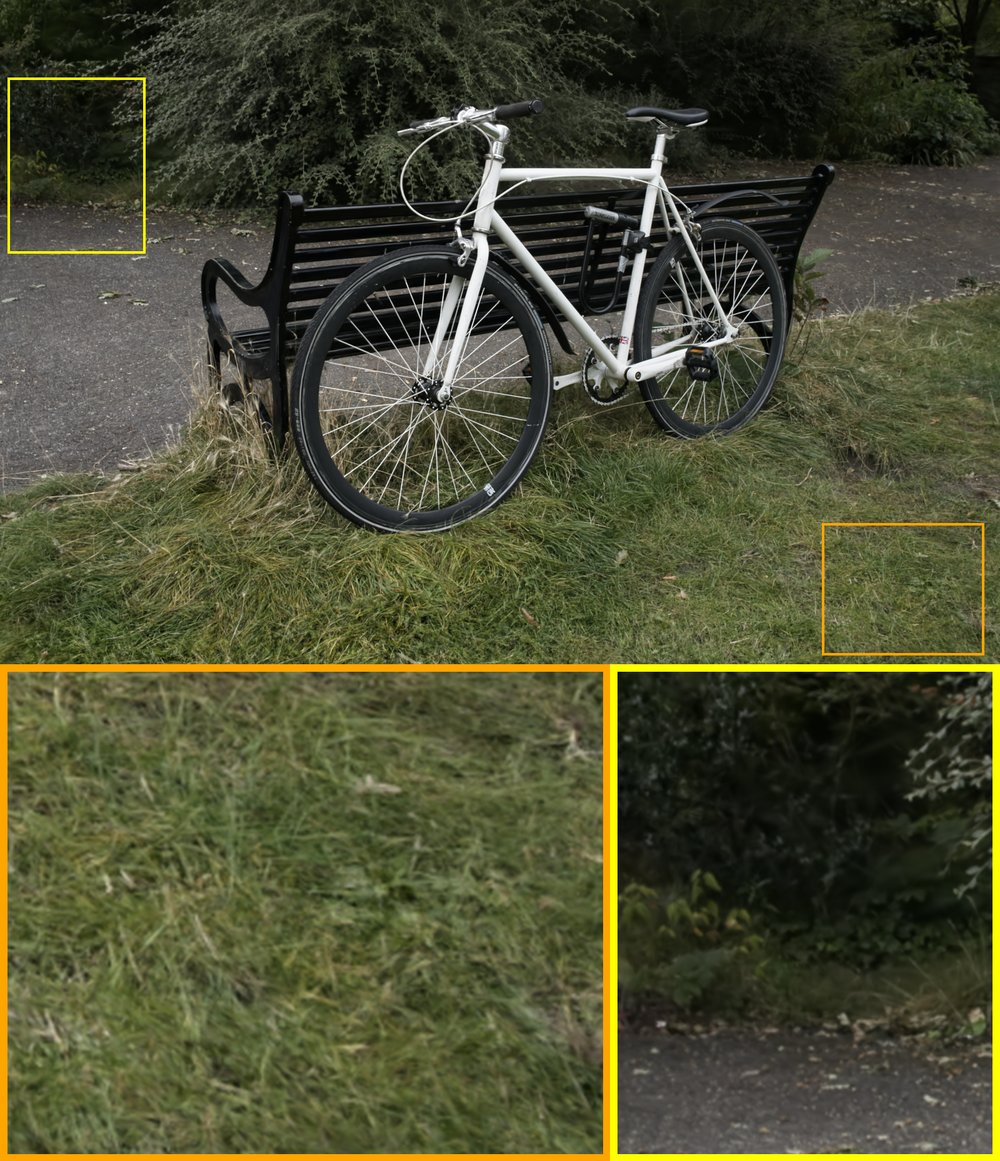} &
        \includegraphics[width=\linewidth]{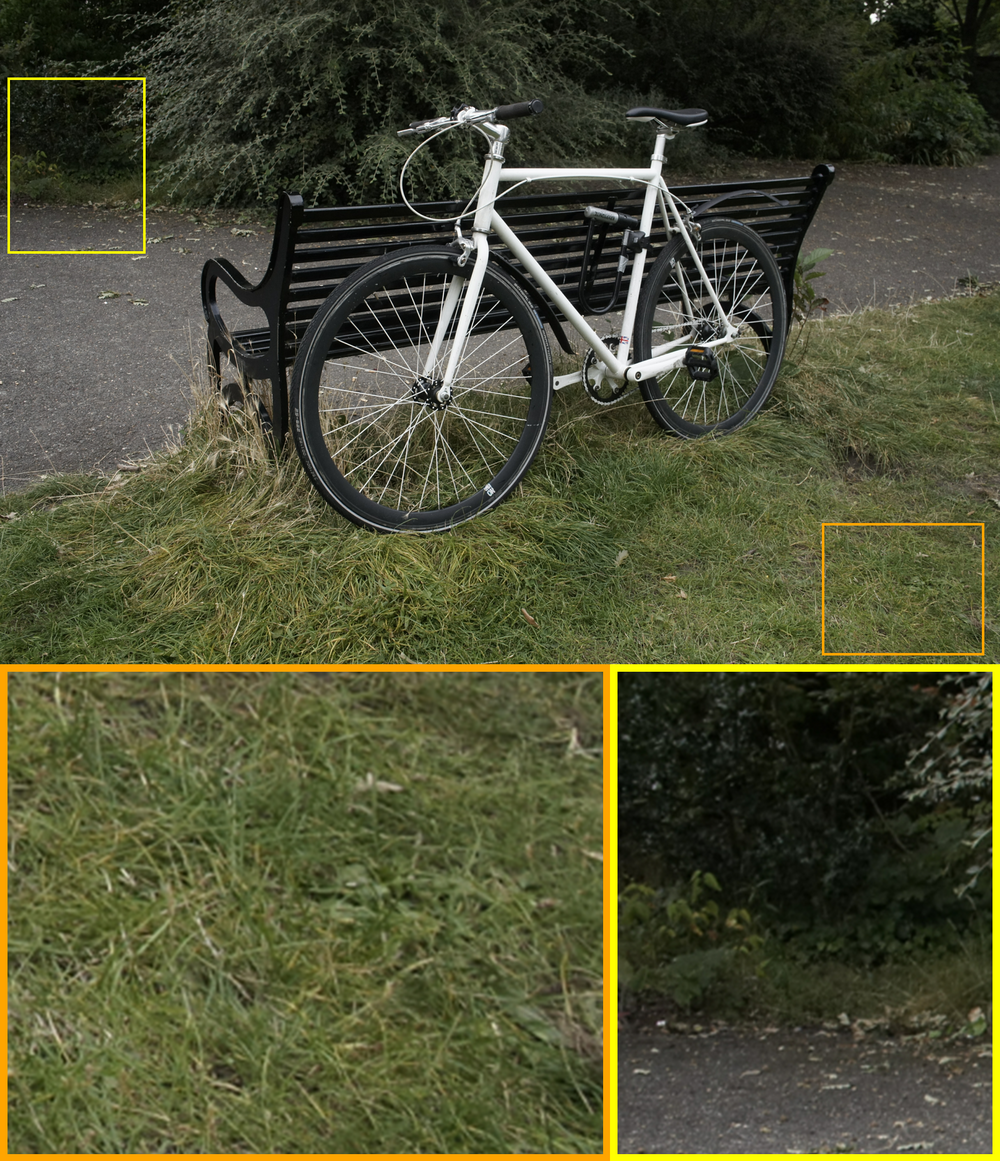} \\
    \end{tabular}
    
    \caption{\revised{Qualitative comparison on diverse Mip-NeRF 360 scenes. 
    (a) In the 'flowers' scene, our method reconstructs both intricate petal structures and distant foliage with significantly higher textural fidelity and fewer artifacts. 
    (b) In the 'room' scene, our method excels at rendering fine geometric details, such as the cabinet handle, while producing sharper textures on the bookshelf and floor. 
    (c) In the 'garden' scene, our method achieves a more complete reconstruction of complex objects (e.g., the bucket) and fine foliage, with a clear reduction in artifacts. 
    (d) In the 'bicycle' scene, our method dramatically improves the realism of the ground plane, rendering grass and soil with superior detail and textural accuracy.}}
    \label{fig:additional_qualitative_comparison}
\end{figure*}

\begin{figure*}[t]
    \centering
    
    % --- 修改1: 稍微减小每张图片的宽度，为左侧的标签列腾出空间 ---
    % 原为 0.23\textwidth，现调整为 0.225\textwidth
    \newcommand{\myimagewidth}{0.225\textwidth}
    
    % 定义列之间的水平间距 (无需修改)
    \newcommand{\mycolumngap}{2mm}
    
    % 定义行之间的垂直间距 (无需修改)
    \newcommand{\myrowgap}{2mm}

    % --- 修改2: 在表格定义的最左侧增加一个新列用于放置旋转的标签 ---
    % c: 一个标准的居中列
    % @{\hspace{4mm}}: 在标签列和第一张图片列之间增加 4mm 的间距
    \begin{tabular}{
        c @{\hspace{4mm}}
        >{\centering\arraybackslash}m{\myimagewidth}
        @{\hspace{\mycolumngap}}
        >{\centering\arraybackslash}m{\myimagewidth}
        @{\hspace{\mycolumngap}}
        >{\centering\arraybackslash}m{\myimagewidth}
        @{\hspace{\mycolumngap}}
        >{\centering\arraybackslash}m{\myimagewidth}
    }
        % === 列标题 ===
        % 在最左侧增加一个空的单元格，以保持对齐
        & \small 3DGS & \small GaussianPro & \small Ours & \small Ground Truth  \\[2mm]

        % === 第1行: Town01 ===
        % --- 修改3: 使用 \rotatebox{90}{...} 来创建竖排标签 ---
        \rotatebox{90}{\small\texttt{Town01}} &
        \includegraphics[width=\linewidth]
        {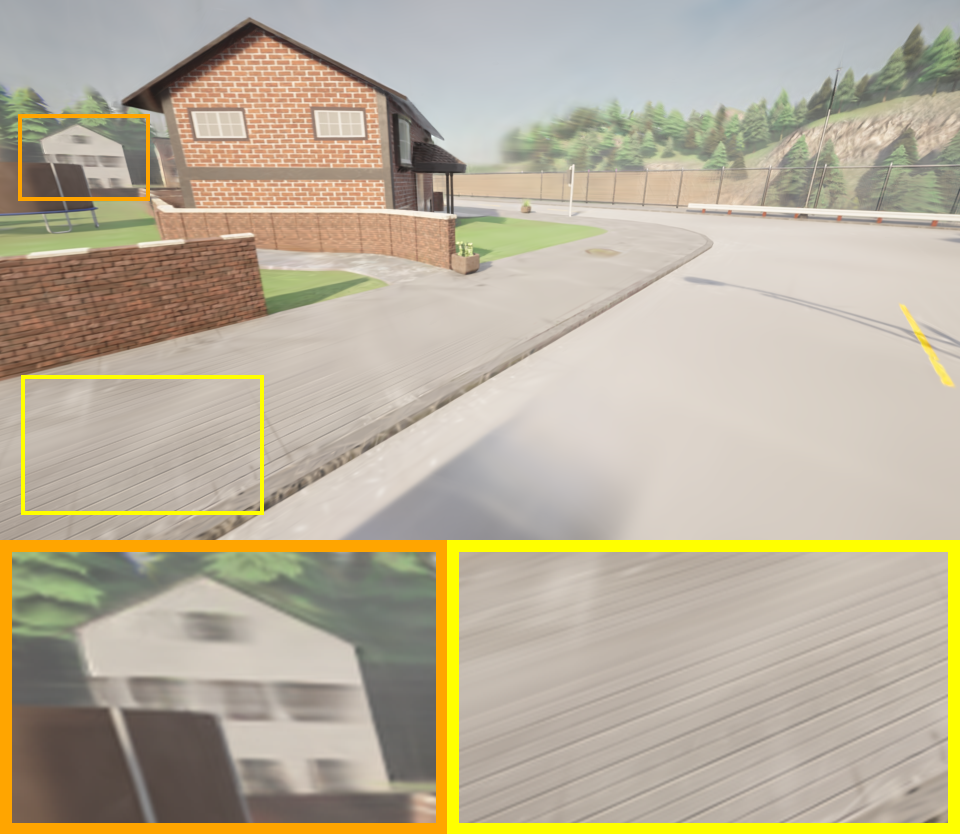} &
        \includegraphics[width=\linewidth]
        {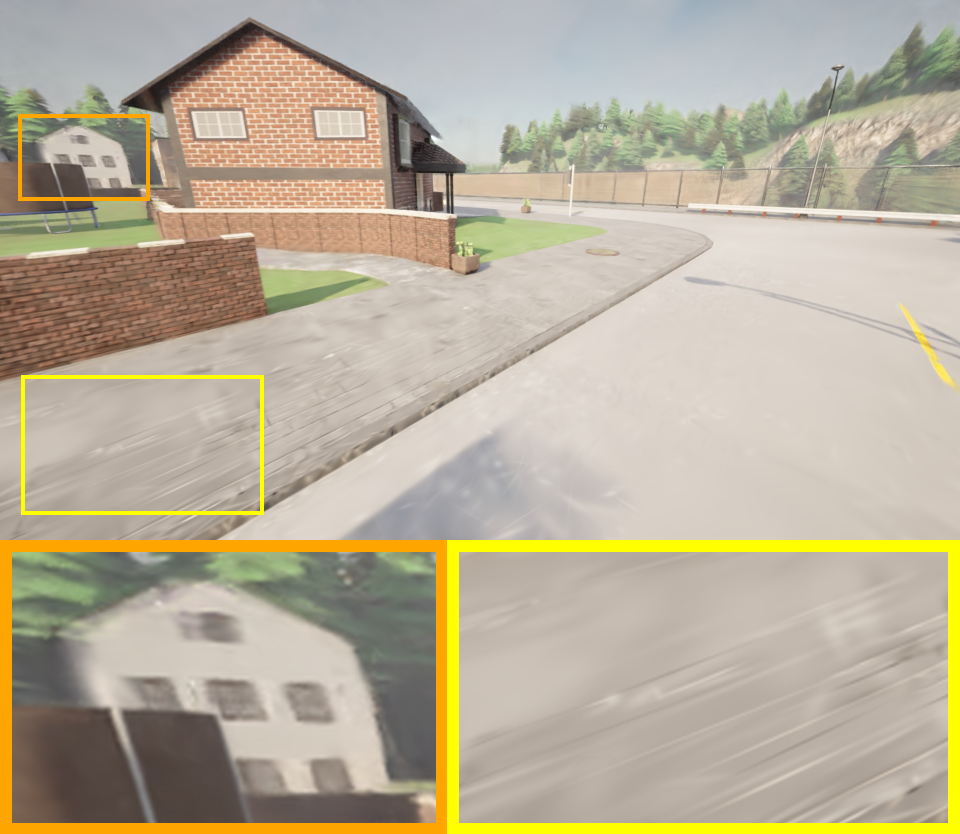} &
        \includegraphics[width=\linewidth]
        {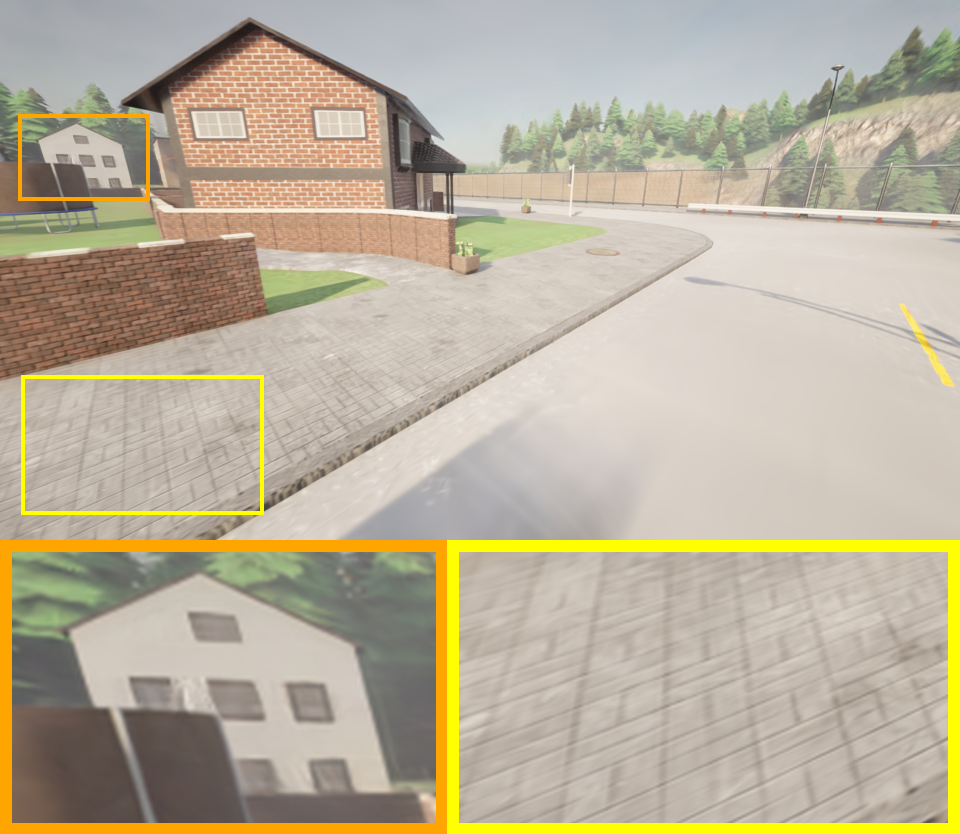} &
        \includegraphics[width=\linewidth]
        {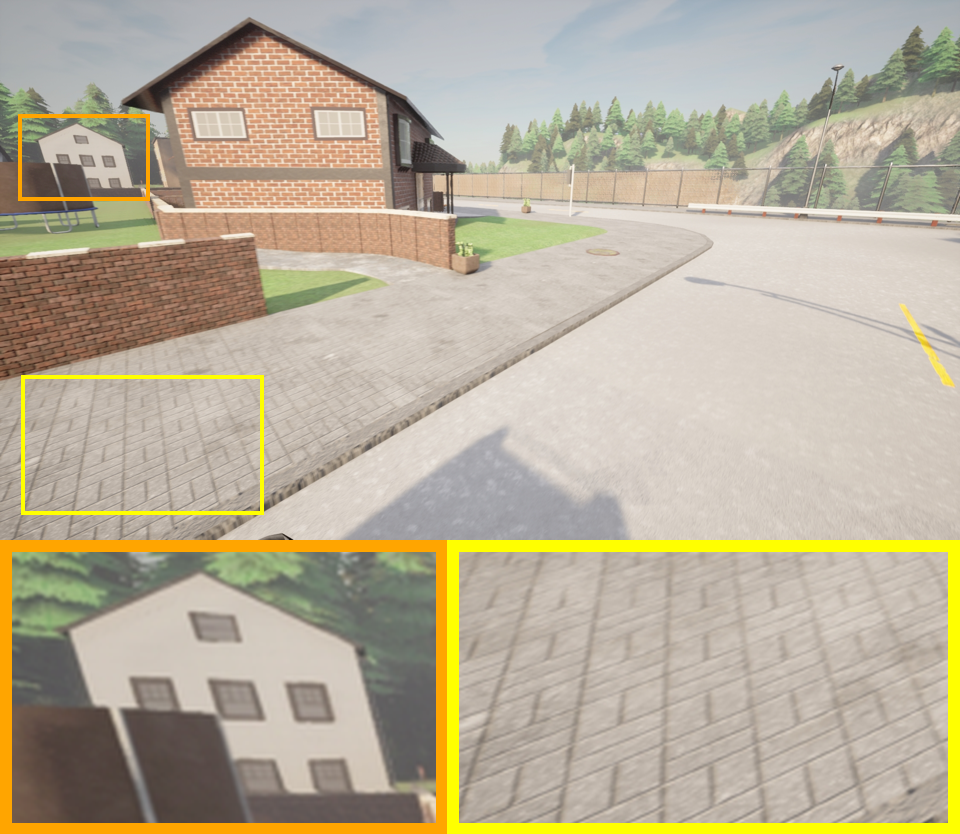} \\[\myrowgap]

        % === 第2行: Town02_v2 ===
        \rotatebox{90}{\small\texttt{Town02}} &
        \includegraphics[width=\linewidth]{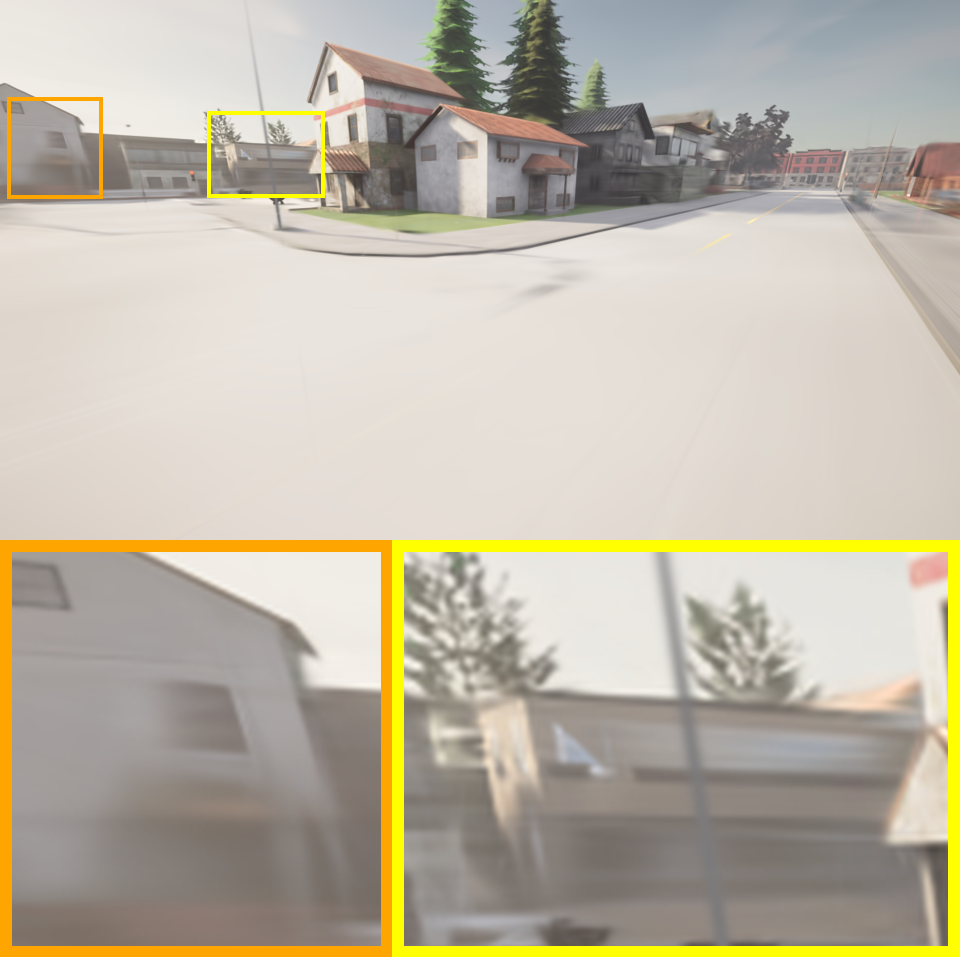} &
        \includegraphics[width=\linewidth]{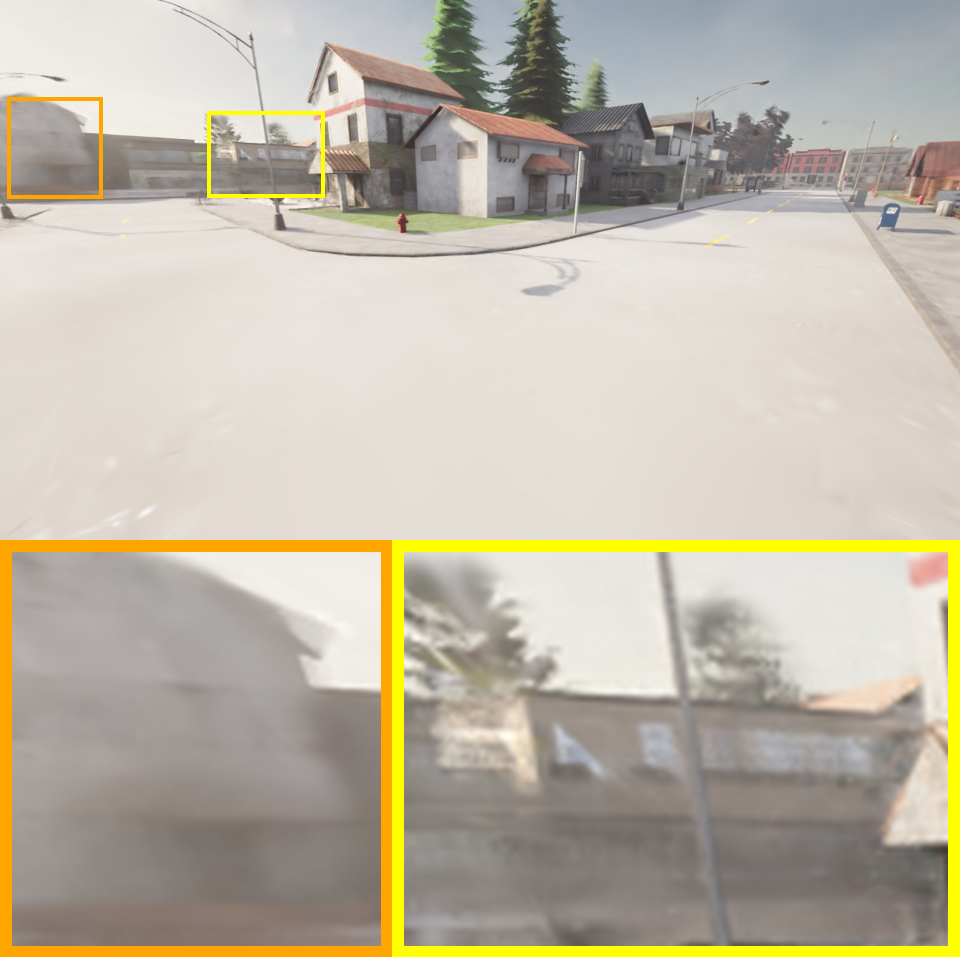} &
        \includegraphics[width=\linewidth]{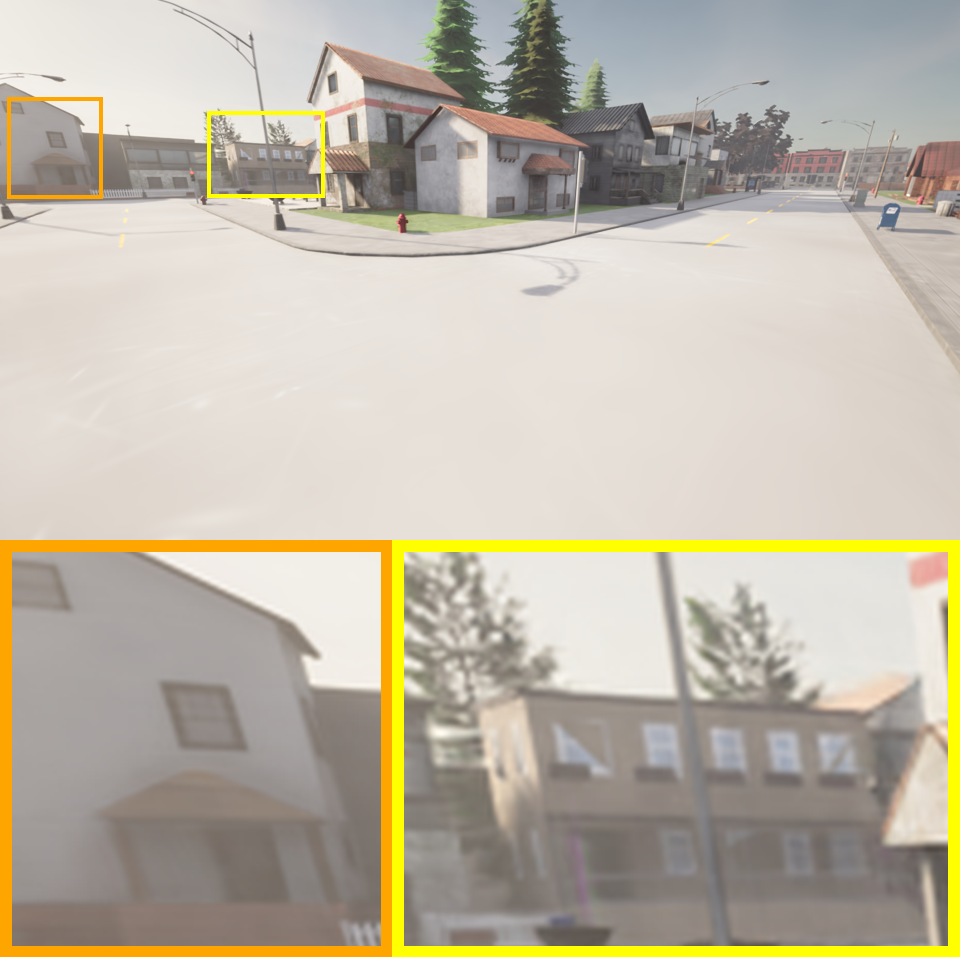} &
        \includegraphics[width=\linewidth]{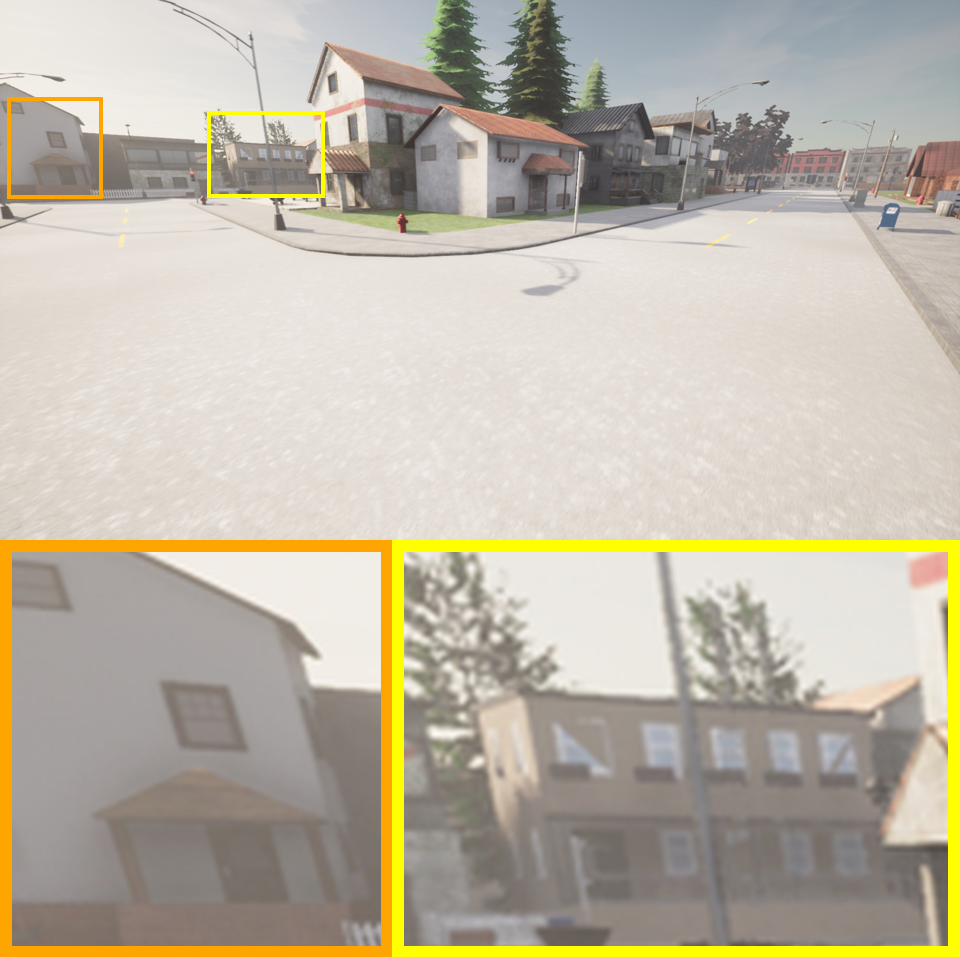} \\[\myrowgap]

        % === 第3行: Town03_v2 ===
        \rotatebox{90}{\small\texttt{Town03}} &
        \includegraphics[width=\linewidth]{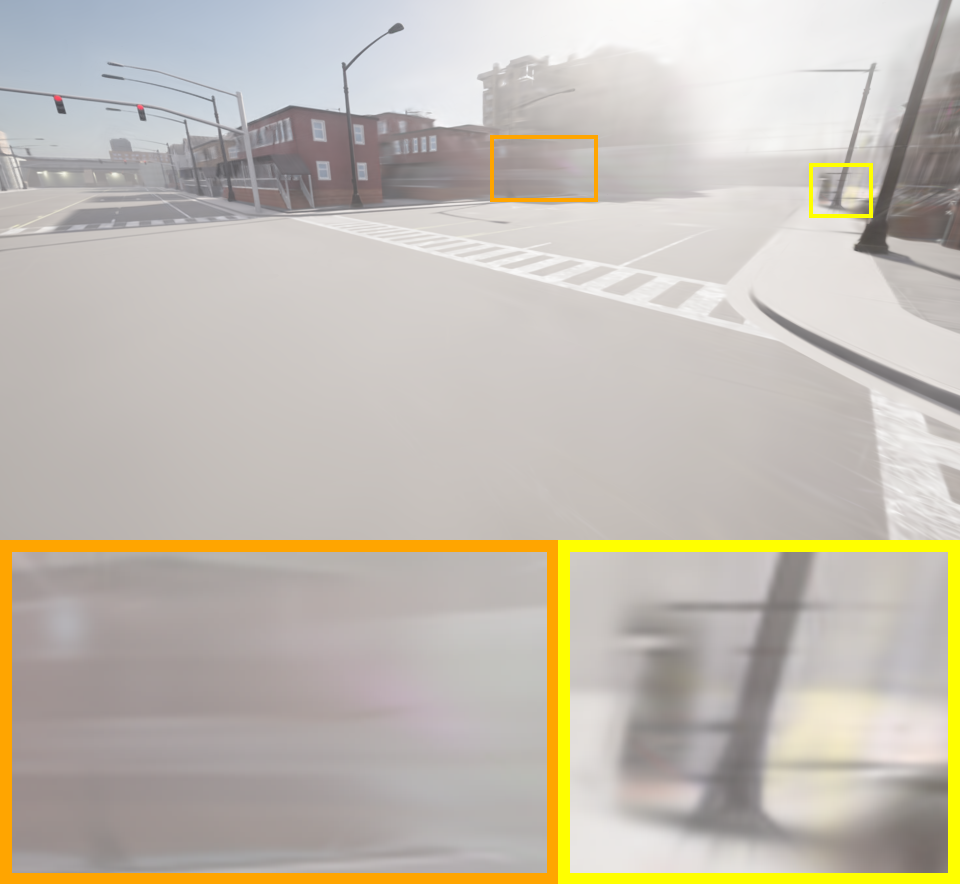} &
        \includegraphics[width=\linewidth]{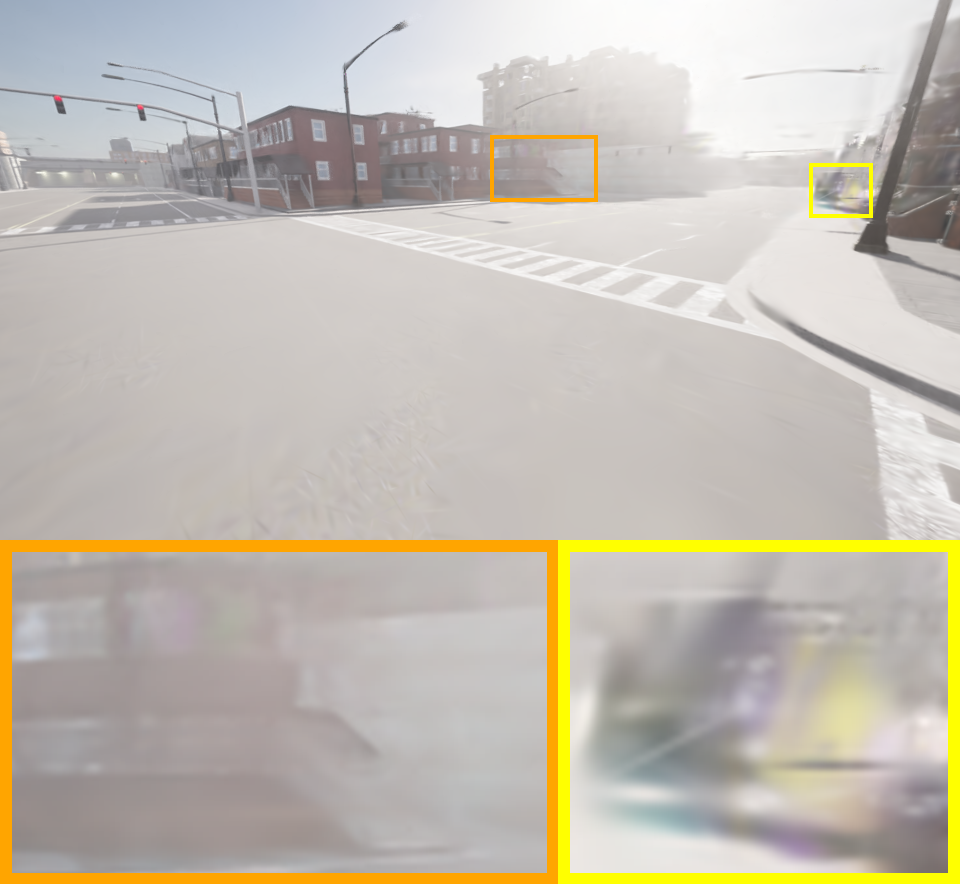} &
        \includegraphics[width=\linewidth]{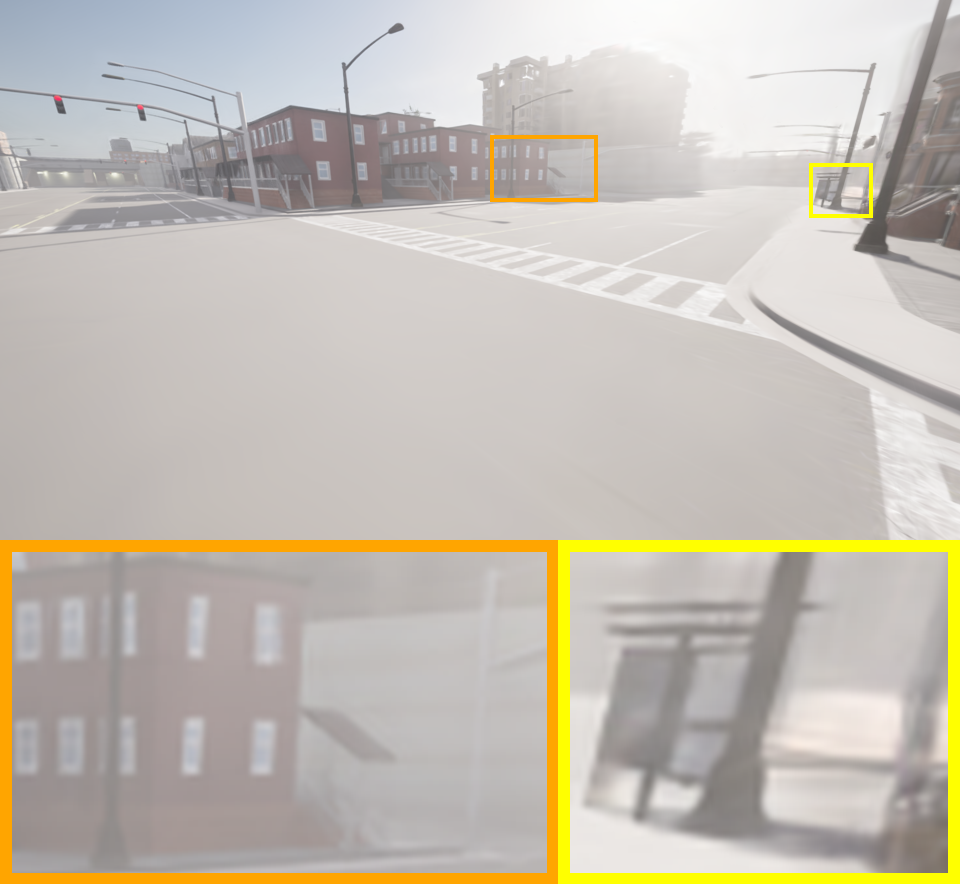} &
        \includegraphics[width=\linewidth]{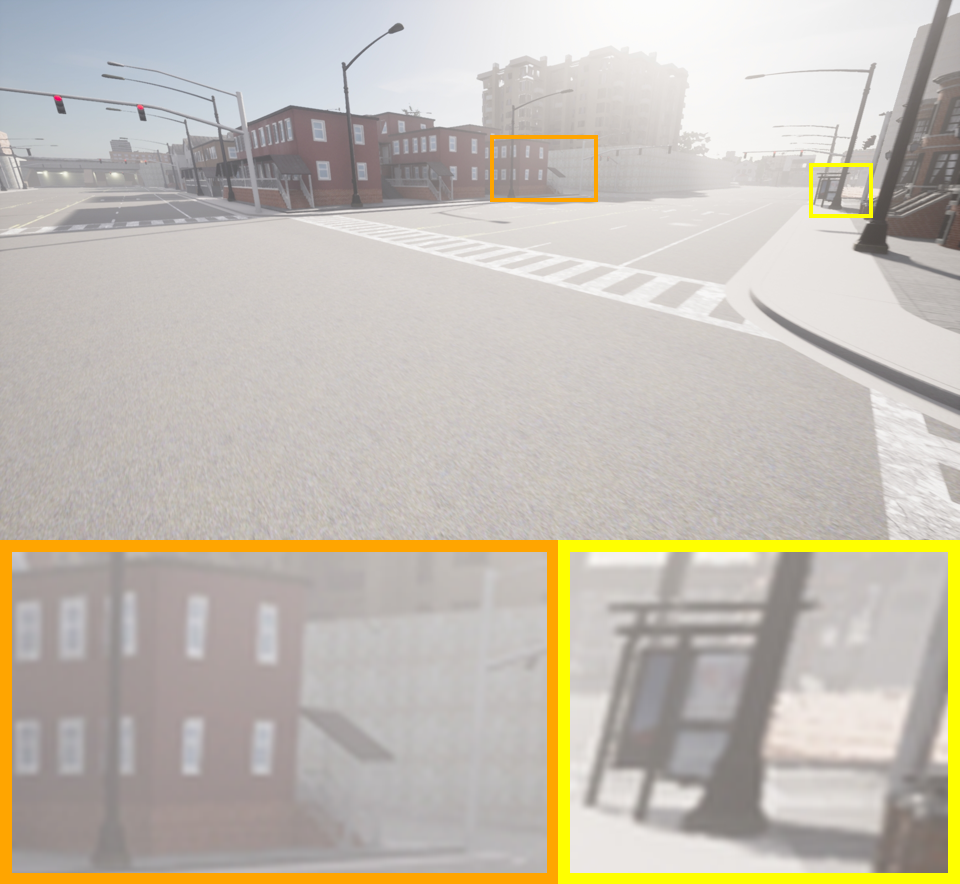} \\[\myrowgap]

        % === 第4行: Town04_v2 ===
        \rotatebox{90}{\small\texttt{Town04}} &
        \includegraphics[width=\linewidth]{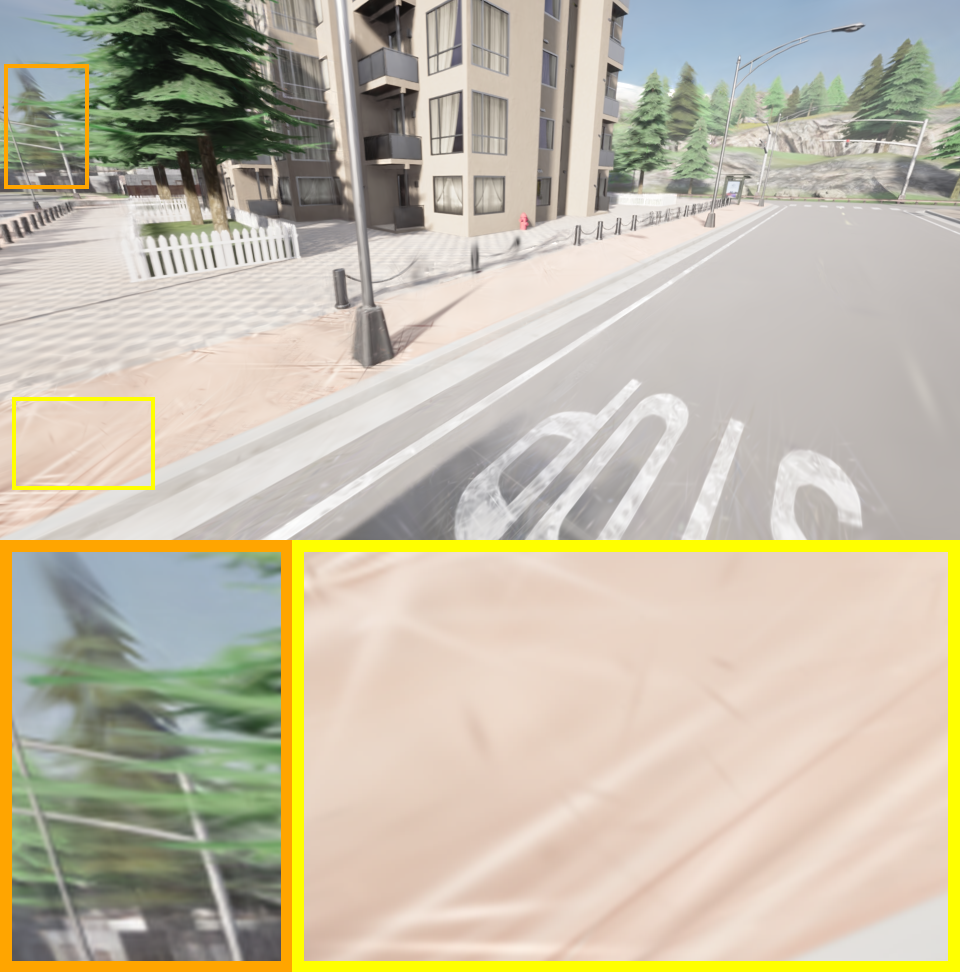} &
        \includegraphics[width=\linewidth]{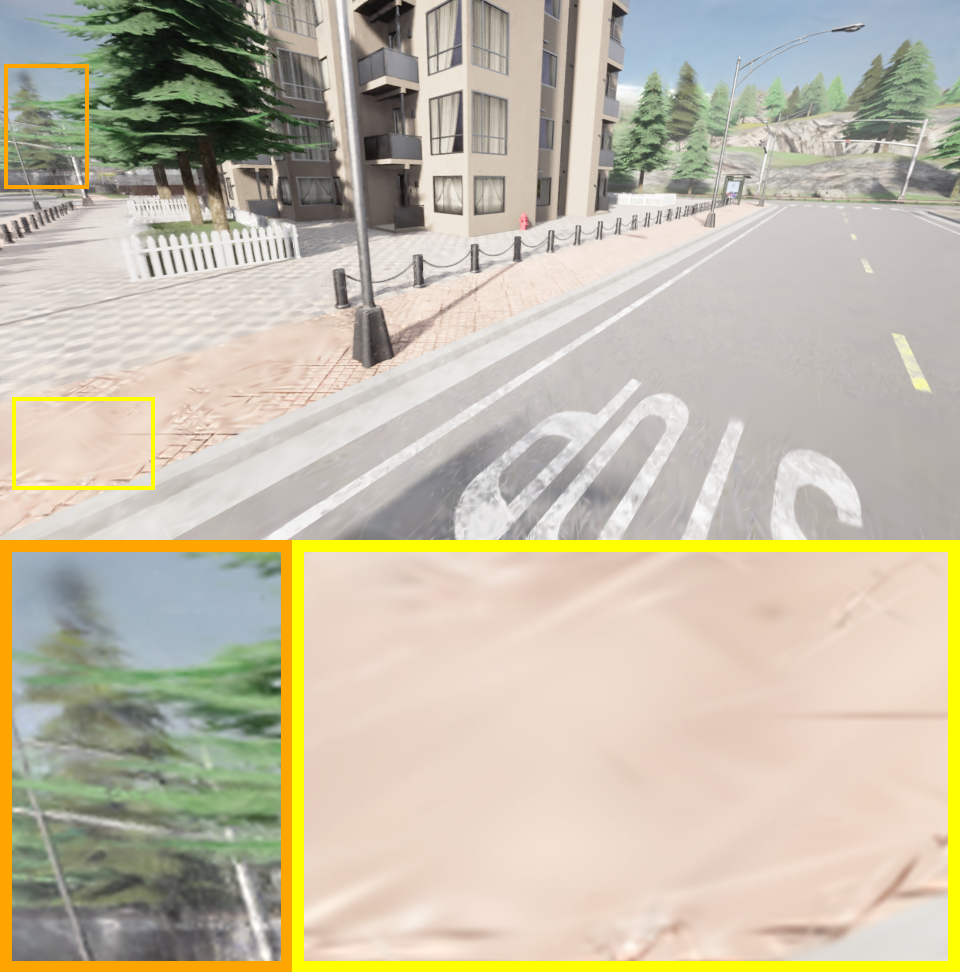} &
        \includegraphics[width=\linewidth]{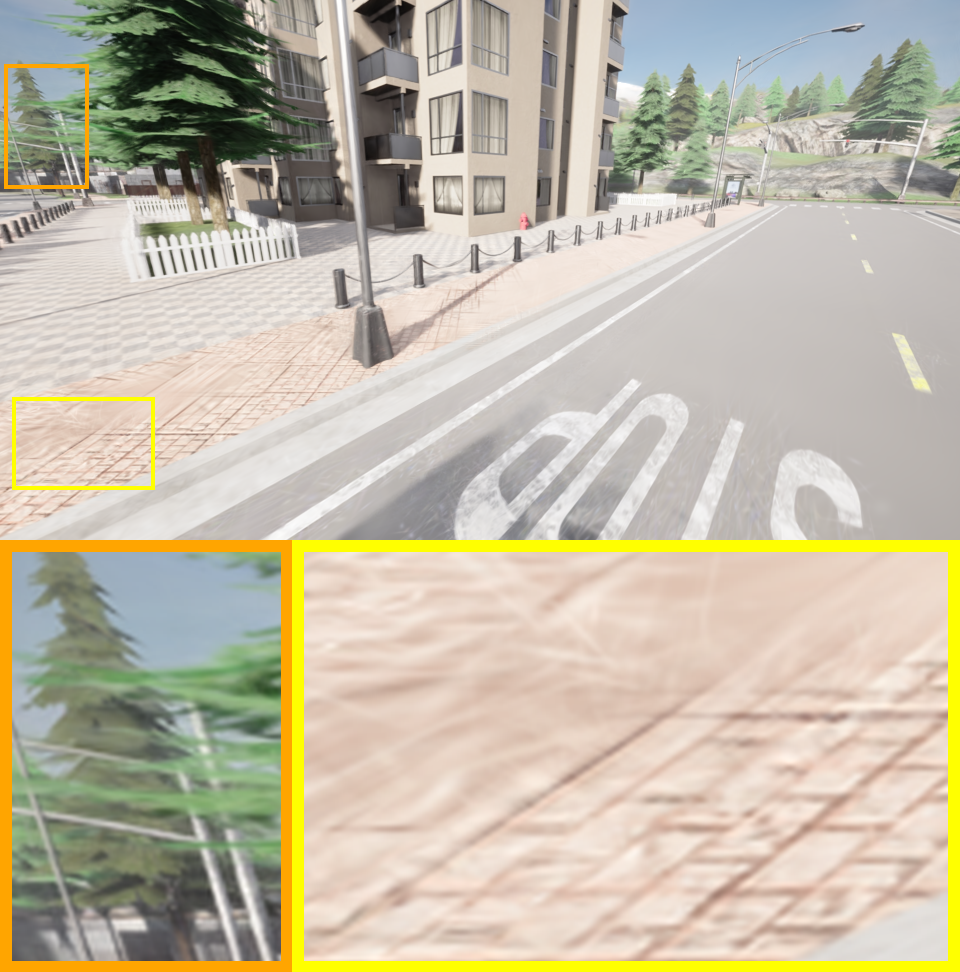} &
        \includegraphics[width=\linewidth]{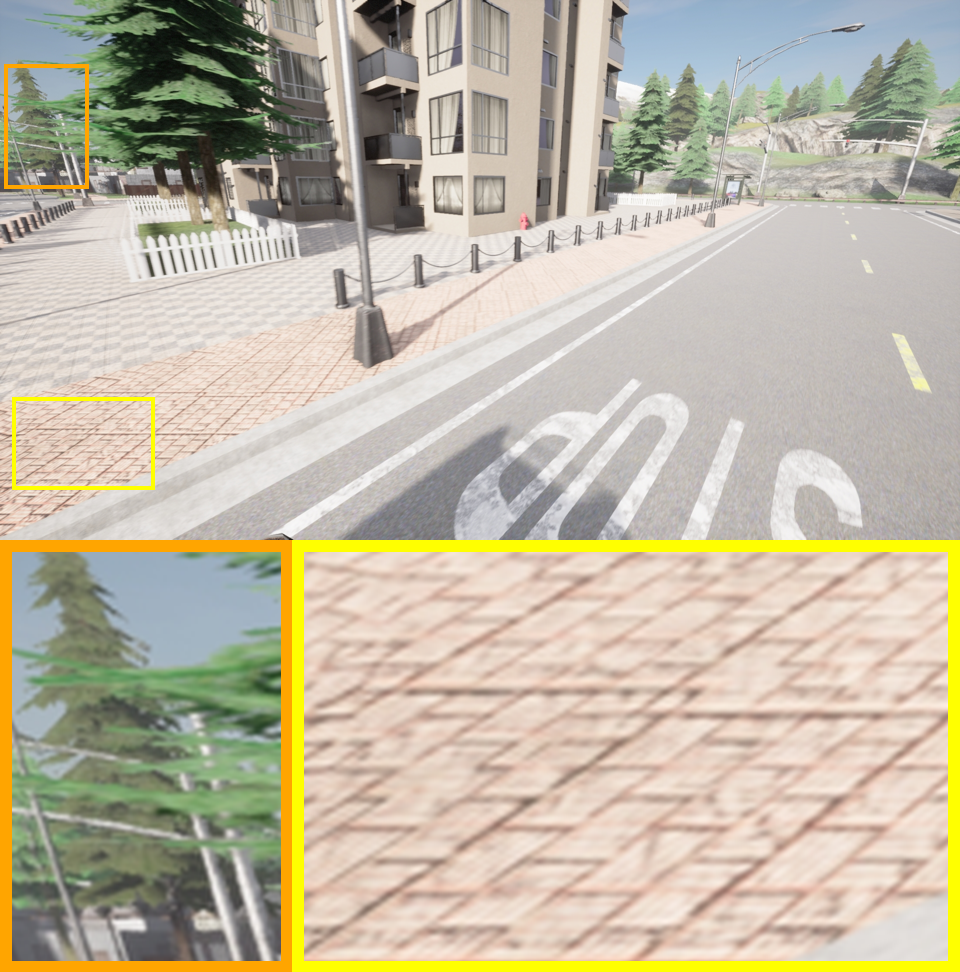} \\
    \end{tabular}
    
    \caption{\revised{Qualitative comparison on the synthetic SS3DM dataset. Our method consistently achieves superior visual quality across various scenes. 
    In \texttt{Town01}, the proposed method achieved higher fidelity in the reconstruction of building facades and road textures. 
    In \texttt{Town02}, our method rendered distant architecture with enhanced realism and detail, particularly at road corners characterized by sparse views.
    In \texttt{Town03}, under complex lighting, our approach preserved sharp structural details on buildings and roadside signs, mitigating the blurring artifacts observed in baseline results. 
    In \texttt{Town04}, our method generated more coherent geometry and clearer details for intricate elements, including road textures at sparsely-viewed corners and distant trees.}}
    \label{fig:ss3dm_qualitative_comparison}
\end{figure*}

\subsubsection{\revised{Maximum Blur Kernel Size}}

\revised{Our analysis of the 'maximum blur kernel size', a key parameter in our defocus model, reveals a nuanced but important finding, as shown in Table~\ref{tab:robustness_kernel_size}. We compared our default size of 7x7 against a smaller 3x3 kernel across four diverse scenes. The results indicate that a smaller kernel is a consistently strong choice. For scenes with prominent fine structures like 'room' and 'bicycle', the 3x3 kernel provides a modest but measurable performance benefit, improving PSNR by +0.23 dB and +0.03 dB, respectively. For other scenes such as 'garden' and 'flowers', the performance is highly robust to the kernel size, with negligible differences between the two settings. This comprehensive analysis validates our default 7x7 as a robust baseline, while also providing a clear, data-driven guideline for practitioners: for maximizing fidelity, especially in scenes with intricate geometry, selecting a smaller 3x3 kernel is a principled and effective optimization.}

\subsection{\revised{Additional Qualitative Comparisons on Mip-NeRF 360}}
\label{sec:appendix_qualitative_ablation}

\revised{To provide further visual evidence of our method's performance on diverse and challenging unbounded scenes, Figure~\ref{fig:additional_qualitative_comparison} presents qualitative comparisons against state-of-the-art methods on three scenes from the Mip-NeRF 360 dataset. These visualizations complement the quantitative results in Table~\ref{tab:results} and demonstrate our method's superior ability to reconstruct complex geometry and fine details.}

\subsection{\revised{Additional Qualitative Results on Diverse Datasets}}
\label{sec:appendix_additional_vis}

\revised{To further demonstrate the generalization capabilities of our framework, this section provides qualitative comparisons on the synthetic SS3DM dataset. As illustrated in Figure~\ref{fig:ss3dm_qualitative_comparison}, our method consistently produces reconstructions with higher fidelity and fewer artifacts compared to baseline methods.}

\printcredits
% \textbf{Baozhu Zhao}: Conceptualization, Methodology, Resources, Investigation. \textbf{Yu Deng}: Methodology, Resources, Software, Validation, Formal analysis, Investigation, Data curation, Visualization, Writing- Original draft preparation. \textbf{Junyan Su}: Resources, Writing - Review & Editing. \textbf{Qi Liu}: Writing - Review \& Editing, Supervision. \textbf{Xiaohan Zhang}: Writing - Review \& Editing, Validation. 

\bibliographystyle{cas-model2-names}

\bibliography{cas-refs}

@incollection{snavely2006photo,
  title={Photo tourism: exploring photo collections in 3D},
  author={Snavely, Noah and Seitz, Steven M and Szeliski, Richard},
  booktitle={ACM siggraph 2006 papers},
  pages={835--846},
  year={2006}
}

@inproceedings{schonberger2016structure,
  title={Structure-from-motion revisited},
  author={Schonberger, Johannes L and Frahm, Jan-Michael},
  booktitle={Proceedings of the IEEE conference on computer vision and pattern recognition},
  pages={4104--4113},
  year={2016}
}

@inproceedings{engel2014lsd,
  title={LSD-SLAM: Large-scale direct monocular SLAM},
  author={Engel, Jakob and Sch{\"o}ps, Thomas and Cremers, Daniel},
  booktitle={European conference on computer vision},
  pages={834--849},
  year={2014},
  organization={Springer}
}

@article{mur2015orb,
  title={ORB-SLAM: a versatile and accurate monocular SLAM system},
  author={Mur-Artal, Raul and Montiel, Jose Maria Martinez and Tardos, Juan D},
  journal={IEEE transactions on robotics},
  volume={31},
  number={5},
  pages={1147--1163},
  year={2015},
  publisher={IEEE}
}

@inproceedings{campbell2008using,
  title={Using multiple hypotheses to improve depth-maps for multi-view stereo},
  author={Campbell, Neill DF and Vogiatzis, George and Hern{\'a}ndez, Carlos and Cipolla, Roberto},
  booktitle={Computer Vision--ECCV 2008: 10th European Conference on Computer Vision, Marseille, France, October 12-18, 2008, Proceedings, Part I 10},
  pages={766--779},
  year={2008},
  organization={Springer}
}

@article{furukawa2015multi,
  title={Multi-view stereo: A tutorial},
  author={Furukawa, Yasutaka and Hern{\'a}ndez, Carlos and others},
  journal={Foundations and Trends{\textregistered} in Computer Graphics and Vision},
  volume={9},
  number={1-2},
  pages={1--148},
  year={2015},
  publisher={Now Publishers, Inc.}
}

@inproceedings{xu2019multi,
  title={Multi-scale geometric consistency guided multi-view stereo},
  author={Xu, Qingshan and Tao, Wenbing},
  booktitle={Proceedings of the IEEE/CVF conference on computer vision and pattern recognition},
  pages={5483--5492},
  year={2019}
}

@article{kar2017learning,
  title={Learning a multi-view stereo machine},
  author={Kar, Abhishek and H{\"a}ne, Christian and Malik, Jitendra},
  journal={Advances in neural information processing systems},
  volume={30},
  year={2017}
}

@inproceedings{ji2017surfacenet,
  title={Surfacenet: An end-to-end 3d neural network for multiview stereopsis},
  author={Ji, Mengqi and Gall, Juergen and Zheng, Haitian and Liu, Yebin and Fang, Lu},
  booktitle={Proceedings of the IEEE international conference on computer vision},
  pages={2307--2315},
  year={2017}
}

@inproceedings{yao2018mvsnet,
  title={Mvsnet: Depth inference for unstructured multi-view stereo},
  author={Yao, Yao and Luo, Zixin and Li, Shiwei and Fang, Tian and Quan, Long},
  booktitle={Proceedings of the European conference on computer vision (ECCV)},
  pages={767--783},
  year={2018}
}

@inproceedings{ma2022multiview,
  title={Multiview stereo with cascaded epipolar raft},
  author={Ma, Zeyu and Teed, Zachary and Deng, Jia},
  booktitle={European Conference on Computer Vision},
  pages={734--750},
  year={2022},
  organization={Springer}
}

@inproceedings{feng2023cvrecon,
  title={CVRecon: Rethinking 3d geometric feature learning for neural reconstruction},
  author={Feng, Ziyue and Yang, Liang and Guo, Pengsheng and Li, Bing},
  booktitle={Proceedings of the IEEE/CVF International Conference on Computer Vision},
  pages={17750--17760},
  year={2023}
}

@inproceedings{gu2020cascade,
  title={Cascade cost volume for high-resolution multi-view stereo and stereo matching},
  author={Gu, Xiaodong and Fan, Zhiwen and Zhu, Siyu and Dai, Zuozhuo and Tan, Feitong and Tan, Ping},
  booktitle={Proceedings of the IEEE/CVF conference on computer vision and pattern recognition},
  pages={2495--2504},
  year={2020}
}

@article{giang2021curvature,
  title={Curvature-guided dynamic scale networks for multi-view stereo},
  author={Giang, Khang Truong and Song, Soohwan and Jo, Sungho},
  journal={arXiv preprint arXiv:2112.05999},
  year={2021}
}

@article{chen2024survey,
  title={A survey on 3d gaussian splatting},
  author={Chen, Guikun and Wang, Wenguan},
  journal={arXiv preprint arXiv:2401.03890},
  year={2024}
}

@article{mildenhall2021nerf,
  title={Nerf: Representing scenes as neural radiance fields for view synthesis},
  author={Mildenhall, Ben and Srinivasan, Pratul P and Tancik, Matthew and Barron, Jonathan T and Ramamoorthi, Ravi and Ng, Ren},
  journal={Communications of the ACM},
  volume={65},
  number={1},
  pages={99--106},
  year={2021},
  publisher={ACM New York, NY, USA}
}

@inproceedings{sitzmann2019deepvoxels,
  title={Deepvoxels: Learning persistent 3d feature embeddings},
  author={Sitzmann, Vincent and Thies, Justus and Heide, Felix and Nie{\ss}ner, Matthias and Wetzstein, Gordon and Zollhofer, Michael},
  booktitle={Proceedings of the IEEE/CVF Conference on Computer Vision and Pattern Recognition},
  pages={2437--2446},
  year={2019}
}

@inproceedings{henzler2019escaping,
  title={Escaping plato's cave: 3d shape from adversarial rendering},
  author={Henzler, Philipp and Mitra, Niloy J and Ritschel, Tobias},
  booktitle={Proceedings of the IEEE/CVF International Conference on Computer Vision},
  pages={9984--9993},
  year={2019}
}

@article{muller2022instant,
  title={Instant neural graphics primitives with a multiresolution hash encoding},
  author={M{\"u}ller, Thomas and Evans, Alex and Schied, Christoph and Keller, Alexander},
  journal={ACM transactions on graphics (TOG)},
  volume={41},
  number={4},
  pages={1--15},
  year={2022},
  publisher={ACM New York, NY, USA}
}

@inproceedings{fridovich2022plenoxels,
  title={Plenoxels: Radiance fields without neural networks},
  author={Fridovich-Keil, Sara and Yu, Alex and Tancik, Matthew and Chen, Qinhong and Recht, Benjamin and Kanazawa, Angjoo},
  booktitle={Proceedings of the IEEE/CVF conference on computer vision and pattern recognition},
  pages={5501--5510},
  year={2022}
}

@inproceedings{barron2022mip,
  title={Mip-nerf 360: Unbounded anti-aliased neural radiance fields},
  author={Barron, Jonathan T and Mildenhall, Ben and Verbin, Dor and Srinivasan, Pratul P and Hedman, Peter},
  booktitle={Proceedings of the IEEE/CVF conference on computer vision and pattern recognition},
  pages={5470--5479},
  year={2022}
}

@inproceedings{wang2023f2,
  title={F2-nerf: Fast neural radiance field training with free camera trajectories},
  author={Wang, Peng and Liu, Yuan and Chen, Zhaoxi and Liu, Lingjie and Liu, Ziwei and Komura, Taku and Theobalt, Christian and Wang, Wenping},
  booktitle={Proceedings of the IEEE/CVF Conference on Computer Vision and Pattern Recognition},
  pages={4150--4159},
  year={2023}
}

@inproceedings{mildenhall2022nerf,
  title={Nerf in the dark: High dynamic range view synthesis from noisy raw images},
  author={Mildenhall, Ben and Hedman, Peter and Martin-Brualla, Ricardo and Srinivasan, Pratul P and Barron, Jonathan T},
  booktitle={Proceedings of the IEEE/CVF conference on computer vision and pattern recognition},
  pages={16190--16199},
  year={2022}
}

@article{wang2022nerfocus,
  title={Nerfocus: Neural radiance field for 3d synthetic defocus},
  author={Wang, Yinhuai and Yang, Shuzhou and Hu, Yujie and Zhang, Jian},
  journal={arXiv preprint arXiv:2203.05189},
  year={2022}
}

@inproceedings{ma2022deblur,
  title={Deblur-nerf: Neural radiance fields from blurry images},
  author={Ma, Li and Li, Xiaoyu and Liao, Jing and Zhang, Qi and Wang, Xuan and Wang, Jue and Sander, Pedro V},
  booktitle={Proceedings of the IEEE/CVF Conference on Computer Vision and Pattern Recognition},
  pages={12861--12870},
  year={2022}
}

@inproceedings{lee2023dp,
  title={Dp-nerf: Deblurred neural radiance field with physical scene priors},
  author={Lee, Dogyoon and Lee, Minhyeok and Shin, Chajin and Lee, Sangyoun},
  booktitle={Proceedings of the IEEE/CVF Conference on Computer Vision and Pattern Recognition},
  pages={12386--12396},
  year={2023}
}

@article{wang2024dof,
  title={DOF-GS: Adjustable Depth-of-Field 3D Gaussian Splatting for Refocusing, Defocus Rendering and Blur Removal},
  author={Wang, Yujie and Chakravarthula, Praneeth and Chen, Baoquan},
  journal={arXiv preprint arXiv:2405.17351},
  year={2024}
}

@article{kerbl20233d,
  title={3d gaussian splatting for real-time radiance field rendering.},
  author={Kerbl, Bernhard and Kopanas, Georgios and Leimk{\"u}hler, Thomas and Drettakis, George},
  journal={ACM Trans. Graph.},
  volume={42},
  number={4},
  pages={139--1},
  year={2023}
}

@misc{blanc2024raygaussvolumetricgaussianbasedray,
      title={RayGauss: Volumetric Gaussian-Based Ray Casting for Photorealistic Novel View Synthesis}, 
      author={Hugo Blanc and Jean-Emmanuel Deschaud and Alexis Paljic},
      year={2024},
      eprint={2408.03356},
      archivePrefix={arXiv},
      primaryClass={cs.CV},
      url={https://arxiv.org/abs/2408.03356}, 
}

@article{zwicker2002ewa,
  title={EWA splatting},
  author={Zwicker, Matthias and Pfister, Hanspeter and Van Baar, Jeroen and Gross, Markus},
  journal={IEEE Transactions on Visualization and Computer Graphics},
  volume={8},
  number={3},
  pages={223--238},
  year={2002},
  publisher={IEEE}
}

@inproceedings{cheng2024gaussianpro,
  title={Gaussianpro: 3d gaussian splatting with progressive propagation},
  author={Cheng, Kai and Long, Xiaoxiao and Yang, Kaizhi and Yao, Yao and Yin, Wei and Ma, Yuexin and Wang, Wenping and Chen, Xuejin},
  booktitle={Forty-first International Conference on Machine Learning},
  year={2024}
}

@article{li2024mipmap,
  title={Mipmap-GS: Let Gaussians Deform with Scale-specific Mipmap for Anti-aliasing Rendering},
  author={Li, Jiameng and Shi, Yue and Cao, Jiezhang and Ni, Bingbing and Zhang, Wenjun and Zhang, Kai and Van Gool, Luc},
  journal={arXiv preprint arXiv:2408.06286},
  year={2024}
}

@inproceedings{lee2024compact,
  title={Compact 3d gaussian representation for radiance field},
  author={Lee, Joo Chan and Rho, Daniel and Sun, Xiangyu and Ko, Jong Hwan and Park, Eunbyung},
  booktitle={Proceedings of the IEEE/CVF Conference on Computer Vision and Pattern Recognition},
  pages={21719--21728},
  year={2024}
}

@article{fan2023lightgaussian,
  title={Lightgaussian: Unbounded 3d gaussian compression with 15x reduction and 200+ fps},
  author={Fan, Zhiwen and Wang, Kevin and Wen, Kairun and Zhu, Zehao and Xu, Dejia and Wang, Zhangyang},
  journal={arXiv preprint arXiv:2311.17245},
  year={2023}
}

@inproceedings{yu2024mip,
  title={Mip-splatting: Alias-free 3d gaussian splatting},
  author={Yu, Zehao and Chen, Anpei and Huang, Binbin and Sattler, Torsten and Geiger, Andreas},
  booktitle={Proceedings of the IEEE/CVF Conference on Computer Vision and Pattern Recognition},
  pages={19447--19456},
  year={2024}
}

@inproceedings{huang20242d,
  title={2d gaussian splatting for geometrically accurate radiance fields},
  author={Huang, Binbin and Yu, Zehao and Chen, Anpei and Geiger, Andreas and Gao, Shenghua},
  booktitle={ACM SIGGRAPH 2024 conference papers},
  pages={1--11},
  year={2024}
}

@inproceedings{girish2024eagles,
  title={Eagles: Efficient accelerated 3d gaussians with lightweight encodings},
  author={Girish, Sharath and Gupta, Kamal and Shrivastava, Abhinav},
  booktitle={European Conference on Computer Vision},
  pages={54--71},
  year={2024},
  organization={Springer}
}

@article{kerbl2024hierarchical,
  title={A hierarchical 3d gaussian representation for real-time rendering of very large datasets},
  author={Kerbl, Bernhard and Meuleman, Andreas and Kopanas, Georgios and Wimmer, Michael and Lanvin, Alexandre and Drettakis, George},
  journal={ACM Transactions on Graphics (TOG)},
  volume={43},
  number={4},
  pages={1--15},
  year={2024},
  publisher={ACM New York, NY, USA}
}

@article{mai2024ever,
  title={Ever: Exact volumetric ellipsoid rendering for real-time view synthesis},
  author={Mai, Alexander and Hedman, Peter and Kopanas, George and Verbin, Dor and Futschik, David and Xu, Qiangeng and Kuester, Falko and Barron, Jonathan T and Zhang, Yinda},
  journal={arXiv preprint arXiv:2410.01804},
  year={2024}
}

@inproceedings{wang2024cinematic,
  title={Cinematic Gaussians: Real-Time HDR Radiance Fields with Depth of Field},
  author={Wang, Chao and Wolski, Krzysztof and Kerbl, Bernhard and Serrano, Ana and Bemana, Mojtaba and Seidel, Hans-Peter and Myszkowski, Karol and Leimk{\"u}hler, Thomas},
  booktitle={Computer Graphics Forum},
  volume={43},
  number={7},
  pages={e15214},
  year={2024},
  organization={Wiley Online Library}
}

@inproceedings{DBLP:conf/nips/0001WZG024,
  author       = {Yubin Hu and
                  Kairui Wen and
                  Heng Zhou and
                  Xiaoyang Guo and
                  Yong{-}Jin Liu},
  title        = {{SS3DM:} Benchmarking Street-View Surface Reconstruction with a Synthetic
                  3D Mesh Dataset},
  booktitle    = {Advances in Neural Information Processing Systems 38: Annual Conference
                  on Neural Information Processing Systems 2024, NeurIPS 2024, Vancouver,
                  BC, Canada, December 10 - 15, 2024},
  year         = {2024}
}

@InProceedings{Sun_2020_CVPR,
author = {Sun, Pei and Kretzschmar, Henrik and Dotiwalla, Xerxes and Chouard, Aurelien and Patnaik, Vijaysai and Tsui, Paul and Guo, James and Zhou, Yin and Chai, Yuning and Caine, Benjamin and Vasudevan, Vijay and Han, Wei and Ngiam, Jiquan and Zhao, Hang and Timofeev, Aleksei and Ettinger, Scott and Krivokon, Maxim and Gao, Amy and Joshi, Aditya and Zhang, Yu and Shlens, Jonathon and Chen, Zhifeng and Anguelov, Dragomir},
title = {Scalability in Perception for Autonomous Driving: Waymo Open Dataset},
booktitle = {Proceedings of the IEEE/CVF Conference on Computer Vision and Pattern Recognition (CVPR)},
month = {June},
year = {2020}
}

@inproceedings{DBLP:conf/icml/ChengLYY0MWC24,
  author       = {Kai Cheng and
                  Xiaoxiao Long and
                  Kaizhi Yang and
                  Yao Yao and
                  Wei Yin and
                  Yuexin Ma and
                  Wenping Wang and
                  Xuejin Chen},
  title        = {GaussianPro: 3D Gaussian Splatting with Progressive Propagation},
  booktitle    = {Forty-first International Conference on Machine Learning, {ICML} 2024,
                  Vienna, Austria, July 21-27, 2024},
  publisher    = {OpenReview.net},
  year         = {2024}
}

@inproceedings{barron2023zip,
  title={Zip-nerf: Anti-aliased grid-based neural radiance fields},
  author={Barron, Jonathan T and Mildenhall, Ben and Verbin, Dor and Srinivasan, Pratul P and Hedman, Peter},
  booktitle={Proceedings of the IEEE/CVF International Conference on Computer Vision},
  pages={19697--19705},
  year={2023}
}

@inproceedings{kaneko2022ar,
  title={Ar-nerf: Unsupervised learning of depth and defocus effects from natural images with aperture rendering neural radiance fields},
  author={Kaneko, Takuhiro},
  booktitle={Proceedings of the IEEE/CVF Conference on Computer Vision and Pattern Recognition},
  pages={18387--18397},
  year={2022}
}

@inproceedings{liu2024citygaussian,
  title={Citygaussian: Real-time high-quality large-scale scene rendering with gaussians},
  author={Liu, Yang and Luo, Chuanchen and Fan, Lue and Wang, Naiyan and Peng, Junran and Zhang, Zhaoxiang},
  booktitle={European Conference on Computer Vision},
  pages={265--282},
  year={2024},
  organization={Springer}
}

@article{radl2024stopthepop,
  title={Stopthepop: Sorted gaussian splatting for view-consistent real-time rendering},
  author={Radl, Lukas and Steiner, Michael and Parger, Mathias and Weinrauch, Alexander and Kerbl, Bernhard and Steinberger, Markus},
  journal={ACM Transactions on Graphics (TOG)},
  volume={43},
  number={4},
  pages={1--17},
  year={2024},
  publisher={ACM New York, NY, USA}
}

@article{cui2024dual,
  title={Dual-domain strip attention for image restoration},
  author={Cui, Yuning and Knoll, Alois},
  journal={Neural Networks},
  volume={171},
  pages={429--439},
  year={2024},
  publisher={Elsevier}
}

@inproceedings{wang2024efficient,
  title={Efficient LoFTR: Semi-dense local feature matching with sparse-like speed},
  author={Wang, Yifan and He, Xingyi and Peng, Sida and Tan, Dongli and Zhou, Xiaowei},
  booktitle={Proceedings of the IEEE/CVF Conference on Computer Vision and Pattern Recognition},
  pages={21666--21675},
  year={2024}
}

@article{hu2024metric3d,
  title={Metric3D v2: A Versatile Monocular Geometric Foundation Model for Zero-shot Metric Depth and Surface Normal Estimation},
  author={Hu, Mu and Yin, Wei and Zhang, Chi and Cai, Zhipeng and Long, Xiaoxiao and Chen, Hao and Wang, Kaixuan and Yu, Gang and Shen, Chunhua and Shen, Shaojie},
  journal={arXiv preprint arXiv:2404.15506},
  year={2024}
}

@article{almalioglu2022selfvio,
  title={SelfVIO: Self-supervised deep monocular Visual--Inertial Odometry and depth estimation},
  author={Almalioglu, Yasin and Turan, Mehmet and Saputra, Muhamad Risqi U and De Gusm{\~a}o, Pedro PB and Markham, Andrew and Trigoni, Niki},
  journal={Neural Networks},
  volume={150},
  pages={119--136},
  year={2022},
  publisher={Elsevier}
}

@article{liu2024joint,
  title={Joint estimation of pose, depth, and optical flow with a competition--cooperation transformer network},
  author={Liu, Xiaochen and Zhang, Tao and Liu, Mingming},
  journal={Neural Networks},
  volume={171},
  pages={263--275},
  year={2024},
  publisher={Elsevier}
}

@article{zhou2023miper,
  title={Miper-MVS: Multi-scale iterative probability estimation with refinement for efficient multi-view stereo},
  author={Zhou, Huizhou and Zhao, Haoliang and Wang, Qi and Hao, Gefei and Lei, Liang},
  journal={Neural Networks},
  volume={162},
  pages={502--515},
  year={2023},
  publisher={Elsevier}
}

@article{keypoint,
title = {Unsupervised distribution-aware keypoints generation from 3D point clouds},
journal = {Neural Networks},
volume = {173},
pages = {106158},
year = {2024},
issn = {0893-6080},
doi = {https://doi.org/10.1016/j.neunet.2024.106158},
url = {https://www.sciencedirect.com/science/article/pii/S0893608024000820},
author = {Yiqi Wu and Xingye Chen and Xuan Huang and Kelin Song and Dejun Zhang}
}

@article{wang2004image,
  title={Image quality assessment: from error visibility to structural similarity},
  author={Wang, Zhou and Bovik, Alan C and Sheikh, Hamid R and Simoncelli, Eero P},
  journal={IEEE transactions on image processing},
  volume={13},
  number={4},
  pages={600--612},
  year={2004},
  publisher={IEEE}
}

@inproceedings{zhang2018unreasonable,
  title={The unreasonable effectiveness of deep features as a perceptual metric},
  author={Zhang, Richard and Isola, Phillip and Efros, Alexei A and Shechtman, Eli and Wang, Oliver},
  booktitle={Proceedings of the IEEE conference on computer vision and pattern recognition},
  pages={586--595},
  year={2018}
}



\end{document}